\documentclass[times,twocolumn,final,longtitle]{elsarticle}

\usepackage[utf8]{inputenc}
\providecommand{\KWD}{\textbf{Keywords: }}
\providecommand{\snm}[1]{#1}
\usepackage{lineno}
\usepackage{siunitx} 

\usepackage{framed,multirow}
\usepackage{pdflscape}
\usepackage{rotating}   
\usepackage{longtable}  
\usepackage{graphicx}   
\usepackage{hyperref}   
\usepackage[table,xcdraw]{xcolor}
\usepackage{amsmath}
\usepackage{amssymb}
\usepackage{latexsym}
\usepackage{booktabs}
\usepackage{url}
\usepackage{xcolor}

\definecolor{newcolor}{rgb}{.8,.349,.1}

\definecolor{tblhead}{HTML}{2E3338}   
\definecolor{tblsub}{HTML}{565C63}    
\definecolor{tblrow}{HTML}{F0F1F2}    
\definecolor{tblline}{HTML}{C2C6CA}   
\hypersetup{colorlinks=true, urlcolor=tblhead, linkcolor=black, citecolor=black, breaklinks=true}

\journal{a peer-reviewed journal}

\begin{document}


\begin{frontmatter}

\title{Deep Learning for Retinal Degeneration Assessment: A Comprehensive Analysis of the MARIO Challenge}%



\author[1,2,3]{Rachid \snm{Zeghlache}}\corref{cor1}\cortext[cor1]{Corresponding author:}\ead{rachid.zeghlache@univ-brest.fr}
\author[1,8]{Ikram \snm{Brahim}}
\author[1,3]{Pierre-Henri \snm{Conze}}
\author[1,2]{Mathieu \snm{Lamard}}
\author[5]{Mohammed El Amine \snm{Lazouni}}
\author[5,6]{Zineb Aziza \snm{Elaouaber}}
\author[5]{Leila Ryma \snm{Lazouni}}



\author[18,19]{Christopher \snm{Nielsen}}
\author[18,19]{Ahmad O. \snm{Ahsan}}
\author[18,20,21,22,23,24]{Matthias \snm{Wilms}}
\author[18,20,21,25]{Nils D. \snm{Forkert}}

\author[17]{Lovre Antonio \snm{Budimir}}
\author[17]{Ivana \snm{Matovinović}}
\author[17]{Donik \snm{Vršnak}}
\author[17]{Sven \snm{Lončarić}}

\author[1,11]{Philippe \snm{Zhang}} 
\author[12]{Weili \snm{Jiang}}
\author[35]{Yihao \snm{Li}}


\author[9]{Yiding \snm{Hao}}


\author[16]{Markus \snm{Frohmann}}
\author[16]{Patrick \snm{Binder}}
\author[16]{Marcel \snm{Huber}}


\author[13]{Taha \snm{Emre}}
\author[13]{Teresa Finisterra \snm{Araújo}}
\author[13]{Marzieh \snm{Oghbaie}}
\author[13]{Hrvoje \snm{Bogunović}}


\author[14]{Amerens A. \snm{Bekkers}}
\author[14]{Nina M. \snm{van Liebergen}}
\author[14,15]{Hugo J. \snm{Kuijf}}


\author[33]{Abdul \snm{Qayyum}}
\author[34]{Moona \snm{Mazher}}
\author[33]{Steven A. \snm{Niederer}}


\author[29]{Alberto J. \snm{Beltrán-Carrero}}
\author[29,30]{Juan J. \snm{Gómez-Valverde}}
\author[31]{Javier \snm{Torresano-Rodríquez}}
\author[29]{Álvaro \snm{Caballero-Sastre}}
\author[29,30]{María J. \snm{Ledesma Carbayo}}


\author[10]{Yosuke \snm{Yamagishi}}


\author[32]{Yi \snm{Ding}}

\author[26,27]{Robin \snm{Peretzke}}
\author[26,27]{Alexandra \snm{Ertl}}
\author[26,27]{Maximilian \snm{Fischer}}
\author[26,27,28]{Jessica \snm{Kächele}}


\author[6]{Sofiane \snm{Zehar}}
\author[6]{Karim Boukli \snm{Hacene}}
\author[4]{Thomas \snm{Monfort}}
\author[1,2,7]{Béatrice \snm{Cochener}}
\author[1,2]{Mostafa El Habib Daho}\fnref{fn1}
\author[4]{Anas-Alexis Benyoussef}\fnref{fn1}
\author[1]{Gwenolé Quellec}

\address[1]{LaTIM UMR 1101, Inserm, Brest, France}
\address[2]{University of Western Brittany, Brest, France}
\address[3]{IMT Atlantique, Brest, France}
\address[4]{Service d'Ophtalmologie, CHRU Brest, Brest, France}
\address[5]{University of Tlemcen, Algeria}
\address[6]{LAZOUNI Ophthalmology Clinic, Tlemcen, Algeria}
\address[7]{Ophthalmology Department, CHRU Brest, Brest, France}
\address[8]{INSERM U1227 Lymphocytes B et Autoimmunite (LBAI), Brest, France}
\address[9]{Imperial College London, United Kingdom}
\address[10]{Division of Radiology and Biomedical Engineering, Graduate School of Medicine, The University of Tokyo, Tokyo, Japan}
\address[11]{Evolucare Technologies, France}
\address[12]{College of Computer Science, Sichuan University, China}
\address[13]{Medical University of Vienna, Austria}
\address[14]{TNO, The Hague, The Netherlands}
\address[15]{Image Sciences Institute, UMC Utrecht, Utrecht, The Netherlands}
\address[16]{Johannes Kepler University Linz, Austria}
\address[17]{University of Zagreb, Faculty of Electrical Engineering and Computing, Croatia}
\address[18]{Department of Radiology, University of Calgary, Calgary, AB, Canada}
\address[19]{Biomedical Engineering Graduate Program, University of Calgary, Calgary, AB, Canada}
\address[20]{Hotchkiss Brain Institute, University of Calgary, Calgary, AB, Canada}
\address[21]{Alberta Children’s Hospital Research Institute, University of Calgary, Calgary, AB, Canada}
\address[22]{Department of Pediatrics, University of Calgary, Calgary, AB, Canada}
\address[23]{Department of Community Health Sciences, University of Calgary, Calgary, AB, Canada}
\address[24]{Department of Clinical Neuroscience, University of Calgary, Calgary, AB, Canada}
\address[25]{University of Calgary, Calgary, AB, Canada}
\address[26]{German Cancer Research Center (DKFZ) Heidelberg, Division of Medical Image Computing, Germany}
\address[27]{Medical Faculty Heidelberg, Heidelberg University, Germany}
\address[28]{German Cancer Consortium (DKTK), DKFZ, Germany}
\address[29]{Biomedical Image Technologies (BIT), ETSI Telecomunicación, Universidad Politécnica de Madrid, Spain}
\address[30]{Centro de Investigación Biomédica en Red de Bioingeniería, Biomateriales y Nanomedicina (CIBER-BBN), Madrid, Spain}
\address[31]{Ophthalmology Service of the Provincial Ophthalmic Institute, Hospital Universitario Gregorio Marañón, Madrid, Spain}
\address[32]{University of Edinburgh, Scotland}
\address[33]{National Heart and Lung Institute, Faculty of Medicine, Imperial College London, United Kingdom}
\address[34]{Hawkes Institute, Department of Computer Science, University College London, London, United
Kingdom}
\address[35]{United Imaging Healthcare, China}
\fntext[fn1]{These authors contributed equally.}


\begin{abstract}
The MARIO challenge, organised at MICCAI 2024 in Marrakesh, is the first international benchmark for automated longitudinal monitoring of neovascular age-related macular degeneration (AMD) from optical coherence tomography (OCT). It comprised two complementary tasks: Task~1 classified the change in neovascular activity between two consecutive OCT B-scans (reduced, stable, or worsened), and Task~2 predicted disease progression over the following three months from a single visit (an ordinal prognostic task). The primary cohort comprised 136 patients from Brest, France (68 training / 34 validation / 34 test), imaged on a Heidelberg Spectralis device with paired infrared localiser images and clinical variables; an independent five-patient cohort from Tlemcen, Algeria, formed a held-out external test set probing population and device shift. Of 35 registered teams, 12 reached the final phase. On Task~1, the best team (MIPLAB) reached a composite score of 0.833 (macro-F1 = 0.859), approaching inter-expert agreement and showing that current models can reliably track short-term activity change. Task~2 proved markedly harder: the best composite score was only 0.306, quadratic weighted kappa stayed near zero, and inter-team agreement collapsed (mean Cohen's $\kappa$ = 0.10), indicating that no method recovered a reliable progression signal from a single time point. Performance fell substantially on the Algerian cohort and the ranking reshuffled entirely---the Task~1 winner dropped to fifth---underscoring the unsolved problem of cross-population, cross-device generalisation. MARIO thus establishes that short-term AMD activity monitoring is within reach of present-day methods, whereas single-visit progression prediction and domain-robust deployment remain open challenges.
\end{abstract}

\begin{keyword}
\KWD MARIO challenge \sep neovascular AMD \sep OCT \sep disease progression \sep longitudinal monitoring \sep anti-VEGF \sep deep learning \sep foundation models \sep domain generalisation \sep MICCAI challenge
\end{keyword}

\end{frontmatter}



\section{Introduction}

Age-related macular degeneration (AMD) represents one of the leading causes of irreversible visual impairment worldwide, affecting approximately 196 million people globally \citep{SCHULTZ20211792}. This progressive neurodegenerative retinal disease primarily impacts individuals over 65 years of age, causing substantial central vision loss while typically preserving peripheral vision. AMD manifests across a spectrum of severity, with advanced stages --- geographic atrophy (GA) and neovascular AMD (nAMD) --- affecting approximately 20\% of patients and constituting the predominant cause of severe visual impairment in developed nations \citep{Wong2014,Fleckenstein2018}. The multifactorial etiology of AMD, involving complex interactions between genetic susceptibility and environmental risk factors, presents significant challenges for both prevention strategies and therapeutic interventions.

\begin{figure*}[t]
	\centering
	\includegraphics[width=\textwidth]{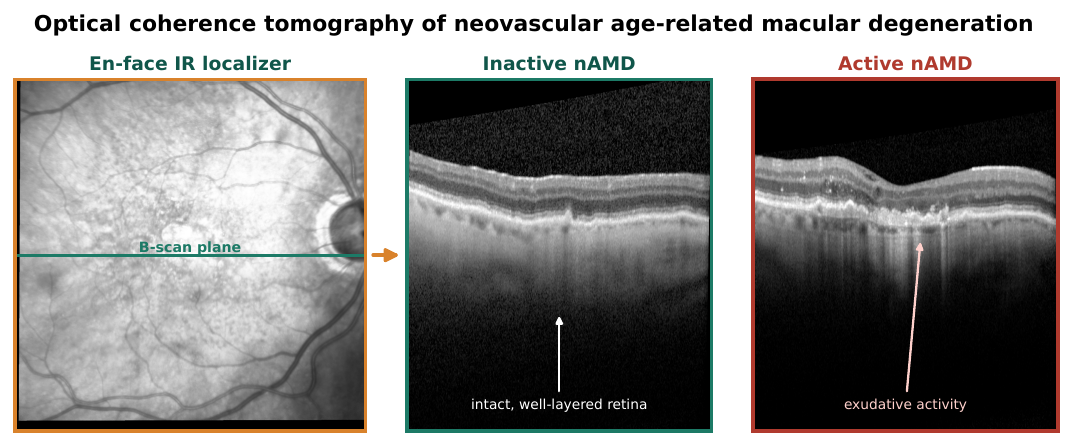}
	\caption{Optical coherence tomography (OCT) of neovascular AMD.
	\textbf{Left:} an en-face infrared localizer indicating the plane of a cross-sectional
	B-scan. \textbf{Middle:} an OCT B-scan of an inactive macula with intact, well-layered
	retina. \textbf{Right:} an OCT B-scan with active neovascular AMD, where exudative
	activity (subretinal/intraretinal fluid and hyperreflective foci) disrupts the normal
	retinal architecture. All images are from the Brest cohort.}
	\label{fig:introimageamd}
\end{figure*}

The introduction of anti-vascular endothelial growth factor (anti-VEGF) therapies in 2007 revolutionized the management of nAMD, demonstrating unprecedented efficacy in halting disease progression and, in some cases, restoring visual function \citep{Rosenfeld2006,Brown2009}. However, the success of anti-VEGF therapy is contingent upon early diagnosis, accurate disease activity assessment, and strategic treatment planning based on regular monitoring. Optical coherence tomography (OCT) has emerged as the cornerstone imaging modality in this context, providing high-resolution, three-dimensional visualization of retinal microstructure capable of detecting critical exudative markers --- subretinal fluid (SRF), intraretinal fluid (IRF), and intraretinal hyperreflective foci --- which serve as main indicators of neovascular activity (Fig.\ref{fig:introimageamd}).

While artificial intelligence (AI) algorithms have demonstrated promising capabilities in identifying early and intermediate AMD stages and in predicting progression to advanced disease \citep{Yim2020,Grassmann2018}, a significant gap exists in AI applications specifically designed for longitudinal monitoring of neovascular activity in patients undergoing anti-VEGF therapy. Current research has predominantly focused on cross-sectional disease classification rather than the dynamic assessment required to detect subtle temporal changes in exudative activity, information crucial for optimizing individualized treatment regimens. Developing algorithms able to accurately characterize disease evolution in these patients could substantially enhance treatment outcomes by enabling more precise, personalized anti-VEGF therapy protocols that respond directly to fluctuations in disease activity \citep{SchmidtErfurth2022}.

Despite the proven efficacy of anti-VEGF therapy, the associated treatment burden on patients and healthcare systems is considerable. Real-world studies report that patients with nAMD receive an average of five to seven intravitreal injections per year under standard care \citep{Lanzetta2021}, and treat-and-extend protocols may require eight or more annual injections during active disease phases \citep{Lalwani2009}. Each injection requires a clinical visit with OCT imaging, creating a substantial monitoring workload for already-stretched ophthalmic services. Automated, accurate assessment of disease activity from OCT could alleviate this burden by enabling confident deferral of injections when disease is stable, reducing unnecessary visits and freeing clinical capacity for newly presenting patients.

The MARIO (Monitoring Age-Related macular degeneration with Intelligent Ophthalmology) challenge\footnote{\url{https://youvenz.github.io/MARIO\_challenge.github.io/}} directly addresses this critical need by evaluating existing and novel AI algorithms specifically for detecting the evolution of neovascular activity in patients with exudative AMD. This initiative provides a diverse OCT dataset encompassing patients from both African and European populations, thereby facilitating assessment of domain generalizability and encouraging development of AI models capable of individualized disease monitoring across demographically diverse populations. By specifically targeting the progression dynamics of nAMD in closely monitored clinical scenarios, this challenge aims to establish new benchmarks for AI in longitudinal disease management, ultimately contributing to more precise, equitable, and effective treatment strategies.

To provide a comprehensive roadmap for the reader, this paper is structured as follows:

Sect.\ref{sec:sect2} reviews related work, contextualizing the challenge within the broader landscape of existing research and previous competitions. Sect.\ref{sec:sect3} elaborates on the key technical challenges addressed by the MARIO benchmark. Sect.\ref{sec:sect4} introduces the MARIO challenge in detail, outlining its specific objectives, clinical significance, and the particular problems it addresses. Sect.\ref{sec:sect5} describes the dataset composition and characteristics, including acquisition protocols, preprocessing methodologies, and inherent limitations. It also presents the evaluation metrics employed to assess algorithmic performance. Sect.\ref{sec:sect6} provides a detailed summary of the methodologies proposed by each participating team. Sect.\ref{sec:sect7} presents the results and outcomes of the competition. Sect.\ref{sec:sect8} offers a critical discussion of key insights, limitations, and promising directions for future research.

\section{Related work}
\label{sec:sect2}

Understanding the clinical context of AMD management is essential for appreciating the significance of MARIO.

\subsection{Anti-VEGF treatment paradigms}

Current anti-VEGF treatment strategies for nAMD follow several established paradigms, including fixed-interval dosing, pro re nata (PRN, as needed), and treat-and-extend protocols \citep{Lanzetta2021}. Each approach presents distinct advantages and limitations regarding treatment burden, visual outcomes, and healthcare resource utilization.
The fixed-interval dosing regimen, derived from pivotal clinical trials like MARINA and ANCHOR \citep{Rosenfeld2006,Brown2009}, involves regular monthly injections regardless of disease activity. While this approach provides optimal visual outcomes, it imposes substantial treatment burden on patients and healthcare systems. Conversely, the PRN approach involves treatment only when active disease is detected during monthly monitoring visits, reducing injection frequency but requiring frequent clinical assessment \citep{Lalwani2009}.
The treat-and-extend protocol, which has gained prominence in recent years, aims to optimize the treatment-monitoring balance by gradually extending the interval between treatments when disease stability is achieved, and shortening intervals when recurrent activity is detected \citep{Freund2015}. This personalized approach necessitates accurate and consistent evaluation of disease activity markers, particularly fluid presence and evolution. This is precisely the clinical need that the MARIO challenge addresses.

\subsection{Clinical significance of fluid dynamics}

The presence, location, and quantity of retinal fluid serve as primary biomarkers for nAMD activity and treatment response \citep{Schmidt-Erfurth2018}. Different fluid compartments---subretinal fluid (SRF), intraretinal fluid (IRF), and sub-retinal pigment epithelium fluid---exhibit distinct associations with visual outcomes and treatment response patterns.
Studies have demonstrated that persistent IRF strongly correlates with poorer visual outcomes and may indicate irreversible retinal damage \citep{Blumenkranz2010,Sharma2016}, while moderate SRF may be tolerated without significant visual impact in some patients \citep{Guymer2019}. This nuanced understanding of fluid dynamics has led to more sophisticated treatment decision-making, moving beyond binary dry/wet classification toward quantitative assessment of specific fluid patterns and their evolution over time.
The development of AI algorithms able to detect subtle changes in fluid distribution and volume across sequential visits therefore addresses a critical unmet need in clinical practice: the ability to objectively quantify disease activity dynamics and personalize treatment decisions based on individual response patterns rather than on standardized protocols.

\subsection{Foundation models for retinal imaging}

Foundation models---large neural networks pretrained on broad datasets and subsequently fine-tuned for downstream tasks---have recently transformed retinal image analysis. RETFound~\citep{zhou2023foundation} is a self-supervised Vision Transformer (ViT-L) pretrained on 1.6 million unlabelled retinal images using masked autoencoders; it demonstrated strong generalisation across diverse retinal classification and grading tasks and is freely available for ophthalmological research. BiomedCLIP~\citep{zhang2023biomedclip} extends CLIP-style contrastive vision--language pretraining to the biomedical domain using 15 million scientific image--text pairs, yielding state-of-the-art performance in zero-shot and fine-tuned settings across multiple medical imaging modalities. The widespread adoption of these models among MARIO finalists (Section~\ref{sec:sect6}) confirms that domain-specific pretraining provides a decisive advantage for rare or complex longitudinal classification tasks, where labelled data are inherently limited.

\subsection{Related OCT benchmarks}

Several public benchmarks have been established for retinal OCT analysis. Table~\ref{tab:challenge_comparison} summarises the most relevant challenges and their relationship to MARIO.

\begin{table*}[t]
\centering
\caption{Comparison of public retinal OCT challenge benchmarks. `Longitudinal' indicates whether the task requires temporal reasoning across visits.}
\label{tab:challenge_comparison}
\setlength{\tabcolsep}{10pt}
\renewcommand{\arraystretch}{1.4}
\arrayrulecolor{tblline}
\resizebox{\textwidth}{!}{%
\rowcolors{2}{tblrow}{white}
\begin{tabular}{lcccccc}
\toprule
\rowcolor{tblhead}
\textcolor{white}{\textbf{Challenge}} & \textcolor{white}{\textbf{Year}} & \textcolor{white}{\textbf{Task type}} & \textcolor{white}{\textbf{Modality}} & \textcolor{white}{\textbf{Longitudinal}} & \textcolor{white}{\textbf{Teams}} & \textcolor{white}{\textbf{Cross-domain}} \\
\midrule
RETOUCH~\citep{Bogunovic2019} & 2017 & Fluid segmentation & OCT (3 vendors) & No & 11 & No \\
AROI~\citep{Melinak2023}      & 2021 & Layer \& fluid segmentation & OCT (Spectralis) & No & -- & No \\
OLIVES~\citep{prabhushankarolives2022} & 2022 & Biomarker \& clinical labelling & OCT + fundus & Yes & -- & No \\
MARIO (ours) & 2024 & Activity change classification \& progression prediction & OCT + IR + clinical & Yes & 35 & Yes \\
\bottomrule
\end{tabular}%
}
\arrayrulecolor{black}
\end{table*}

RETOUCH~\citep{Bogunovic2019} established the first large-scale benchmark for retinal fluid segmentation across multiple OCT devices, covering AMD and diabetic macular oedema. It focused on cross-sectional segmentation of individual volumes, without temporal reasoning across visits. MARIO extends this scope by requiring temporal reasoning between paired visits (Task~1), providing a prognostic prediction task (Task~2), and including a cross-domain hold-out cohort to assess generalisation across populations and devices.

\section{Technical challenges in longitudinal OCT analysis}
\label{sec:sect3}

The development of robust algorithms for longitudinal OCT analysis in nAMD presents several unique technical challenges that are distinct from conventional cross-sectional classification or segmentation tasks.

\subsection{Registration and alignment}

Accurate spatial alignment between sequential OCT volumes is fundamental for reliable detection of temporal changes. Patient movement, variations in scan protocols, and alterations in retinal morphology due to disease progression or treatment effects can complicate registration \citep{Lang2016}. While commercial OCT systems incorporate basic eye-tracking functionality, suboptimal alignment remains common in clinical datasets.
Advanced registration techniques are required to establish precise spatial correspondence across sequential scans. These approaches must account for both global transformation parameters and local deformations resulting from disease activity changes. Methods ranging from feature-based registration to deformable transformation models have been explored, though their application in routine clinical analysis remains limited \citep{Golabbakhsh2013}.

\subsection{Handling irregular time intervals}

Clinical follow-up schedules in nAMD management frequently involve irregular time intervals between visits, with intervals ranging from 4 weeks to several months depending on disease activity and treatment protocol. This temporal irregularity poses significant challenges for conventional sequence modeling approaches, which typically assume uniform sampling intervals.
Recent deep learning approaches have adapted to address this challenge, including temporal convolutional networks with dilated convolutions \citep{Lea2017}, continuous-time models like neural ordinary differential equations (NODEs) \citep{Chen2018}, and attention-based architectures that can model long-range dependencies regardless of temporal distance \citep{Vaswani2017}. The MARIO challenge provides an opportunity to evaluate these approaches in a standardized clinical context.

\subsection{Balancing sensitivity and specificity}

The clinical utility of AMD monitoring algorithms depends on achieving an optimal balance between sensitivity (detecting all instances of disease activity) and specificity (avoiding false positive detections that could lead to unnecessary treatment). This balance is particularly crucial in treatment decision-making, where false negatives could result in undertreatment and disease progression, while false positives might lead to overtreatment and increased iatrogenic risks \citep{Liu_2018Hanruo}.
Conventional evaluation metrics like accuracy may be insufficient or even misleading in this context, particularly when dealing with class imbalance or when considering the differential clinical impact of various error types. The MARIO challenge therefore incorporates clinically relevant evaluation metrics that reflect the actual utility of algorithms in treatment decision support.

Research in AI-assisted AMD analysis using OCT imaging has evolved substantially in recent years, focusing on several interconnected domains \citep{Crincoli2024,Romond2021,Muntean2023,SchmidtErfurth2016}: automated retinal layer segmentation, detection and quantification of exudative features, longitudinal analysis of disease progression, and cross-domain generalizability. These research directions collectively provide the foundation for comprehensive understanding of AMD pathophysiology and evidence-based clinical decision-making. We review these key areas with particular emphasis on their relevance to monitoring disease progression in patients undergoing anti-VEGF treatments.

\subsection{Exudative feature detection and quantification}

The detection and precise quantification of exudative markers --- including SRF, IRF, and hyperreflective foci - represents a critical component in evaluating neovascular AMD activity, as these features directly inform anti-VEGF treatment decisions and retreatment intervals. Early approaches relied on traditional computer vision techniques with handcrafted feature extraction, but deep learning architectures have since transformed the landscape of exudative feature detection \citep{hassan_deep_2021,Schlegl2018}.
Convolutional neural networks, particularly implementations of mask R-CNN, U-Net derivatives, and DenseNet-based architectures, have significantly improved accuracy in identifying and segmenting fluid compartments and hyperreflective material \citep{Venhuizen2020}. However, these models continue to face challenges with variations in image quality, scan protocols, and patient-specific anatomical differences. Recent advancements have explored specialized architectural adaptations, including attention mechanisms \citep{Xu2023}, Transformer-based models \citep{Melinak2023}, and hybrid approaches that integrate anatomical priors with data-driven learning \citep{Roy2017}.

While these innovations have improved performance in controlled settings, the reliable detection of subtle fluid changes over time remains challenging, particularly in patients undergoing active treatment. The MARIO challenge format, with its focus on patients receiving ongoing anti-VEGF therapy, provides an ideal testbed for evaluating models capable of sensitive and specific fluid detection across sequential visits. This capability is crucial for longitudinal disease monitoring and capturing the temporal evolution of exudative markers, which directly impacts treatment decision-making \citep{SchmidtErfurth2022,Simader2014}.

\subsection{Longitudinal analysis and progression modeling}

Recent advances in longitudinal OCT analysis have enabled more sophisticated approaches to understanding and predicting AMD progression. Traditional methods relied primarily on statistical modeling of discrete time points, but contemporary approaches increasingly leverage the temporal continuity, inherent in disease progression.

Self-supervised learning techniques have demonstrated promise in modeling disease trajectories and predicting the onset of advanced AMD stages without requiring extensive labeled data \citep{rivail2019modeling,Jung2024}. These approaches enable the extraction of meaningful representations from temporal sequences of OCT volumes, capturing subtle changes that might escape human detection. Recurrent neural networks, particularly LSTM and GRU architectures, have been applied to model sequential dependencies in OCT data. However, they often struggle with irregular sampling intervals common in clinical practice \citep{Lad2022}.

More recently, differential equation-based models, especially NODEs, have emerged as powerful frameworks for modeling continuous-time disease dynamics in various conditions including AMD \citep{Yellapragada2022,Chakravarty2024}. These approaches offer distinct advantages in handling irregular sampling intervals and providing interpretable trajectories of disease states, allowing for more natural integration of multimodal data including imaging biomarkers and clinical factors.

Despite these advances, existing longitudinal models typically focus on predicting conversion to advanced disease rather than characterizing fluctuations in disease activity within patients already diagnosed with nAMD. The MARIO challenge addresses this gap by specifically targeting the detection of disease activity changes in patients undergoing treatment, representing a more clinically relevant scenario for therapeutic decision-making.

\subsection{Domain adaptation and generalizability}

Cross-domain generalizability remains one of the most significant challenges in clinical OCT analysis for AMD \citep{kugelman2022review,MATTA2024109256}. Substantial variability in OCT data can arise from differences in scanner types, acquisition protocols, image quality, and patient demographics, often resulting in significant performance degradation when algorithms are applied outside their training domain.
Various domain adaptation techniques have been proposed to address these challenges, including adversarial learning approaches \citep{guan2021domain}, transfer learning strategies \citep{liu2022deep}, and generative models such as CycleGANs for style transfer between domains \citep{Gomariz2022}. These methods aim to reduce the domain shift between source and target distributions, thereby improving model robustness across heterogeneous data sources.
However, the fundamental limitation of restricted data diversity persists in most current research, with algorithms developed and validated predominantly on homogeneous populations showing diminished performance when applied to underrepresented groups. 

Addressing algorithmic bias is therefore essential for ensuring equitable performance across diverse populations \citep{ueda2024fairness,LIM2024100096}. The MARIO challenge uniquely addresses these limitations by providing a dataset inclusive of both African and European patient data, promoting the development of adaptable models suitable for global deployment in diverse clinical settings.

\subsection{Publicly-available datasets and benchmarking efforts}

Public datasets and standardized benchmarks have been instrumental in advancing AMD research, particularly in the domains of fluid detection and retinal layer segmentation. Notable contributions include the RETOUCH challenge dataset \citep{Bogunovic2019}, which features OCT images from multiple devices and various retinal diseases, establishing important benchmarks for fluid detection and segmentation tasks. Similarly, the OCTAGON dataset \citep{Daz2019} provides valuable resources for algorithm development and validation in retinal OCT analysis. More recently, the OLIVES dataset \citep{prabhushankarolives2022} paired OCT and fundus images with clinical labels and biomarkers across repeated visits, offering a richer multimodal resource for retinal disease analysis. However, these existing datasets have limitations that restrict their utility for developing comprehensive AMD monitoring solutions. Most critically, they often lack:

\begin{itemize}
    \item longitudinal data with sufficient temporal resolution to capture disease dynamics during treatment,
    \item geographic and demographic diversity needed to evaluate algorithm generalizability,
    \item standardized annotations of subtle changes in disease activity relevant to treatment decisions,
    \item integration of clinical metadata with imaging findings to provide contextual information.
\end{itemize}

The MARIO challenge addresses these limitations by establishing a unique benchmarking platform with several distinguishing features. First, it offers a dataset specifically designed to evaluate algorithms for monitoring disease progression in treated patients rather than simple binary classification or segmentation tasks. Second, it promotes equitable AI development by accounting for demographic variability across different geographic regions. Finally, it provides a standardized framework for evaluating algorithmic performance on clinically relevant outcomes related to disease activity changes.
This initiative represents the first MICCAI-affiliated challenge specifically focused on AMD progression monitoring, providing a valuable resource for the research community. By offering participants the opportunity to test their methods on both European and African datasets, the challenge enables rigorous evaluation of algorithmic performance across shifted populations with respect to both ethnicity and OCT acquisition devices. This is a critical step toward developing truly generalizable solutions for global application.

\section{Challenge description}
\label{sec:sect4}

\begin{figure*}[h!]
    \centering
    \includegraphics[width=2\columnwidth]{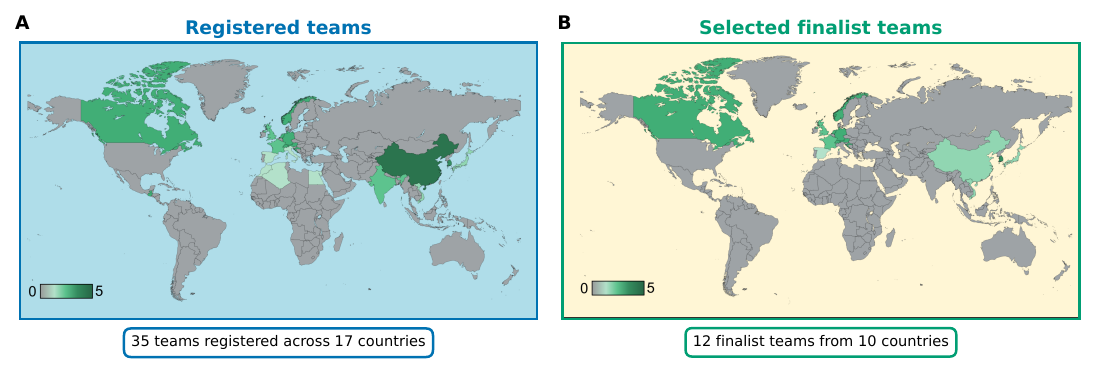}
    \caption{Geographic distribution of MARIO participants. \textbf{A:} the 35 teams
    registered to the challenge, spanning 17 countries. \textbf{B:} the 12 finalist teams
    that reached the final phase, from 10 countries. Colour encodes the number of teams per
    country (grey = none).}
    \label{fig:dist_world_mario}
\end{figure*}

\begin{figure*}[h!]
    \centering
    \includegraphics[width=1\linewidth]{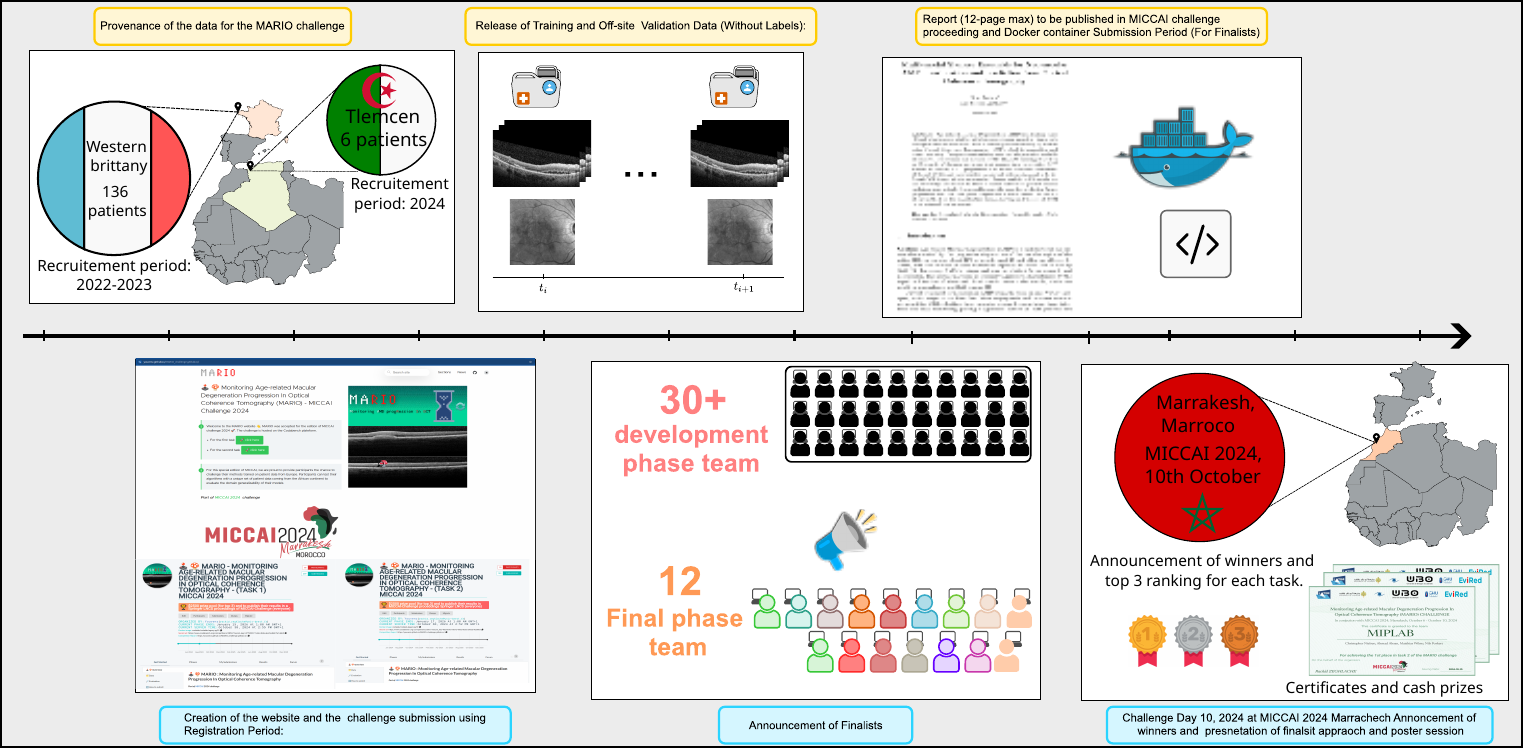}
    \caption{Time line of the challenge since inception.}
    \label{fig:timeline}
\end{figure*}

The MARIO challenge is designed following the BIAS
Reporting Guideline \cite{MaierHein2018} for enhanced quality and
transparency of biomedical research. The proposal was approved after two rounds
of MICCAI review followed by a call for participation circulated online
and offline. The challenge was officially launched on April 1, 2024, and run
through a 6-month window as shown in Fig.~\ref{fig:timeline}, using the Codabench platform for hosting \cite{codabench}. Participants were supported throughout with training data, a metrics library, a Slack channel, starter code, and GitHub repositories. Fig.~\ref{fig:dist_world_mario} shows the number of registered teams and their countries, alongside those of the finalists. The timeline also included a validation phase (self-validation system, validation data, and a Docker submission template), concluding with the presentation of methods, results, and awards at the MICCAI 2024 conference in Marrakesh on October 10, 2024.

\subsection{Challenge Format}

The challenge comprised three phases:
\begin{itemize}
    \item \textbf{Training Phase}: Release of training data for algorithm development.
    \item \textbf{Off-Site Validation}: Submission of results on validation data to determine finalists.
    \item \textbf{Final Round}: Top teams submitted Docker containers for on-site evaluation.
\end{itemize}

Participants addressed two core tasks:

\subsection{Proposed tasks }

\begin{figure*}[t]
    \centering
    \includegraphics[width=\textwidth]{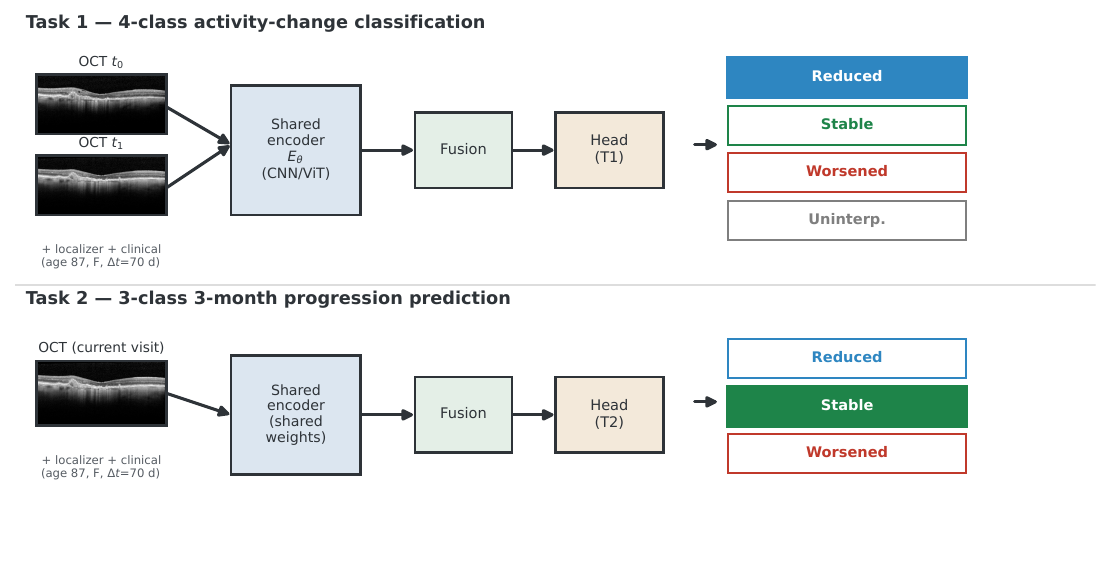}
    \caption{Overview of the MARIO tasks and the canonical deep-learning pipeline used by
    the finalist teams, shown on a representative Brest example. Inputs (OCT B-scan(s), the
    paired infrared localizer, and clinical variables) are embedded by a shared encoder
    ($E_\theta$; a CNN or Vision Transformer with shared weights), the features are fused
    (concatenation or difference), and a classification head with a softmax output produces
    the class probabilities. \textbf{Task~1} (top) classifies the change in neovascular
    activity between two consecutive B-scans ($t_0, t_1$) into \emph{four} classes---reduced,
    stable, worsened, or uninterpretable. \textbf{Task~2} (bottom) predicts the activity
    state three months after a single visit into \emph{three} classes---reduced, stable, or
    worsened.}
    \label{fig:tasks}
\end{figure*}

The challenge comprised two complementary tasks, illustrated together in
Figure~\ref{fig:tasks}. \textbf{Task~1} was the classification of the evolution of
neovascular activity between two consecutive 2D OCT B-scans (baseline $t_0$ versus
follow-up $t_1$), into reduced, stable, or worsened activity.
\textbf{Task~2} was the prediction of future AMD evolution over the following three
months (a single-visit prognostic task) for patients undergoing anti-VEGF therapy.

All submissions were evaluated based on F1-score, specificity, and Rk-correlation coefficients for both tasks, with an additional Quadratic Weighted Kappa metric for Task 2. Members of the organizer's institutes could participate but were not eligible for awards. Regarding the publication policy, up to 5 members of the individual top teams (ranks 1--6) according to the leaderboard were invited to contribute to this joint challenge paper as co-authors, and up to 3 members for the top 7--12. The participating teams may publish their own results separately with citations to the assigned papers on the online challenge platform after the challenge paper is published. To access the training and testing datasets, participants first register on the challenge website and sign a non-disclosure contract on the usage of the datasets. Afterward, participants are provided with a download link to the online repository containing the dataset. For submission, the participants could perform cross validation on the training set. The evaluation code \footnote{\url{https://github.com/YouvenZ/MARIO-Challenge-MICCAI-2024}} was made available prior to submission. The participating teams were encouraged to make their source code publicly available.

\section{Data description}
\label{sec:sect5}

Imaging data were collected with a Heidelberg Spectralis OCT device and include high-resolution 2D B-scan and 3D volume sequences from 136 patients enrolled at the University Hospital of Brest, France. Each patient's longitudinal record contains up to 10 consecutive OCT C-scans per eye, each paired with an infrared (IR) localizer image. The Brest cohort is partitioned into 68 training, 34 validation, and 34 test patients; an additional 5 patients from the University Hospital of Tlemcen, Algeria, form a separate cross-domain test set to assess generalisation across populations and imaging conditions. Table~\ref{tab:data_split} summarises the dataset splits and annotation protocols.

\begin{table}[h!]
\centering
\caption{MARIO dataset splits and annotation protocols.}
\label{tab:data_split}
\setlength{\tabcolsep}{4pt}
\renewcommand{\arraystretch}{1.25}
\arrayrulecolor{tblline}
\resizebox{\columnwidth}{!}{%
\rowcolors{2}{tblrow}{white}
\begin{tabular}{lccl}
\toprule
\rowcolor{tblhead}
\textcolor{white}{\textbf{Split}} & \textcolor{white}{\textbf{Patients}} & \textcolor{white}{\textbf{Annotators}} & \textcolor{white}{\textbf{Annotation protocol}} \\
\midrule
Training (Brest)      & 68 & 1 ophthalmologist  & Single \\
Validation (Brest)    & 34 & 1 ophthalmologist  & Single \\
Test (Brest)          & 34 & 2 ophthalmologists & Independent, adjudicated \\
Test (Tlemcen)        &  5 & 2 ophthalmologists & Independent, adjudicated \\
\bottomrule
\end{tabular}%
}
\arrayrulecolor{black}
\end{table}

\begin{figure*}
    \centering
    \includegraphics[width=1.9\columnwidth]{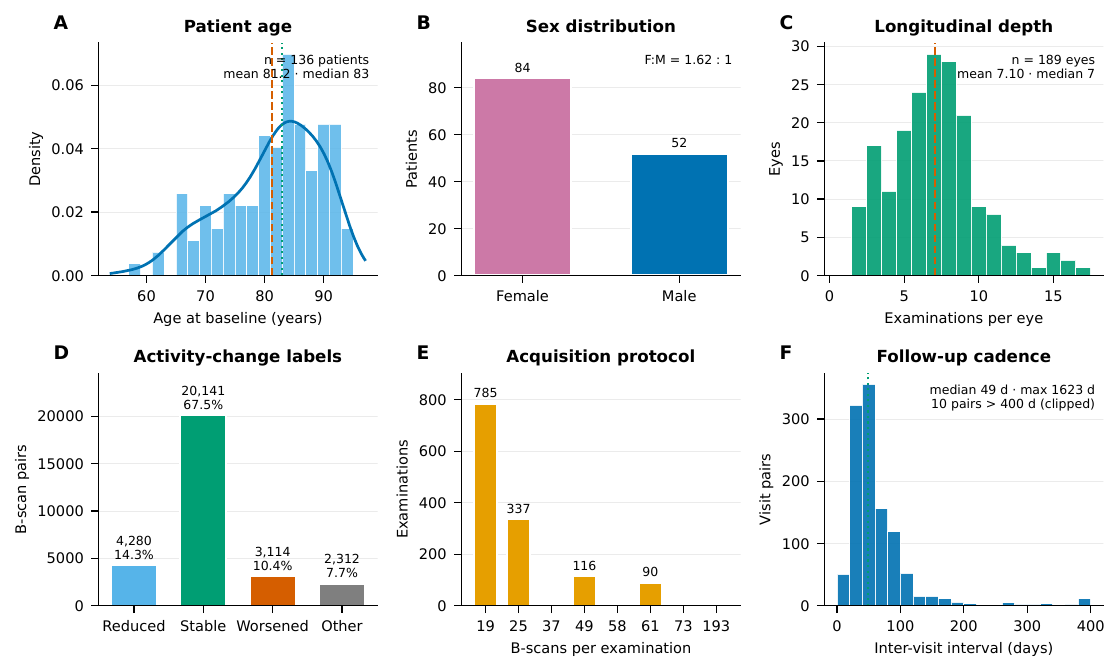}
    \caption{Exploratory analysis of the Brest MARIO cohort (136 patients, 189 eyes).
    \textbf{A:} distribution of patient age at baseline (histogram with kernel-density
    estimate; dashed line = mean, dotted line = median).
    \textbf{B:} sex distribution with female-to-male ratio.
    \textbf{C:} number of examinations per eye, quantifying longitudinal depth.
    \textbf{D:} distribution of the simplified three-class activity-change labels
    (Reduced, Stable, Worsened) plus the excluded \textit{Other} category, over all B-scan pairs.
    \textbf{E:} number of B-scans acquired per examination, reflecting two acquisition protocols.
    \textbf{F:} inter-visit time interval; intervals beyond 400~days are clipped into the final
    bin (median 49~days, maximum 1{,}623~days).}
    \label{fig:mario_eda}
\end{figure*}

The MARIO database, comprises a comprehensive longitudinal collection of ophthalmic data, facilitating detailed analysis of neovascular age-related macular degeneration (nAMD). As illustrated in Figure \ref{fig:mario_eda}, the dataset's composition and characteristics provide a robust foundation for investigating disease progression and treatment efficacy. The gender distribution reveals a notable skew towards female patients: across the full 136-patient Brest database (training, validation, and test splits combined), 84 participants are female and 52 are male (61.8\,\%), suggesting a potential demographic bias in the patient population or a higher prevalence of nAMD among females \cite{Rudnicka2012}. This skew is more pronounced within the 34-patient held-out test split specifically (76\,\%\ female; Section~\ref{sec:overall_rankings}).

The imaging frequency followed standardized protocols for most patients, with 785 examinations performed using 19 B-scans and 337 examinations using 25 B-scans. The observed variation in B-scan count likely reflects differences in clinical assessments or adjustments to protocols tailored to individual patient needs. In terms of follow-up timing, the majority of patients demonstrated short inter-visit intervals, suggesting consistent and diligent monitoring. Nevertheless, the presence of outliers, with inter-visit periods extending to 1600 days, points to instances of irregular follow-up, possibly due to patient compliance issues or clinical scheduling limitations.

The age distribution at baseline demonstrates a concentration within the elderly population, a demographic typically associated with nAMD. The mean age at baseline was 81.22 years, with a median of 83 years, and the majority of patients falling within the 70-90 year age range. This age profile aligns with the known prevalence of nAMD in older adults. The follow-up duration, calculated from the initial visit to the baseline, averaged 379.10 days, with a median of 371 days, signifying a substantial monitoring period. This longitudinal depth is essential for analyzing disease trajectories and evaluating long-term treatment outcomes.

Finally, the frequency of patient visits per eye, with a mean of 7.10 visits and a typical range of 4 to 10 visits, underscores the intensive clinical management required for nAMD. This frequent monitoring regimen is indicative of the chronic and progressive nature of the disease, necessitating consistent evaluation and intervention. Collectively, these descriptive statistics provide a comprehensive overview of the MARIO database, highlighting its suitability for detailed investigation into the clinical dynamics of nAMD.

\subsection{Data Pre-Processing}
The XML metadata files were de-identified, and BMP images were converted to PNG format. Consecutive 3D volumes were registered using the device follow-up mode to account for patient movement. Inter-annotator agreement was established as a baseline for algorithm evaluation, given the variability in distinguishing between stable and reduced activity.

\subsection{Manual annotation}

For test cases, two ophthalmologists (A.B, T.M) performed the annotation independently. For training and validation cases, one ophthalmologist performed the annotation. An online annotation tool was designed for the study. Consecutive C-scans were viewed jointly on the same screen (older examination at the top, newer examination at the bottom). With a slider, the annotator could browse through pairs of matched B-scans from the two consecutive C-scans. Moreover, 5 patients (from the training set) were used to train the annotators. After annotating these 5 patients: 1) annotations were compared by the challenge organizers and presented to the annotators, 2) the organizers discussed their annotation strategy with each other and 3) they were given the opportunity to revise their annotations. Next, they annotated the remaining 131 patients independently.

\begin{table}[h!]
\centering
\resizebox{\columnwidth}{!}{%
\begin{tabular}{l S[table-format=2.0] S[table-format=5.0]}
\toprule
\textbf{Name Class} & \textbf{Class Label} & \textbf{Number of Pairs} \\
\midrule
\rowcolor{gray!20} \textsc{Uninterpretable}          & -1 & 2303  \\
\rowcolor{green!10} \textsc{Inactive}                & 0  & 18852 \\
\rowcolor{orange!20} \textsc{Eliminated}              & 1  & 2170  \\
\rowcolor{orange!20} \textsc{Persistent Reduced}      & 2  & 2110  \\
\rowcolor{yellow!20} \textsc{Persistent Stable}       & 3  & 1289  \\
\rowcolor{red!20} \textsc{Persistent Worsened}     & 4  & 1238  \\
\rowcolor{red!40} \textsc{Appeared and Eliminated} & 5  & 9     \\
\rowcolor{orange!20} \textsc{Appeared}                & 6  & 1876  \\
\bottomrule
\end{tabular}%
}
\caption{Data distribution of the manual annotation.}
\label{table:label_dist}
\end{table}

\noindent For each pair of matched B-scans, the annotator assigned one of the following 7 labels presented in Tab~\ref{table:label_dist} along with their corresponding number of pairs.

\subsection{Simplified classification}

\begin{table}[h!]
\centering
\begin{tabular}{l S[table-format=2.0] S[table-format=5.0]}
\toprule
\textbf{Name Class} & \textbf{Class Label} & \textbf{Number of Pairs} \\
\midrule
\rowcolor{orange!20} Reduced  & 0 & 4280 \\
\rowcolor{green!10} Stable   & 1 & 20141 \\
\rowcolor{red!20} Worsened & 2 & 3114 \\
\rowcolor{gray!20} Other    & -1 & 2312 \\
\bottomrule
\end{tabular}
\caption{Data distribution of the simplified annotation}
\label{table:label_dist_simplflied_anno}
\end{table}

The task proposed in this challenge focuses on pairs of 2-D slices (B-scans) from two consecutive OCT acquisitions. The goal is to classify evolution between these two slices (`before' and `after'), which clinicians typically look at side by side on their screen. For the evolution assessment, three classes are defined based on the image level annotation presented in Tab~\ref{table:label_dist_simplflied_anno} along with their corresponding number of pairs.

\begin{enumerate}
        \item Reduced (class 0)
        \begin{itemize}

        \item Contain labels:  eliminated (1) or persistent reduced (2)
        \end{itemize}
        \item  Stable (class 1) 
        \begin{itemize}

        \item Contain labels:  inactive (0) or persistent stable (3)
\end{itemize}
        \item  Worsened (class 2) 
   \begin{itemize}
     
\item Contain labels: persistent worsened (4) or appeared (6)
\end{itemize}
    \end{enumerate}

\subsection{Justification for dataset size}

Data export was manual and, therefore, time-consuming. In addition, manually annotating thousands of image pairs by two retinal experts was also time-consuming. This was the reason for not collecting a larger dataset. However, this remains larger than a similar study published by one of the organizers \cite{Quellec2019}, which involved 70 patients, as opposed to 136 here.

\subsection{Sources of errors from data and annotation}

We observed two main sources of inter-annotator variability:

\begin{enumerate}
    \item The distinction between an absence of disease activity and a stable activity (two types of non-evolution).
    \item The distinction between an eliminated activity and a reduced (but not fully eliminated) activity. Experts disagreed for about 25\% of the B-scan pairs, corresponding to an approximate pairwise agreement rate of 75\% on the simplified three-class task for the Brest test set.
\end{enumerate}

\noindent Another source of error arose during image acquisition. The operator sometimes forgot to activate the follow-up mode, resulting in non-registered OCT volumes. This occurred in about 10\% of consecutive acquisitions.

\subsection{Evaluation Metrics}
The evaluation of algorithms in Task 1 and Task 2 relies on the following metrics:

\begin{itemize}
\item \textbf{F1-score}: Evaluates classification accuracy by balancing precision and recall, which is particularly relevant for handling class imbalances \citep{Sokolova2009}.
\item \textbf{Specificity}: Measures the ability to correctly identify non-progression cases, ensuring a reliable detection of negative instances \citep{Sokolova2009}. Sensitivity is not emphasized here as the focus is on minimizing false positives rather than maximizing true positives.
\item \textbf{Rank Correlation Coefficient} ($R_k$): The multi-class generalisation of the Matthews correlation coefficient \citep{Gorodkin2004}, assessing the agreement between algorithmic predictions and human grading, which is crucial for ordinal classification tasks.
\item \textbf{Quadratic Weighted Kappa} \citep{Cohen1968} (Only for Task 2): Used to evaluate prediction accuracy for disease progression. This metric is relevant in Task 2 because it accounts for the ordinal nature of disease stages, penalizing larger misclassifications more heavily. However, it is not used in Task 1 since that task involves ordinal categories where alternative correlation-based metrics, such as the rank correlation coefficient, are more appropriate for assessing agreement.
\end{itemize}

\section{Proposed solutions}
\label{sec:sect6}

A total of 35 teams participated, with 12 teams selected for the final phase as described in Fig.\ref{fig:dist_world_mario}. To be selected as a finalist, teams were required to submit at least 5 unique submissions for both tasks and achieve a score higher than the provided baseline in the development phase leaderboard.

\begin{table*}[htbp]
\centering
\scriptsize
\setlength{\tabcolsep}{4pt}
\renewcommand{\arraystretch}{1.3}
\arrayrulecolor{tblline}
\resizebox{\textwidth}{!}{%
\rowcolors{3}{tblrow}{white}
\begin{tabular}{>{\raggedright\arraybackslash}p{0.9cm} p{1.6cm} p{1.7cm} p{1.1cm} p{1.5cm} p{2.1cm} p{1.0cm} p{1.2cm} p{1.1cm} p{1.7cm} p{1.9cm}}
\toprule
\rowcolor{tblhead}
\textcolor{white}{\textbf{Team}} & \textcolor{white}{\textbf{Preprocessing}} & \textcolor{white}{\textbf{Backbone}} & \textcolor{white}{\textbf{Loss}} & \textcolor{white}{\textbf{Post-proc.}} & \textcolor{white}{\textbf{Data augmentation}} & \multicolumn{4}{c}{\textcolor{white}{\textbf{Use of}}} & \textcolor{white}{\textbf{GitHub}} \\
\rowcolor{tblsub}
 & & & & & & \textcolor{white}{\textbf{Pretext}} & \textcolor{white}{\textbf{Found.\ model}} & \textcolor{white}{\textbf{Multi-mod.}} & \textcolor{white}{\textbf{Public data}} & \\
\midrule
\textbf{Lumine} & Normalization, resize & Siamese ConvNeXt\_large & Weighted CE & No & ColorJitter, GaussianBlur, RandomAffine, HorizontalFlip, GaussianNoise & N & Y / ConvNeXt\_large & Y / multiple Siamese & ImageNet (timm) & \url{https://github.com/lumine-1/MARIO_Project} \\
\textbf{yyama} & concatenation, resize, normalization & MaxViT Tiny & CE & Ensembling (5-fold CV) & Random resized crop, horizontal flip, rotation, Coarse dropout & N & N & N & ImageNet1k (timm) & \url{https://github.com/yamagishi0824/MARIO24-MaxVit-Fused} \\
\textbf{DF41} & OCTIP\footnote{\url{https://github.com/leto-atreides-2/octip}} & ResNet50 & CE & Ensembling (average) & RandomHorizontalFlip, RandomVerticalFlip, RandomRotation, ColorJitter, RandomPerspective, GaussianBlur & N & N & N & N & \url{https://github.com/pzhangwj/mario_challenge_code} \\
\textbf{Optima} & crop, resize, normalization & RETFound (Large ViT) & Cross-entropy & Ensembling & rotation, horizontal flipping & Y / change detection (Kermany) & Y / RETFound & N & Y / Kermany & \url{https://github.com/EmreTaha/Siamese-EMD-for-AMD-Change} \\
\textbf{TONIC} & Resampled to 224$\times$224 & ResNet18 & CE & -- & Random horizontal/vertical flips, rotations, brightness/contrast, resized cropping & N & N & Y / two ResNet18's & ImageNet & \url{https://github.com/ninamalou/TONIC-MICCAI-2024} \\
\textbf{jkulinzstudents} & crop, resize, normalization & ViT-B/14 distilled & CE & Ensembling (patient-ID split folds) & Resize to 224$\times$224, normalization & N & Y / ViT-B/14 distilled & N & Y / from foundation-model pretraining & \url{https://github.com/marceljhuber/mario-miccai2024} \\
\textbf{FERLIV} & resize, crop, normalization & ViT-Large & weighted CCE & -- & random resized crop, horizontal flip & Y / Segmenter with linear decoder & Y / OCT RETFound & N & Y / OCT MS \& Healthy Controls & \url{https://github.com/LovreAB17/FERLIV-MARIO} \\
\textbf{MIPLAB} & Resize, intensity normalization & RETFound (ViT-large) + EfficientNetV2 & CE + hinge & Pseudo-labelling & Horizontal/vertical flips, rotation, translation, contrast & Y / MAE (via RETFound) & Y / RETFound & Y / OCT + localizer + clinical & N & \url{https://github.com/chrisnielsen/miccai-2024-mario-challenge} \\
\textbf{MIC group 6} & z-score norm., CenterCrop 224$\times$224 & ViT & CE & -- & ResizedCrop, Rotation, HorizontalFlip, Affine, GaussianBlur, GaussianNoise & N & Y / BiomedCLIP & Y / BiomedCLIP & N & \url{https://github.com/MIC-DKFZ/mario} \\
\textbf{STEP} & 3D volume creation, resize, normalization & Vision Transformer & Focal Loss & -- & Intensity scaling, Gaussian noise, horizontal/vertical flip & -- & Y / RETFound & N & N & \url{https://github.com/BIT-UPM/mario_miccai_24_step_amd} \\
\textbf{scyyd4} & crop, resize, normalization & ConvNeXt V2-Large & CE & Ensembling & Rotation, Zoom & N & Y / ConvNeXt V2-Large & N & N & \url{https://github.com/YIDING4869/MARIO2024} \\
\textbf{Cemrg} & Resize, normalization & MobileNetV3 & CE & Ensembling + TTA & Random rotations, flips, brightness, contrast & N & Y / MobileNetV3 & N & N & \url{https://github.com/RespectKnowledge/MARIO_DL_solution/tree/main} \\
\bottomrule
\end{tabular}%
}
\arrayrulecolor{black}
\caption{Summary of the best method submitted by each finalist team for Task~1. ``Use of'' columns indicate whether a pretext task, foundation model, multi-modal input, or public pre-training dataset was employed (with the specific component).}
\label{tab:task1summary}
\end{table*}

\begin{table*}[htbp]
\centering
\scriptsize
\setlength{\tabcolsep}{4pt}
\renewcommand{\arraystretch}{1.3}
\arrayrulecolor{tblline}
\resizebox{\textwidth}{!}{%
\rowcolors{3}{tblrow}{white}
\begin{tabular}{>{\raggedright\arraybackslash}p{0.9cm} p{1.6cm} p{1.8cm} p{1.4cm} p{1.4cm} p{2.1cm} p{1.3cm} p{1.2cm} p{1.1cm} p{1.7cm} p{1.9cm}}
\toprule
\rowcolor{tblhead}
\textcolor{white}{\textbf{Team}} & \textcolor{white}{\textbf{Preprocessing}} & \textcolor{white}{\textbf{Backbone}} & \textcolor{white}{\textbf{Loss}} & \textcolor{white}{\textbf{Post-proc.}} & \textcolor{white}{\textbf{Data augmentation}} & \multicolumn{4}{c}{\textcolor{white}{\textbf{Use of}}} & \textcolor{white}{\textbf{GitHub}} \\
\rowcolor{tblsub}
 & & & & & & \textcolor{white}{\textbf{Pretext}} & \textcolor{white}{\textbf{Found.\ model}} & \textcolor{white}{\textbf{Multi-mod.}} & \textcolor{white}{\textbf{Public data}} & \\
\midrule
\textbf{Lumine} & Normalization, resize & ConvNeXt\_large & Weighted CE & Threshold-based label adjustment by localizer & ColorJitter, GaussianBlur, RandomAffine, HorizontalFlip, GaussianNoise & N & Y / ConvNeXt\_large & N & Y / ImageNet & \url{https://github.com/lumine-1/MARIO_Project} \\
\textbf{yyama} & concatenation, resize, normalization & EfficientNet V2 & CE & -- & Random resized crop, horizontal flip, rotation, Coarse dropout & N & N & Y / patient metadata & ImageNet1k + 22K & \url{https://github.com/yamagishi0824/MARIO24-MaxVit-Fused} \\
\textbf{DF41} & OCTIP & ResNet50, ViT-Large-MAE (PPMAE) & CE & Ensembling (average) & RandomHorizontalFlip, RandomVerticalFlip, RandomRotation, ColorJitter, RandomPerspective, GaussianBlur & N & N & N & Y / PPMAE & \url{https://github.com/pzhangwj/mario_challenge_code} \\
\textbf{Optima} & crop, resize, normalization & ViT-16 & Focal + Wasserstein-2 & Majority voting & rotation, horizontal flipping & Y / MAE & N & N & N & \url{https://github.com/EmreTaha/Siamese-EMD-for-AMD-Change} \\
\textbf{TONIC} & Resampled to 224$\times$224 & ResNet18 & CE & Majority voting & Random horizontal/vertical flips, rotations, brightness/contrast, resized cropping & N & N & N & ImageNet & \url{https://github.com/ninamalou/TONIC-MICCAI-2024} \\
\textbf{jkulinzstudents} & Resize to 224$\times$224, normalization & RETFound & Neg.\ cosine sim.\ with predicted Task~1 embeddings & -- & Task~1 data, class 3 relabelled to class 1 & Y / latent matching on Task~1 embeddings & Y & N & Y / foundation-model pretraining + Task~1 data & \url{https://github.com/marceljhuber/mario-miccai2024} \\
\textbf{FERLIV} & resize, crop, normalization & ViT-Large & weighted CCE & -- & random resized crop, horizontal flip & Y / Task~1 & Y / OCT RETFound & N & Y / MARIO Task~1 & \url{https://github.com/LovreAB17/FERLIV-MARIO} \\
\textbf{MIPLAB} & Resize, intensity normalization & RETFound (ViT-large) + EfficientNetV2 & CE + ordinal logistic & None & Horizontal/vertical flips, rotation, translation, contrast & Y / MAE (via RETFound) & Y / RETFound & Y / OCT + localizer + clinical & N & \url{https://github.com/chrisnielsen/miccai-2024-mario-challenge} \\
\textbf{MIC group 6} & z-score norm., CenterCrop 224$\times$224 & ResNet50 & CE & -- & ResizedCrop, Rotation, HorizontalFlip, Affine, GaussianBlur, GaussianNoise & N & Y / ImageNet & Y / ImageNet & N & \url{https://github.com/MIC-DKFZ/mario} \\
\textbf{STEP} & 3D volume creation, resize, normalization & Vision Transformer & Focal Loss & -- & Intensity scaling, Gaussian noise, horizontal/vertical flip & N & Y / RETFound & N & N & \url{https://github.com/BIT-UPM/mario_miccai_24_step_amd} \\
\textbf{scyyd4} & resize, normalization & ConvNeXt V2-Large (frozen) + CLAM-SB & CE & Majority voting & -- & N & Y / ConvNeXt V2-Large, CLAM-SB & N & N & \url{https://github.com/YIDING4869/MARIO2024} \\
\textbf{Cemrg} & Resize, normalization & MobileNetV3 & CE & Ensembling + TTA & Random rotations, flips, brightness, contrast & N & Y / MobileNetV3 & N & N & \url{https://github.com/RespectKnowledge/MARIO_DL_solution/tree/main} \\
\bottomrule
\end{tabular}%
}
\arrayrulecolor{black}
\caption{Summary of the best method submitted by each finalist team for Task~2. Several teams transferred components or data from their Task~1 solution (e.g.\ FERLIV, jkulinzstudents, MIPLAB).}
\label{tab:task2summary}
\end{table*}

\subsection{Summary of methods proposed for each team}

\begin{figure*}[h!]
	\centering
	\includegraphics[width=\linewidth]{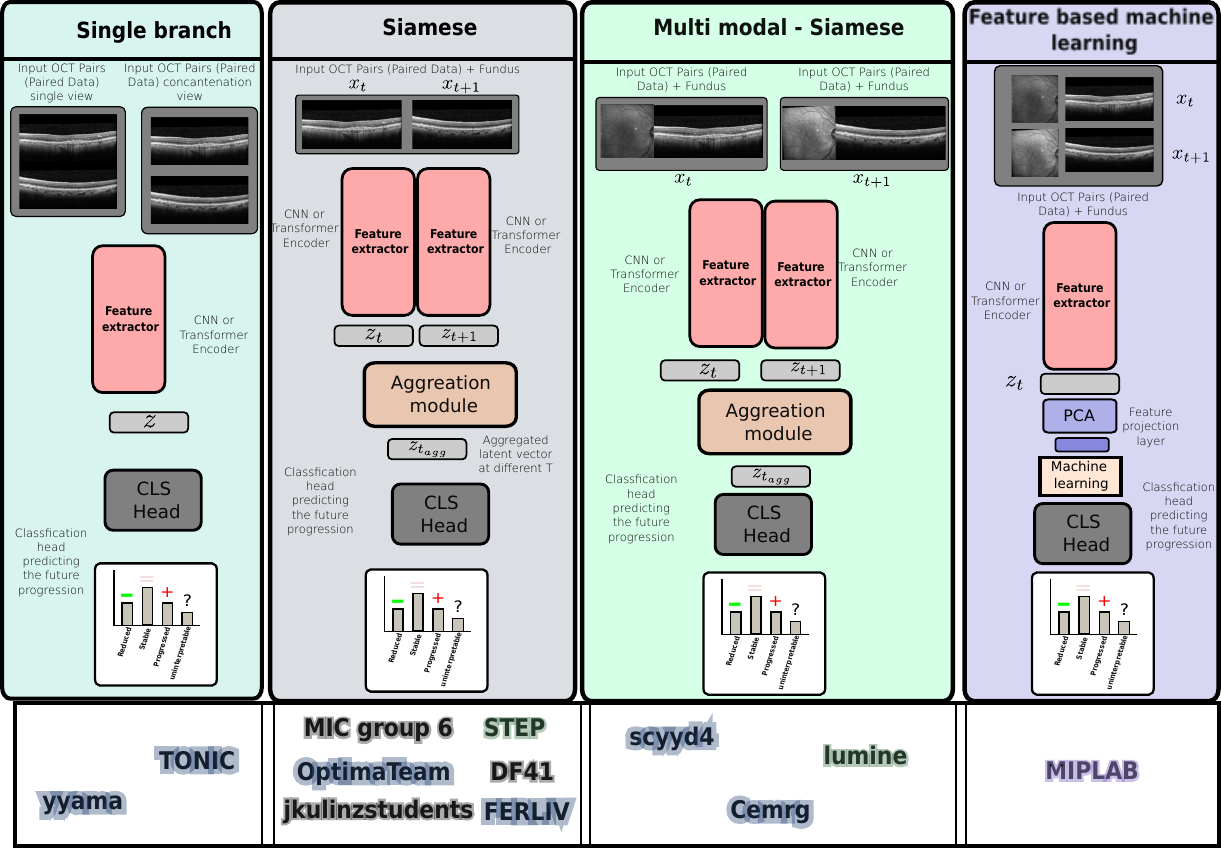}
	\caption{Overview of the methodologies proposed by all teams for Task 1.}
	\label{fig:description_method_task1}
\end{figure*}

The following section provides a concise, cross-team overview of the methodologies developed by the twelve finalists; full per-team narrative descriptions are provided in Supplementary Sections~\nameref{sec:supp_methods_t1} and~\nameref{sec:supp_methods_t2} to keep the main text focused. A visual summary illustrating the common pipeline structures employed by the participants is presented in Figures~\ref{fig:description_method_task1} and~\ref{fig:description_method_task2}, and comprehensive per-team architectural, training, and pre-processing details, together with GitHub implementations, are documented in Tables~\ref{tab:task1summary} and~\ref{tab:task2summary}.

\subsection{Task 1 -- Summary of Methods}

Across the twelve finalist teams, Task~1 solutions converged on a small set of architectural strategies for comparing the two-timepoint OCT pair. Most teams cast the problem as a Siamese or late-fusion comparison, extracting features independently from each B-scan and combining them by subtraction, concatenation, or cross-attention before classification (lumine, MIC~group~6, DF41, scyyd4, STEP). Foundation-model backbones pretrained on retinal imaging -- most commonly RETFound -- were used by more than half of the teams (OptimaTeam, FERLIV, MIPLAB, STEP), typically fine-tuned with a pretext task such as change detection (OptimaTeam), retinal-layer segmentation (FERLIV), or masked-image modelling (via RETFound, MIPLAB), while jkulinzstudents instead relied on a self-supervised DINOv2 backbone with registers. MIPLAB was the only team to fuse all available modalities -- OCT B-scans, infrared localiser images, clinical variables, and 3D volumetric context via slice-level pooling -- into a single classifier. Weighted or focal losses to counter class imbalance, moderate photometric and geometric augmentation, and $k$-fold ensembling with test-time augmentation were near-universal choices across teams.

\subsection{Task 2 -- Summary of Methods}

\begin{figure*}[h!]
	\centering
	\includegraphics[width=\linewidth]{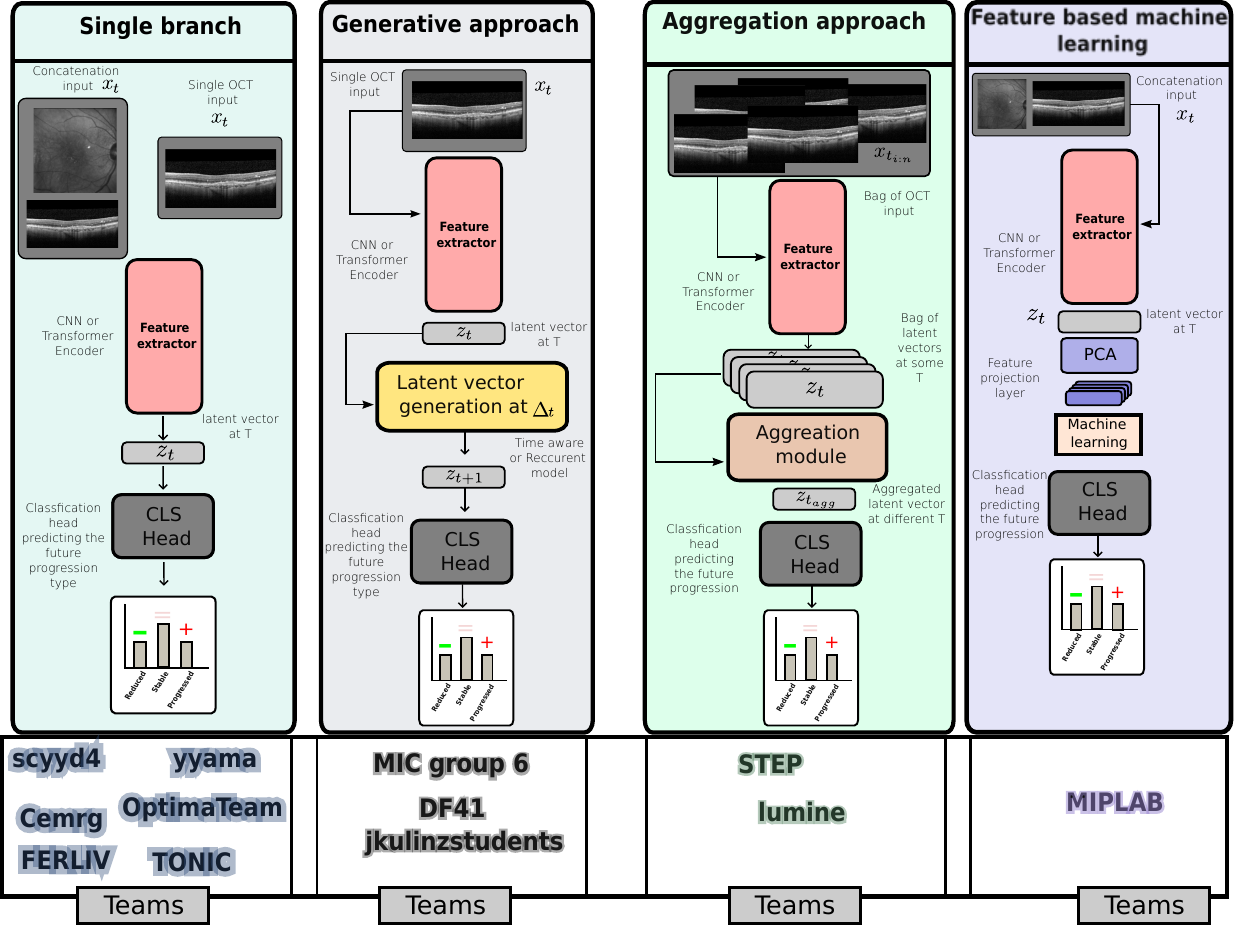}
	\caption{Overview of the methodologies proposed by all teams for Task 2.}
	\label{fig:description_method_task2}
\end{figure*}

Because Task~2 predicts three-month progression from a single visit, without a second real follow-up image, teams adopted one of three broad strategies. Most re-used or fine-tuned their Task~1 architecture and, where applicable, its pretrained weights (lumine, FERLIV, MIPLAB, MIC~group~6, TONIC, Cemrg), effectively treating Task~2 as a single-image variant of Task~1. Two teams instead attempted to hallucinate the missing future state before classifying it: DF41's Patch Progression Masked Autoencoder (PPMAE) reconstructs the follow-up B-scan via masked-patch prediction, and jkulinzstudents' latent-matching model predicts the future RETFound embedding with a negative cosine-similarity objective. STEP and scyyd4 instead framed the problem as multiple-instance learning over the OCT volume, pooling B-scan-level features with an attention mechanism (RETFound backbone and CLAM-SB, respectively). Given the ordinal nature of the progression labels, OptimaTeam and MIPLAB incorporated ordinality-aware objectives -- a discrete Wasserstein-2 term and an ordinal-logistic loss with an immediate-threshold variant, respectively. As in Task~1, weighted losses and ensembling were used throughout to counter the pronounced class imbalance of the \textit{Worsened} class.


\section{Results}
\label{sec:sect7}

Three findings dominate the challenge outcome. First, short-term activity-change
classification (Task~1) is largely tractable: every finalist exceeded a macro-F1 of
0.69 on the primary Brest cohort, and the best team approached inter-expert agreement.
Second, three-month progression prediction (Task~2) is not: composite scores were
compressed near the majority-class baseline, quadratic weighted kappa remained close to
zero, and teams agreed with one another no better than chance---evidence that no method
recovered a genuine prognostic signal from a single visit. Third, and cutting across both
tasks, the models proved fragile under domain shift: transfer to the external Algerian
cohort degraded Task~1 performance for every team and reshuffled the ranking entirely.
We develop each finding below, beginning with the evaluation framework and dataset composition.

\subsection{Evaluation Framework}
\label{sec:eval_framework}

Each team was evaluated separately on the two tasks and both acquisition sites (Brest and Algeria).
The composite score for Task~1 combined macro-F1 ($w=0.50$), Matthews Correlation Coefficient
(MCC, $w=0.25$), and specificity ($w=0.25$):
\begin{equation}
  S_1(m) = 0.50 \cdot F_1 + 0.25 \cdot \mathrm{MCC} + 0.25 \cdot \mathrm{Spec},
  \label{eq:score_t1}
\end{equation}
reflecting the primary importance of balanced class recognition.
Because Task~2 requires ordinal discrimination of longitudinal change,
quadratic weighted kappa (QWK) received the dominant weight:
\begin{equation}
  S_2(m) = 0.10 \cdot F_1 + 0.20 \cdot \mathrm{MCC} + 0.10 \cdot \mathrm{Spec} + 0.60 \cdot \mathrm{QWK}.
  \label{eq:score_t2}
\end{equation}
The overall ranking score was then computed as a weighted combination favouring Task~2:
\begin{equation}
  S_{\mathrm{overall}} = 0.35 \cdot S_1 + 0.65 \cdot S_2.
  \label{eq:score_overall}
\end{equation}
Statistical uncertainty was quantified by stratified bootstrap resampling \citep{Efron1979} ($n=500$) at
the patient level, preserving the longitudinal dependency structure of the data.
Resulting 95\% percentile intervals accompany every reported score.

\subsection{Dataset Characteristics}
\label{sec:dataset_chars}

Panel~A of Figure~\ref{fig:panel1} summarises the class-label distributions and demographic
characteristics for all four site\,$\times$\,task configurations.
The Brest dataset comprises 8,341 visit-pairs across 34 patients for Task~1 and
5,709 visit-pairs across 31 patients for Task~2.
The Algeria (Tlemcen) dataset is substantially smaller:
703 visit-pairs across 5 patients for Task~1 and 574 across 4 patients for Task~2.

The dominant label in both tasks and both sites is \textit{Stable}, reflecting the
natural disease trajectory of neovascular AMD under anti-VEGF treatment.
In Brest Task~1, \textit{Stable} accounts for 71.5\,\% of samples, leaving
13.1\,\% \textit{Reduced}, 9.4\,\% \textit{Worsened}, and 6.0\,\% \textit{Uninterpretable}.
The Algeria cohort exhibits a higher proportion of \textit{Reduced} (22.3\,\%)
and \textit{Worsened} (11.4\,\%) relative to Brest, reflecting different patient-selection criteria.
Task~2 (three-class, excluding \textit{Uninterpretable}) amplifies the class imbalance:
\textit{Stable} reaches 84.3\,\% in Brest and 66.9\,\% in Algeria.

Demographically, the 34-patient Brest \emph{test} cohort is predominantly female (76\,\%;
F:M ratio 3.25:1; mean age 81.5$\pm$8.2 years, range 66--93), a higher female proportion
than the 61.8\,\% (84/136) reported for the full Brest database in
Section~\ref{sec:sect5}, though consistent with the same known prevalence of neovascular
AMD in elderly women~\cite{Rudnicka2012}.
The Algeria cohort is smaller, slightly younger (mean age 76.8$\pm$6.0 years),
and male-dominant (60\,\%).

\subsection{Overall Rankings}
\label{sec:overall_rankings}

Figure~\ref{fig:panel1} (panels B and C) shows the final team rankings on Brest and Algeria,
respectively, with 95\,\% bootstrap confidence intervals.
The full per-metric performance is shown in Figure~\ref{fig:perf_heatmap}.

\subsubsection{Brest (primary test set)}
MIPLAB achieved the highest overall score on Brest (0.491; 95\,\%\,CI [0.474, 0.507]),
followed by MIC group~6 (0.453; [0.439, 0.468]).
The confidence intervals of the top two teams do not overlap, confirming a statistically
significant gap.
From rank~3 downwards (scyyd4 through Cemrg, scores 0.387--0.427) the CIs overlap
substantially, indicating that differences among mid-ranked teams cannot be distinguished
with the available sample size.
jkulinzstudents ranked last (0.346; [0.333, 0.359]).

\subsubsection{Algeria (external validation)}
The Algeria rankings differ markedly from Brest (Figure~\ref{fig:panel1}C).
FERLIV emerged as the top team (0.413; 95\,\%\,CI [0.372, 0.455]),
while Brest winner MIPLAB dropped to rank~5 (0.324; [0.288, 0.354]).
The mean absolute rank shift between the two sites is 3.2 positions,
with STEP showing the greatest instability (rank~6 on Brest, rank~12 on Algeria;
$|\Delta\mathrm{rank}|=6$).
The wider confidence intervals on Algeria reflect the much smaller patient cohort
and highlight the difficulty of drawing firm conclusions from limited external data.

\subsection{Multi-Metric Performance}
\label{sec:multi_metric}

\begin{figure*}[t]
  \centering
  \includegraphics[width=\textwidth]{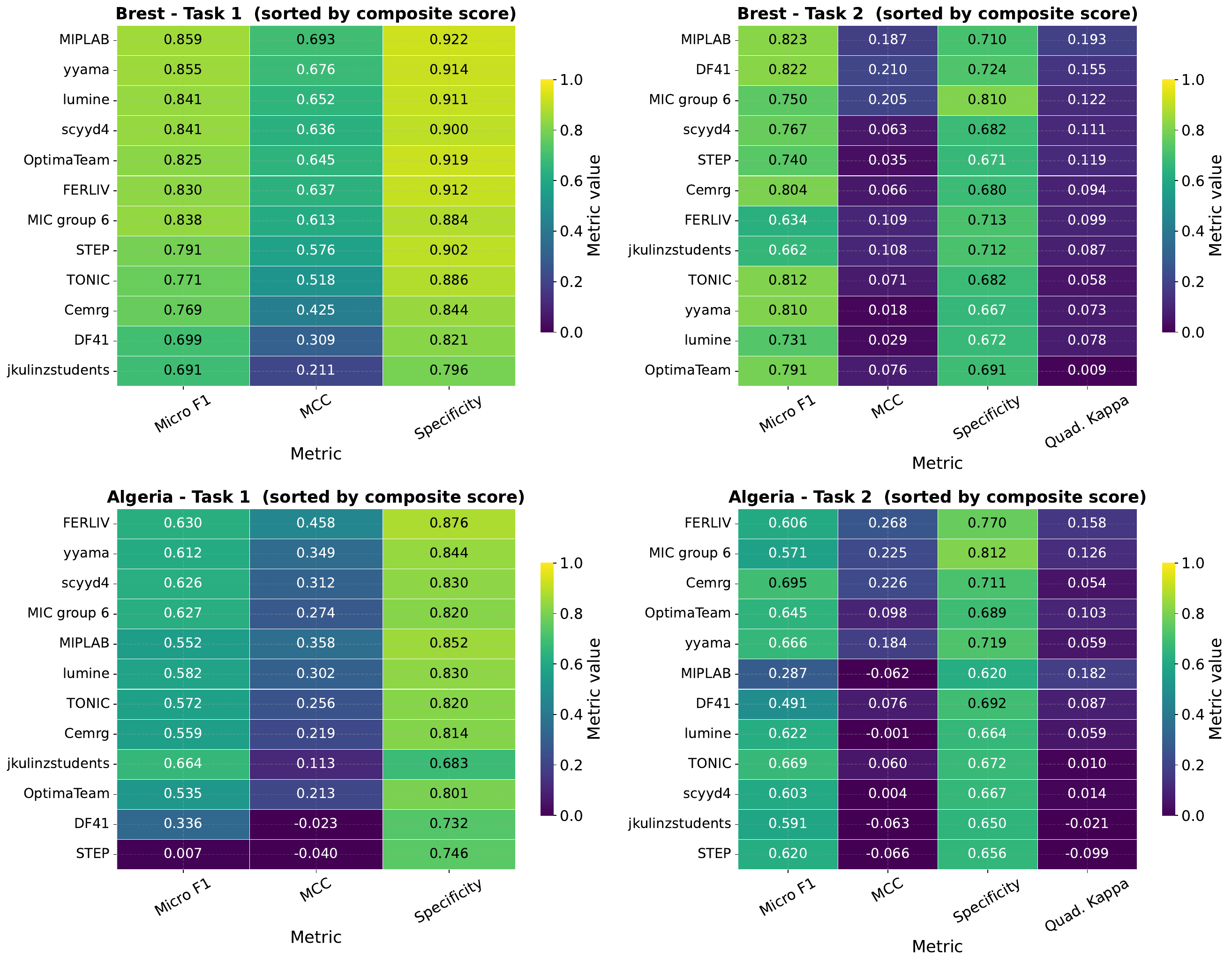}
  \caption{%
    \textbf{Multi-metric performance.}
    Per-metric values for all 12 teams across both sites and tasks, shown as heatmaps
    (perceptually uniform, colour-blind-safe \textit{viridis} scale; brighter~=~higher).
    Teams are sorted by composite score within each panel. Columns give macro-F1, MCC,
    specificity, and---for Task~2---quadratic weighted kappa.
  }
  \label{fig:perf_heatmap}
\end{figure*}

The heatmap in Figure~\ref{fig:perf_heatmap} presents individual metric values (F1, MCC,
specificity, and QWK for Task~2) for all 12 teams across both sites and tasks.

\subsubsection{Task 1}
On Brest, MIPLAB achieved F1$=$0.859 and MCC$=$0.693, with a Task~1 score of 0.833.
All teams attained F1~$>$~0.69 on Brest, suggesting that four-class boundary OCT classification
is tractable with modern deep-learning approaches.
The performance spread (best--worst score difference) was 0.236.
On Algeria, the best F1 was 0.664 (jkulinzstudents), yet the highest composite score
belonged to FERLIV (0.648) owing to stronger MCC (0.458) and specificity (0.876),
highlighting the importance of balanced metric weighting beyond raw F1.
The performance spread on Algeria (0.468) was approximately twice that on Brest,
indicating heterogeneous generalisation.

\subsubsection{Task 2}
Task~2 scores are substantially lower across all teams compared with Task~1.
The best Brest score is 0.306 (MIPLAB), with the worst at 0.159 (OptimaTeam),
a spread of only 0.147.
This compression of scores is attributable to near-zero QWK values: even the best team
achieved QWK$=$0.193, reflecting the extreme difficulty of predicting ordinal activity
change from a single visit-pair.
The MCC values for Task~2 are mostly below 0.25, indicating that most models
are effectively learning to predict the dominant \textit{Stable} class
rather than discriminating between \textit{Reduced} and \textit{Worsened}.
On Algeria Task~2, FERLIV again led with a score of 0.286, while STEP collapsed
to 0.055, the joint-lowest score across any team-site-task combination.

\subsection{Statistical Validation}
\label{sec:statistical_validation}

\subsubsection{Bootstrap confidence intervals and pairwise significance}
\begin{figure*}[t]
  \centering
  \includegraphics[width=\textwidth]{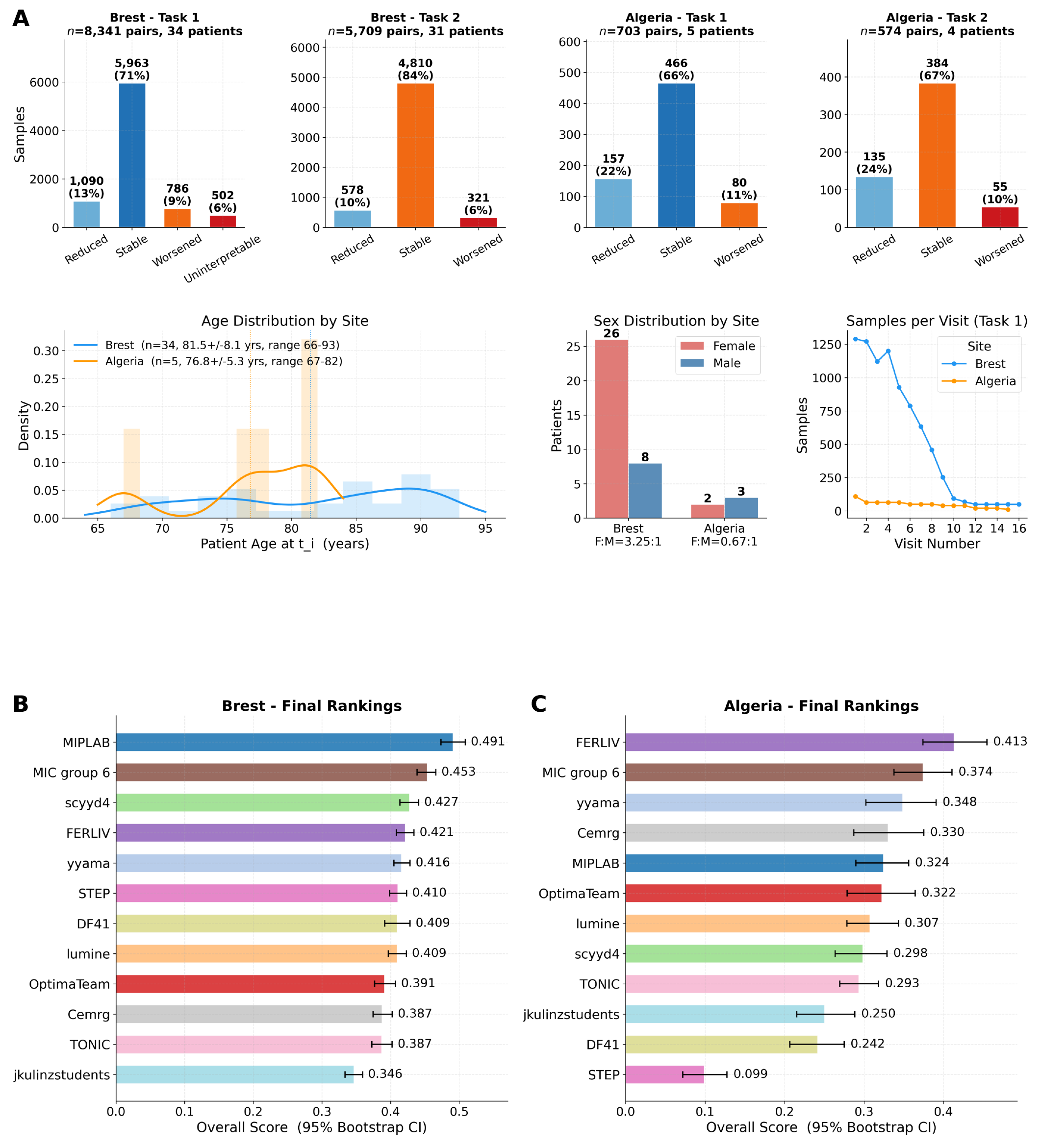}
  \caption{%
    \textbf{Challenge overview and rankings.}
    \textbf{A:} Class-label distribution of the held-out \textbf{test sets} used for evaluation,
    for all four site\,$\times$\,task configurations (Brest Task~1/2, Algeria Task~1/2),
    with absolute counts and proportions.
    \textbf{B:} Final team rankings on the primary Brest test set with 95\,\%
    bootstrap confidence intervals ($n=500$ stratified patient-level resamples).
    \textbf{C:} Final team rankings on the Algeria external-validation set.
    Overall score combines Task~1 and Task~2 composite scores (weights 0.35/0.65; see
    Equations~\ref{eq:score_t1}--\ref{eq:score_overall}).
  }
  \label{fig:panel1}
\end{figure*}

Figure~\ref{fig:mcnemar} shows pairwise McNemar significance maps \citep{McNemar1947} for all four
site\,$\times$\,task configurations.
On Brest Task~1, MIPLAB's predictions are significantly different ($p<0.05$) from all
other teams after Bonferroni correction.
Among mid-ranked teams (ranks 3--9), significant pairwise differences are sparse,
consistent with the overlapping confidence intervals in Figure~\ref{fig:panel1}B.
For Task~2 (Brest and Algeria), very few team pairs show significant disagreement,
reflecting the near-identical Stable-biased prediction patterns.

Figure~\ref{fig:forest} shows bootstrap forest plots of the individual
metrics (F1, MCC, specificity, QWK) for Tasks~1 and~2.
The intervals for Task~1 F1 are narrow and non-overlapping for the top three teams,
confirming robust rank stability.
The Task~2 MCC intervals are wide and centred near zero for most teams,
confirming that no team reliably discriminates between Stable and non-Stable
in the longitudinal change prediction task.

Supplementary Figure~\ref{fig:panel_bootstrap} shows the complete bootstrapped score distributions
(panel~A) and rank stability under resampling (panel~B) for both sites.
On Brest, the violin plots confirm that MIPLAB's overall-score distribution is well
separated from all other teams, while mid-range teams (scyyd4 through lumine)
have overlapping distributions, indicating ranking uncertainty.
On Algeria, all distributions widen considerably due to the smaller sample size.
The ranking stability box plots (panel~B) show that the Task~1 rank order is more
stable than the Task~2 and overall ranks for most teams, consistent with the
higher discriminative power of Task~1.

\subsubsection{Inter-team agreement}
For Brest Task~1, mean pairwise Cohen's $\kappa$~\citep{Cohen1960} $= 0.494$ (range [0.158, 0.769];
23 of 66 pairs $>0.6$), indicating moderate-to-substantial agreement among top teams.
For Brest Task~2, mean $\kappa$ collapses to $0.096$ (range [$-0.048$, $0.466$];
zero pairs $>0.6$), confirming near-random agreement:
teams predict Task~2 labels essentially independently, with no shared signal.
Algeria Task~1 sits between the two Brest tasks ($\kappa=0.292$),
while Algeria Task~2 shows near-zero agreement ($\kappa=0.043$).
Supplementary Figure~\ref{fig:panel_agreement} shows the full pairwise $\kappa$ matrices for all
four site\,$\times$\,task configurations.
High-$\kappa$ clusters are visible among the top-3 Brest Task~1 teams (MIPLAB,
yyama, and MIC group~6), confirming that their strong performance reflects a
shared prediction signal rather than independent error patterns.

\subsection{Classification Error Analysis}
\label{sec:error_analysis}

\subsubsection{Per-class performance}
Figure~\ref{fig:perclass} shows per-class F1 scores for all teams across both sites
and tasks.
In Brest Task~1, the \textit{Stable} class is predicted most reliably
(mean F1$=$0.880, std$=$0.030 across teams), while
\textit{Uninterpretable} is the most challenging class (mean F1$=$0.357),
followed by \textit{Worsened} (0.584) and \textit{Reduced} (0.675).
The near-zero variance for \textit{Stable} reflects the class imbalance:
all models default to \textit{Stable} to maximise accuracy.

In Brest Task~2, the pattern is even more extreme:
\textit{Stable} achieves mean F1$=$0.863 while \textit{Worsened} mean F1$=$0.018
(minimum $= 0.000$ for several teams), indicating that current architectures
completely fail to identify worsening visits.
\textit{Reduced} is marginally better (mean F1$=$0.232) but still far below clinically
useful thresholds.

On Algeria, per-class F1 values are consistently lower, and the \textit{Stable} class
advantage is less pronounced (mean 0.625 in T1, versus 0.880 on Brest),
reflecting the different class distribution and the absence of in-domain training data.

\subsubsection{Confusion matrices: best-performing teams}
\begin{figure*}[t]
  \centering
  \includegraphics[width=\textwidth]{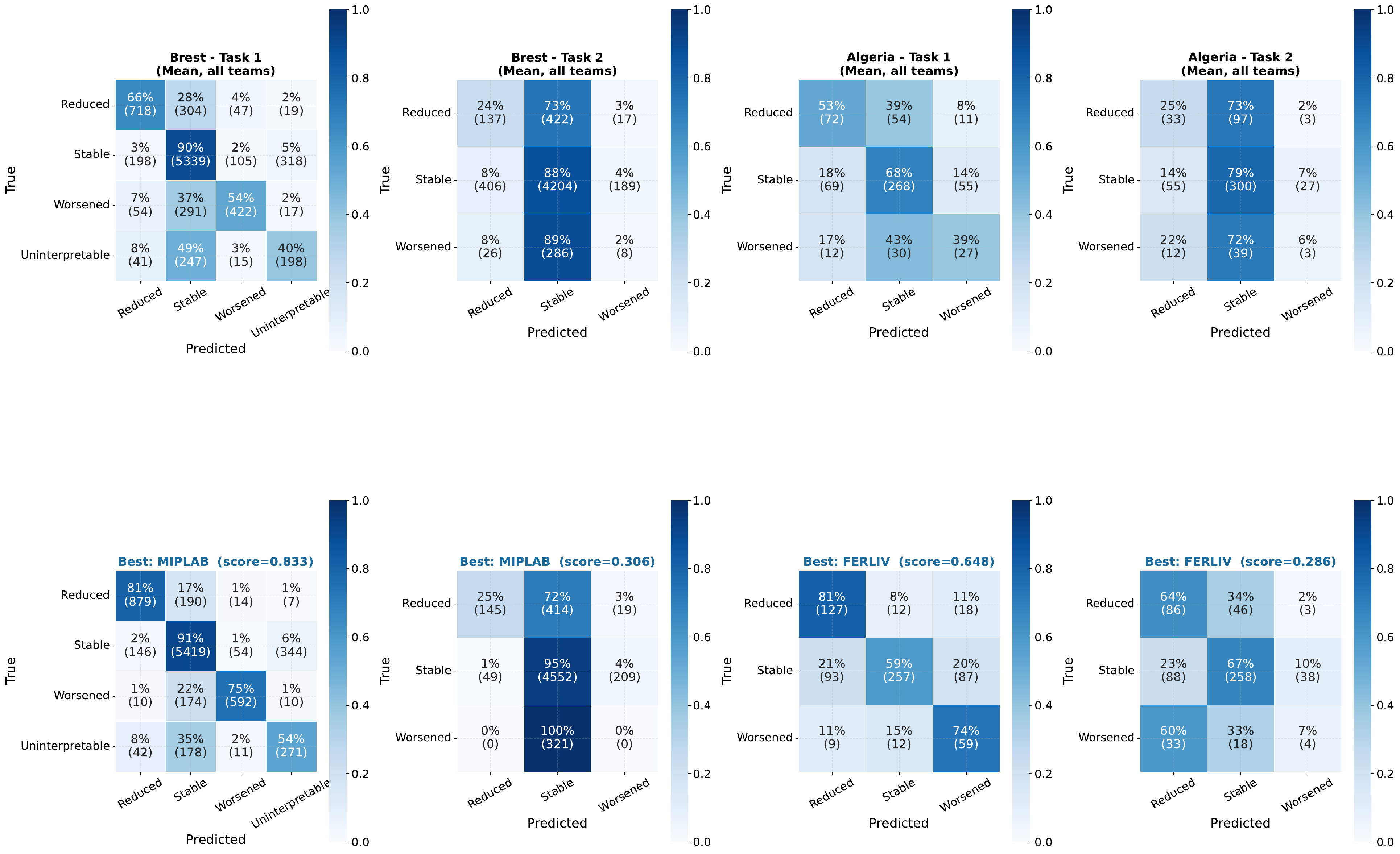}
  \caption{%
    \textbf{Confusion matrices.}
    Row-normalised confusion matrices for each site\,$\times$\,task scenario.
    Top row: mean confusion matrix across all 12 teams; bottom row: the
    best-performing team (MIPLAB for Brest, FERLIV for Algeria).
    Each cell gives the row-normalised fraction (percentage) with the absolute count
    in parentheses. The complete per-team matrices are provided in
    Supplementary Figures~\ref{fig:panel_allcm_brest_t1}--\ref{fig:panel_allcm_brest_t2} and~\ref{fig:panel_allcm_algeria_t1}--\ref{fig:panel_allcm_algeria_t2}.
  }
  \label{fig:cms}
\end{figure*}

Figure~\ref{fig:cms} shows, for each site\,$\times$\,task scenario, the mean
confusion matrix across all 12 teams (top row) and that of the best-performing team
(bottom row); the complete per-team confusion matrices are provided in Supplementary
Figures~\ref{fig:panel_allcm_brest_t1}--\ref{fig:panel_allcm_brest_t2} and~\ref{fig:panel_allcm_algeria_t1}--\ref{fig:panel_allcm_algeria_t2}.
On Brest Task~1 (MIPLAB), most errors occur at the \textit{Stable}--\textit{Reduced}
boundary, with \textit{Uninterpretable} frequently confused with \textit{Stable}.
On Algeria Task~1 (FERLIV), the \textit{Stable} mass absorbs many true \textit{Reduced}
and \textit{Worsened} samples, reflecting domain shift.
In Task~2 (both sites), the matrices confirm the systematic collapse towards
\textit{Stable}: almost no visit is classified as \textit{Worsened}.

\subsubsection{All-team confusion matrices}
Supplementary Figures~\ref{fig:panel_allcm_brest_t1}--\ref{fig:panel_allcm_brest_t2} and~\ref{fig:panel_allcm_algeria_t1}--\ref{fig:panel_allcm_algeria_t2}
present the normalised confusion matrices for all 12 teams on both tasks for the Brest and
Algeria datasets, respectively.
On Brest Task~1 (Supplementary Figure~\ref{fig:panel_allcm_brest_t1}), even the lowest-ranked teams
achieve high diagonal mass on the \textit{Stable} class, while the \textit{Worsened}
and \textit{Uninterpretable} cells exhibit systematic off-diagonal leakage into \textit{Stable}.
Top teams (MIPLAB, yyama) show considerably fuller diagonals than bottom teams
(DF41, jkulinzstudents), particularly for the \textit{Reduced} class.
On Brest Task~2 (Supplementary Figure~\ref{fig:panel_allcm_brest_t2}), the \textit{Worsened} row
collapse is universal: all 12 teams predict fewer than 10\,\% of true \textit{Worsened}
visits correctly, with predictions concentrated entirely in \textit{Stable}.
On the Algeria site (Supplementary Figures~\ref{fig:panel_allcm_algeria_t1} and~\ref{fig:panel_allcm_algeria_t2}), the domain-shift effect
is visible as increased off-diagonal mass and reduced diagonal values across all teams
for Task~1, while Task~2 shows the same \textit{Worsened}-collapse pattern.

\subsubsection{Patient-level misclassification}
Figure~\ref{fig:patient} presents per-patient error-rate heatmaps for all four
site\,$\times$\,task scenarios, with teams as rows (sorted by composite score) and
individual patients as columns (sorted from hardest to easiest).
Two distinct patient archetypes emerge, most clearly on Brest Task~1:
(i)~patients classified correctly by almost every team across their longitudinal
sequence, largely those with predominantly \textit{Stable} records; and
(ii)~a small set of consistently hard patients (the left-most columns) that almost
every team misclassifies, often those whose disease course involves oscillating
\textit{Reduced}/\textit{Worsened} transitions.
Because the hardest patients are shared across teams rather than team-specific, simple
ensembling is unlikely to resolve them.

\subsubsection{Qualitative examples}
\begin{figure*}[t]
  \centering
  \includegraphics[width=\textwidth]{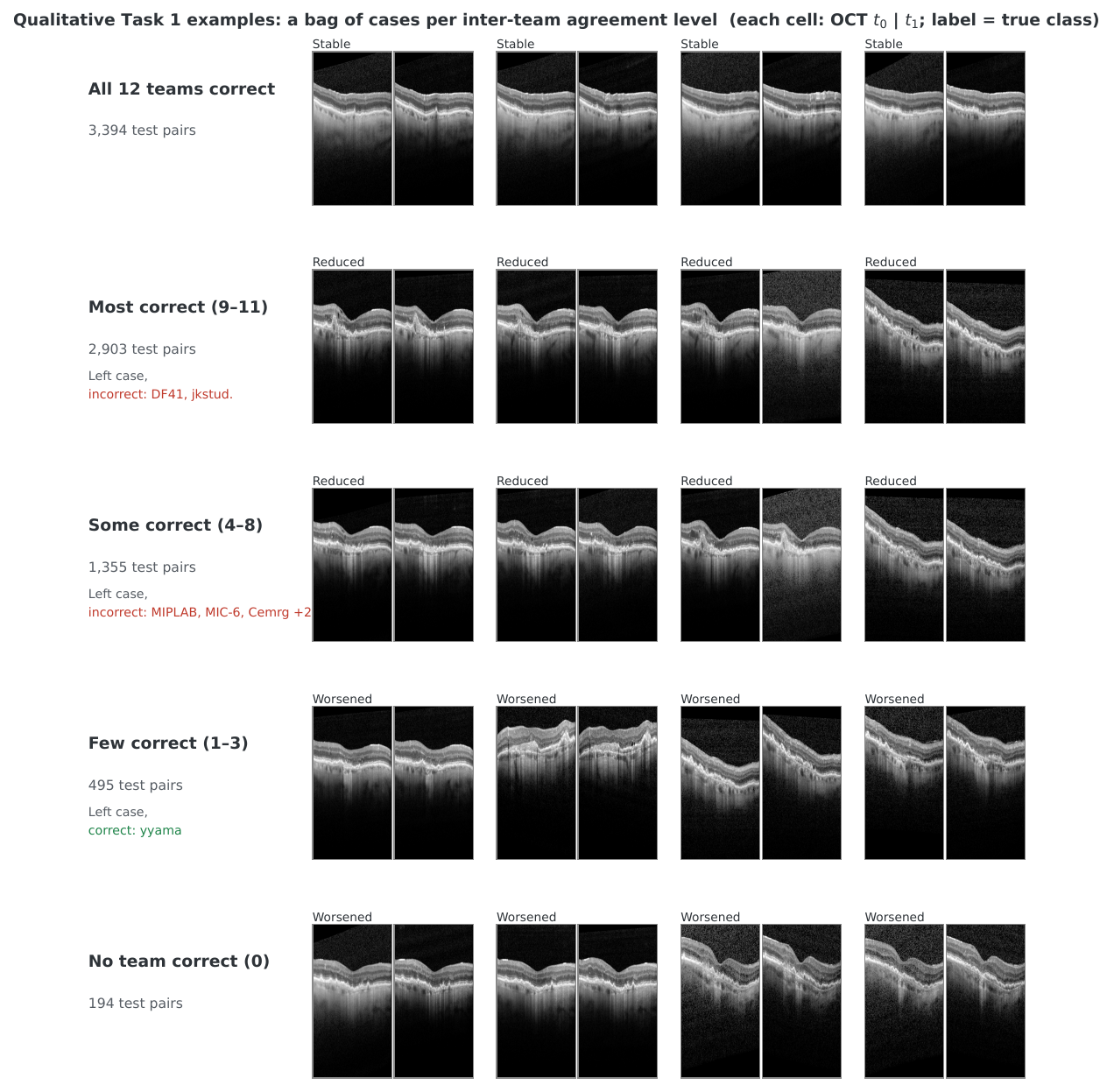}
  \caption{%
    \textbf{Qualitative Task~1 examples by inter-team agreement (Brest test set).}
    Each row corresponds to a level of inter-team agreement---from all 12 finalist teams
    correct (top) to no team correct (bottom)---and shows a \emph{bag} of four
    representative consecutive OCT B-scan pairs (each cell: baseline $t_0$ \textbar{}
    follow-up $t_1$; the label above each pair is the true class). The left column reports
    the number of Brest test pairs in that agreement bin and, for the leftmost case, names
    the minority team set (the few teams that succeeded, or the few that failed).
  }
  \label{fig:octcases}
\end{figure*}

Figure~\ref{fig:octcases} shows representative Brest Task~1 visit-pairs stratified by
inter-team agreement, from cases solved by all 12 finalist teams down to cases missed by
every team, together with the identity of the correct and incorrect teams.
A clear pattern emerges: the cases solved by all teams are predominantly clear
\textit{Stable} scans with well-defined layers and little inter-visit change, whereas the
cases missed by all teams carry genuine \textit{Worsened} or \textit{Reduced} changes---
often subtle fluid shifts or motion-degraded acquisitions---that every model instead
predicts as \textit{Stable}. Difficulty is therefore largely shared across teams rather
than method-specific, reinforcing the quantitative finding that current models default to
the majority class precisely when a clinically important change is present.

\subsection{Cross-Site Generalisation}
\label{sec:cross_site}

Figure~\ref{fig:crosssite} shows paired Brest--Algeria comparisons for each metric and team.
For Task~1, all 12 teams exhibit F1 degradation when moving from Brest to Algeria,
with declines ranging from $-0.027$ (jkulinzstudents) to $-0.784$ (STEP).
The median F1 drop is approximately 0.25, indicating a consistent but variable
domain-shift effect.
MCC degrades more severely for several teams (e.g.\  OptimaTeam: $-0.432$),
suggesting that imbalance-corrected metrics are particularly sensitive to distribution shift.

STEP's Algeria failure is the most pronounced: Task~1 F1$=$0.007 and composite
score$=$0.180, representing near-chance performance.
Post-challenge analysis indicated that STEP's bidirectional cross-attention module
collapsed on the Algeria distribution, likely due to overfit to the temporal scan-sequence
statistics of the Brest site.

For Task~2, cross-site generalisation is more mixed.
Several teams show slight MCC or specificity improvements on Algeria (e.g.\  FERLIV:
$+0.159$ MCC; yyama: $+0.166$ MCC), which is partly explained by the different
class distribution in Algeria Task~2 (less extreme Stable dominance at 66.9\,\% vs.\  84.3\,\%),
making non-Stable prediction marginally easier.
However, QWK remains low on both sites.

\subsection{Temporal and Clinical Factors}
\label{sec:temporal}

\subsubsection{Visit-level performance}
\begin{figure*}[tp]
  \centering
  \includegraphics[width=\textwidth,height=0.94\textheight,keepaspectratio]{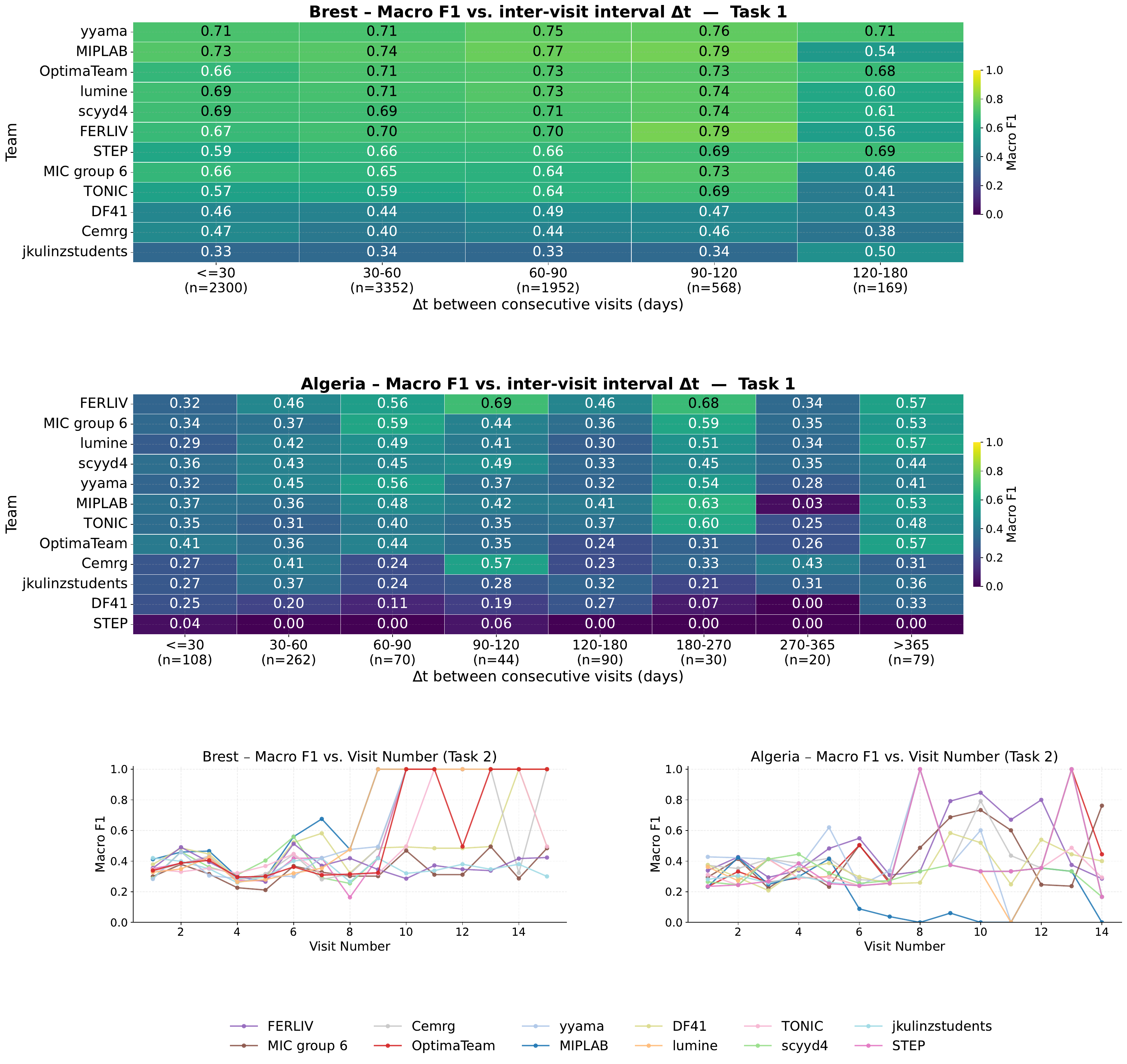}
  \caption{%
    \textbf{Temporal analysis.}
    \textbf{A:} Per-team macro-F1 stratified by inter-visit time interval
    ($\Delta t$ in days) for Brest; Task~1.
    \textbf{B:} The same stratification for Algeria; Task~1.
    \textbf{C--D:} Team macro-F1 as a function of visit index for
    Brest~(C) and Algeria~(D); Task~2.
  }
  \label{fig:panel_temporal}
\end{figure*}

Figure~\ref{fig:panel_temporal} (panels A--B) shows the inter-visit time interval ($\Delta t$) macro-F1
heatmaps for Brest (A) and Algeria (B), both for Task~1.
Performance does not uniformly degrade with longer inter-visit intervals,
which would be expected if temporal context were the primary driver.
Instead, bins with very short ($<30$\,days) and very long ($>180$\,days) intervals
tend to have higher F1 for some teams, while mid-range intervals are harder.
This non-monotonic behaviour suggests that the difficulty of classification is driven
more by the biological variability at the visit than by the elapsed time per se.

Panels C--D (Figure~\ref{fig:panel_temporal}) show Task~2 visit-level macro-F1 as a function
of visit index for Brest (C) and Algeria (D).
On Brest, team-averaged visit-level F1 ranges from 0.353 (visit~13) to 0.639 (visit~16),
with no monotonic trend across the longitudinal sequence.
The dip at visit~13 coincides with a cluster of patients undergoing treatment switches,
a confounder not captured by the imaging alone.
Algeria shows lower overall F1 (range 0.208--0.461) and greater volatility due to the
smaller number of patients.

\subsubsection{Demographic stratification}
Supplementary Figure~\ref{fig:panel_clinical} presents performance stratified by age group, sex, and
laterality (eye side).
No significant interaction between sex and Task~1 F1 was detected across teams,
consistent with the absence of known sex-specific differences in AMD disease course
under treatment.
Similarly, laterality (left vs.\ right eye) did not produce systematic performance
differences.
Age group shows a modest trend: patients older than 85 years exhibit slightly higher
Task~1 F1, likely because older patients have longer, more stable longitudinal trajectories
that are easier to classify as \textit{Stable}.


\begin{figure*}[t]
  \centering
  \includegraphics[width=\textwidth]{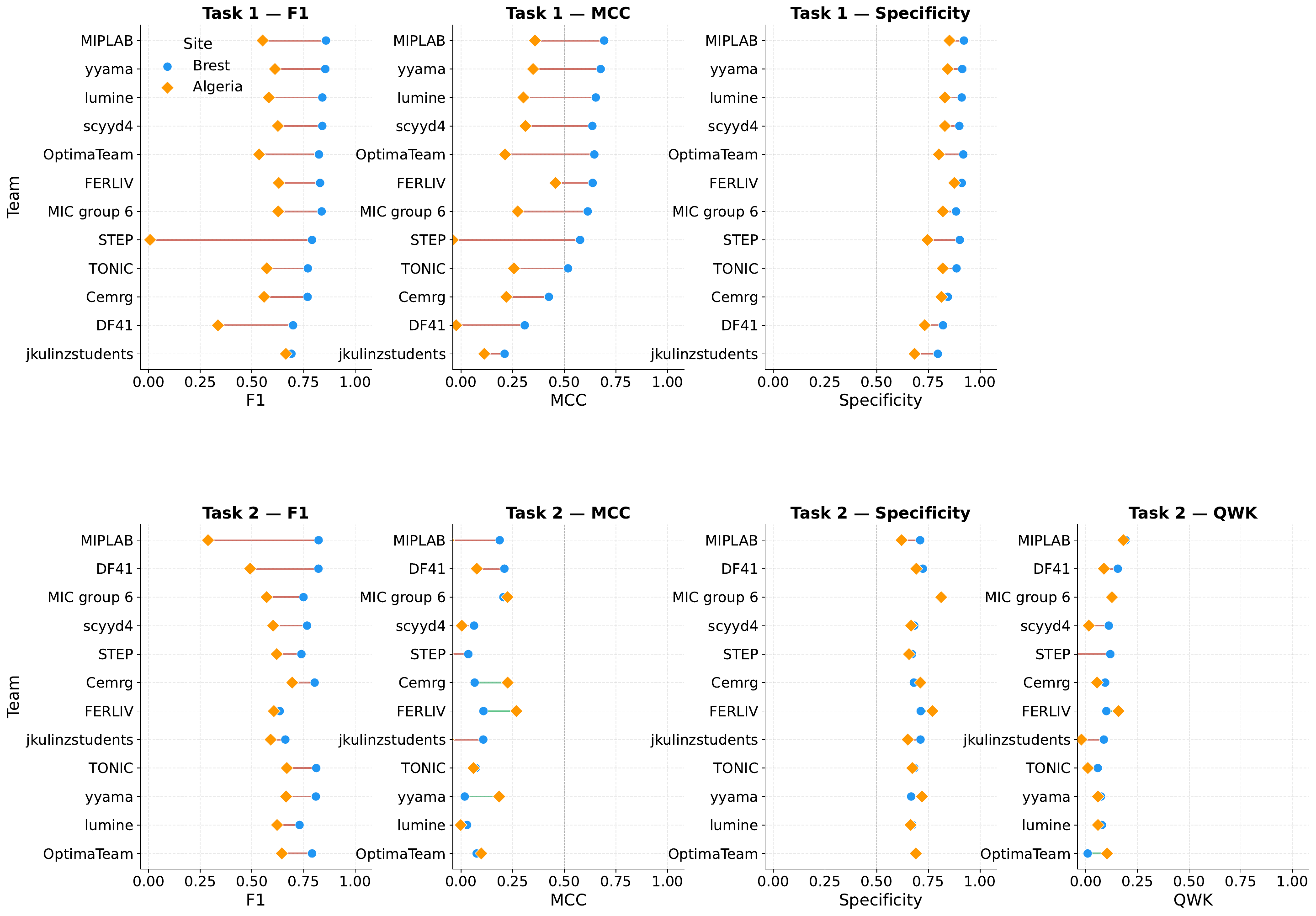}
  \caption{%
    \textbf{Cross-site generalisation.}
    For each metric and team, the Brest value (blue circle) and the Algeria value
    (orange diamond) are joined by a line; the horizontal gap gives the magnitude and
    direction of the between-site change. Top row: Task~1; bottom row: Task~2.
  }
  \label{fig:crosssite}
\end{figure*}

\begin{figure*}[t]
  \centering
  \includegraphics[width=0.92\textwidth]{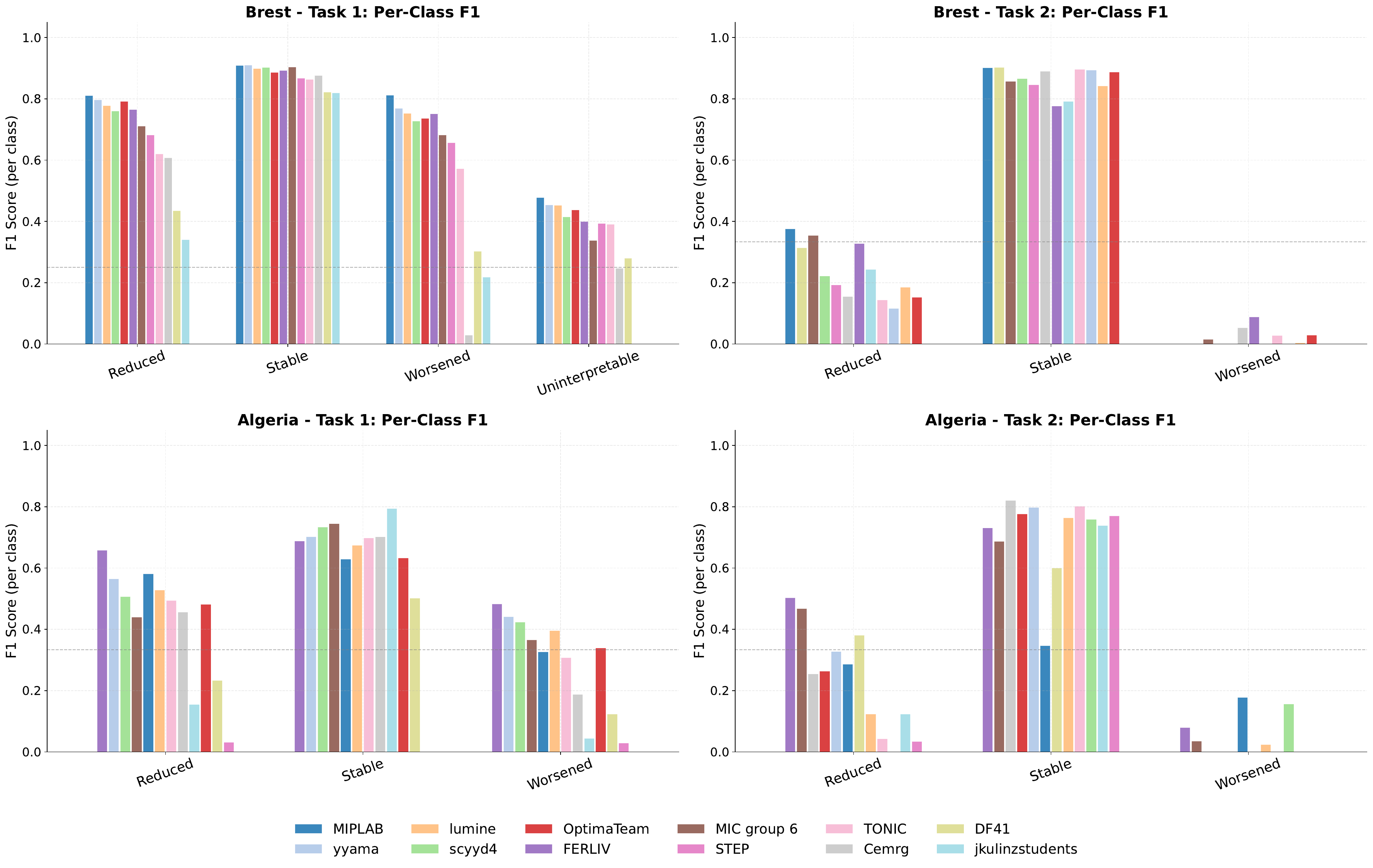}
  \caption{%
    \textbf{Per-class F1.}
    Per-class F1 scores (grouped bars) for all 12 teams across the four
    site\,$\times$\,task panels; the dashed horizontal line marks the
    random-baseline level ($1/n_\mathrm{classes}$).
  }
  \label{fig:perclass}
\end{figure*}

\begin{figure*}[t]
  \centering
  \includegraphics[width=\textwidth]{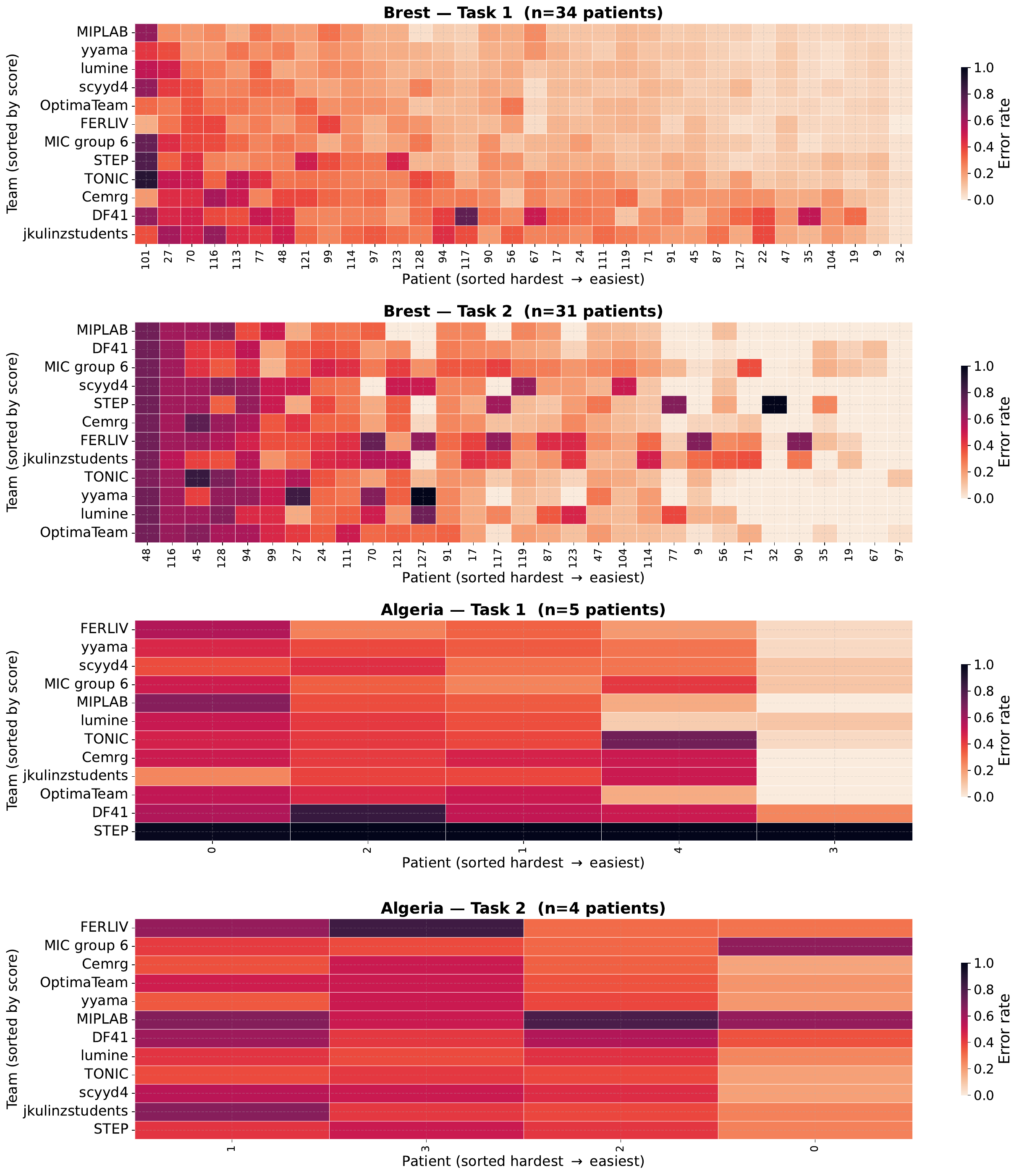}
  \caption{%
    \textbf{Patient-level error rates.}
    Per-patient error-rate heatmaps for all four site\,$\times$\,task scenarios:
    rows are teams (sorted by composite score), columns are individual patients
    (sorted hardest to easiest), and colour encodes the fraction of that patient's
    visits misclassified by the team.
  }
  \label{fig:patient}
\end{figure*}

\begin{figure*}[t]
  \centering
  \includegraphics[width=\textwidth]{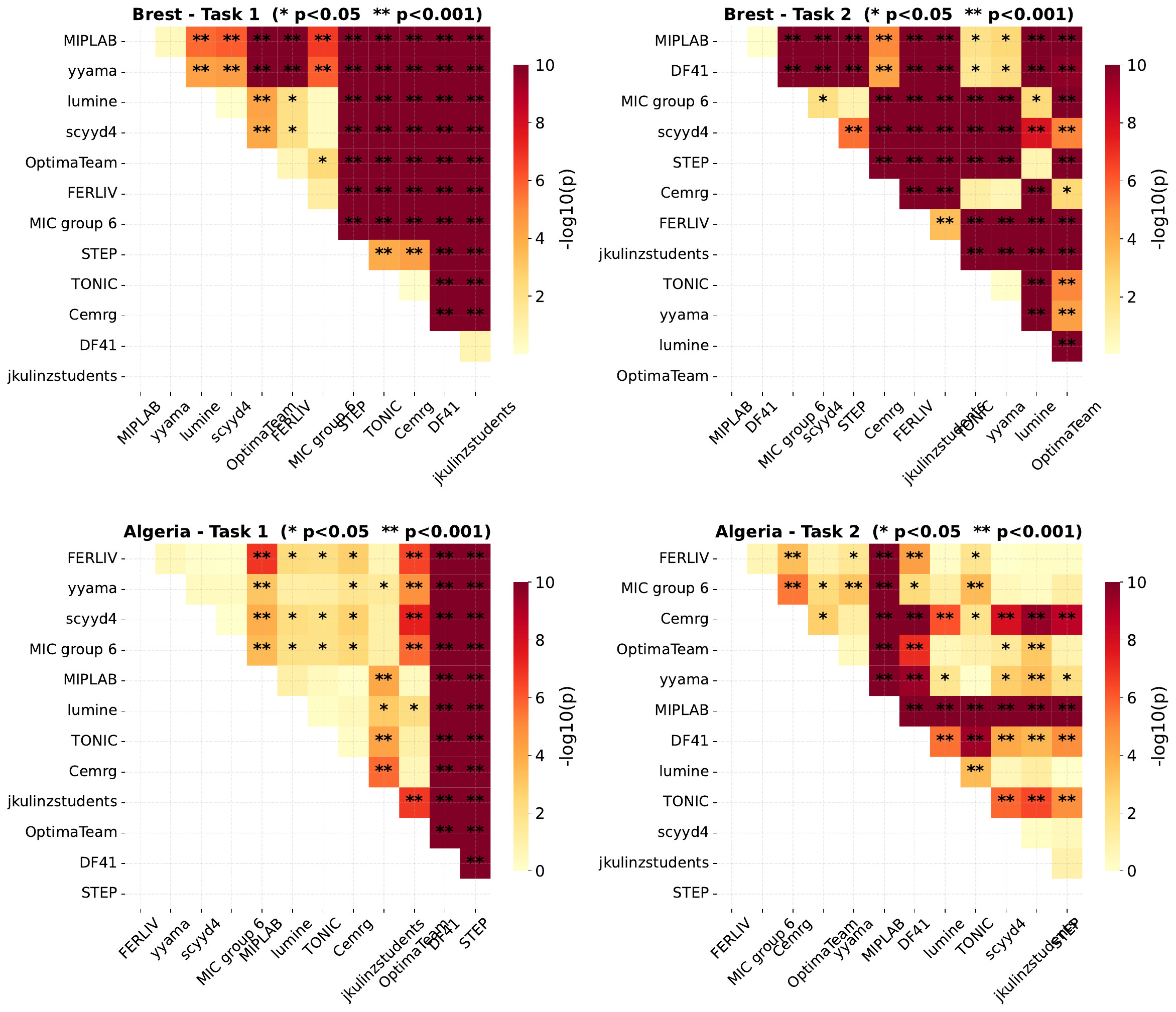}
  \caption{%
    \textbf{Pairwise statistical significance.}
    McNemar test significance maps for all four site\,$\times$\,task configurations.
    Cell colour encodes $-\log_{10}(p)$ for the difference in error patterns between
    each team pair; asterisks mark significance after Bonferroni correction for the
    66 team pairs (\textasteriskcentered\ $p<0.05$; \textasteriskcentered\textasteriskcentered\ $p<0.001$).
  }
  \label{fig:mcnemar}
\end{figure*}

\begin{figure*}[t]
  \centering
  \includegraphics[width=\textwidth]{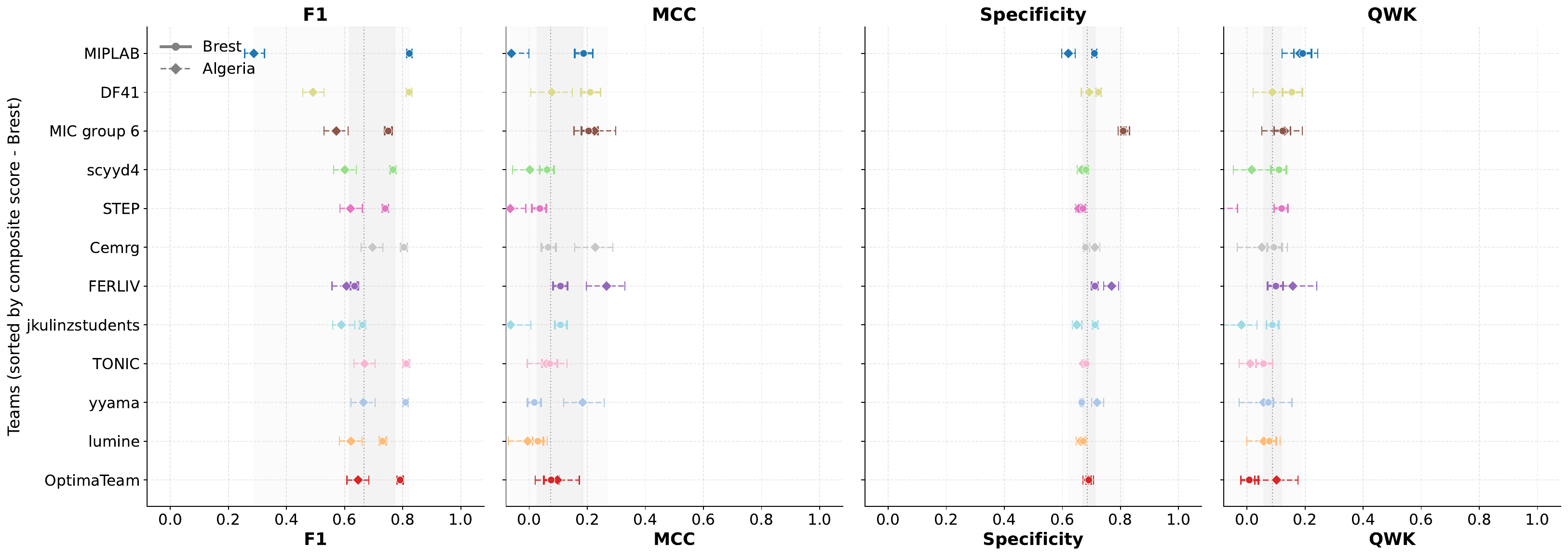}\\[4pt]
  \includegraphics[width=0.75\textwidth]{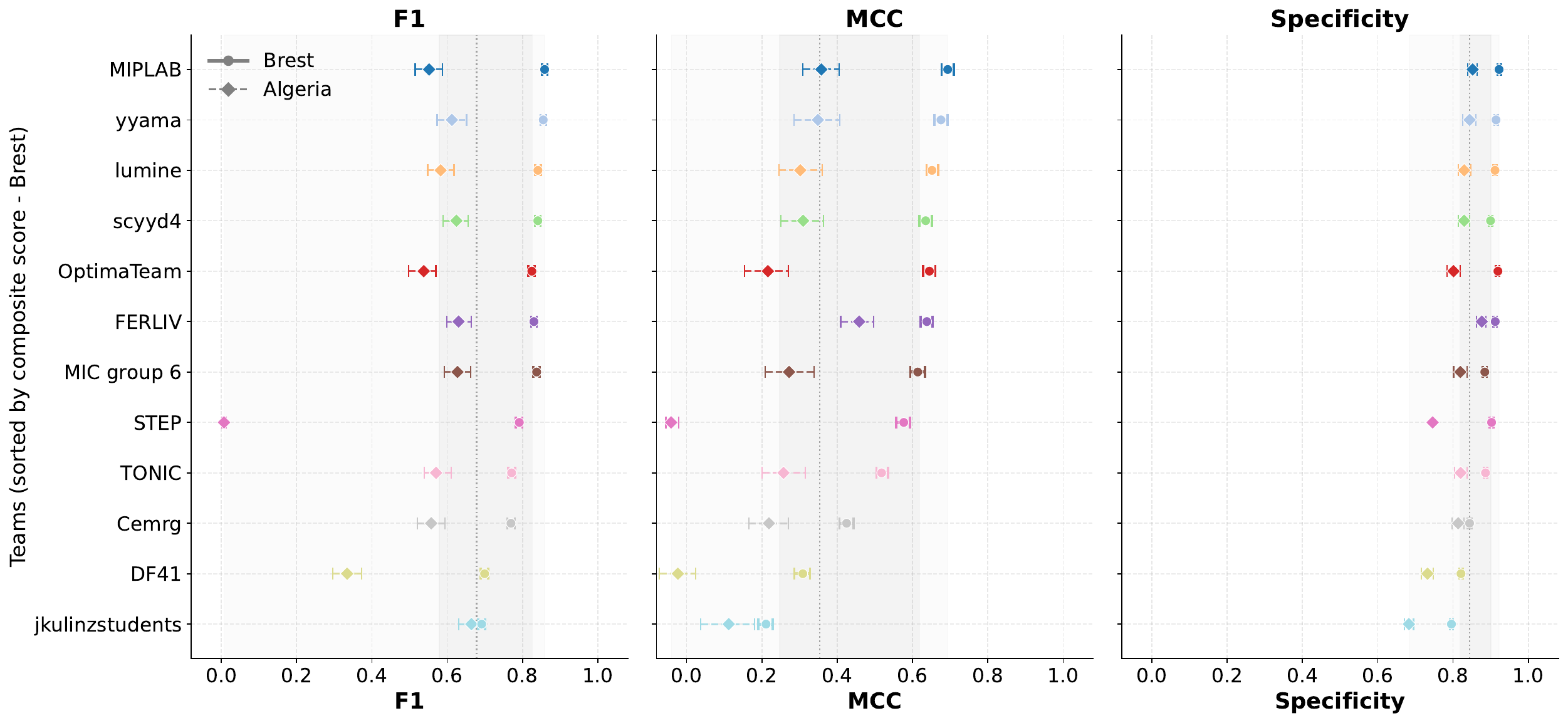}
  \caption{%
    \textbf{Bootstrap metric intervals.}
    Forest plots of the individual metrics for Task~2 (top: F1, MCC, specificity, QWK)
    and Task~1 (bottom: F1, MCC, specificity).
    Each marker is the team's point estimate and the horizontal bar spans the
    2.5th--97.5th bootstrap percentile ($n=500$ patient-level resamples);
    blue = Brest, orange = Algeria.
  }
  \label{fig:forest}
\end{figure*}

\section{Discussion}
\label{sec:sect8}

The MARIO challenge yields a clear two-part message. Automated monitoring of short-term
neovascular activity---the judgement a clinician makes when comparing two consecutive OCT
examinations---is now within reach of current deep-learning methods: the best model reached a
macro-F1 of 0.859, close to the agreement observed between two retinal experts on the same data.
Predicting whether a lesion will progress over the following three months from a single visit,
by contrast, remains unsolved: no finalist meaningfully outperformed a majority-class predictor,
and the near-zero inter-team agreement indicates that the difficulty is intrinsic to the problem
as posed rather than a shortcoming of any individual method. A third result cuts across both
tasks: every method degraded sharply when transferred to an unseen population and device.
Below we interpret these findings and their clinical implications, examine the sensitivity to
domain shift through the lens of fairness, and review the methodological trends that
distinguished the strongest submissions.

\subsection{Statistical Robustness of the Rankings}

The bootstrap analysis (Figure~\ref{fig:forest}) confirms that the Brest Task~1 ranking is
statistically well-grounded for the top two positions: MIPLAB and MIC group~6 have
non-overlapping 95\,\% confidence intervals, and MIPLAB's predictions differ significantly
from all other teams according to the McNemar test (Figure~\ref{fig:mcnemar}).
However, for ranks 3--9 the CIs overlap substantially, meaning that the precise ordering
of mid-field teams should not be over-interpreted.
This pattern is consistent with findings from other multi-team biomedical image analysis
challenges where the top performers are reliably identified but mid-range rankings lack
statistical power~\cite{Wiesenfarth2021,MaierHein2018}.

For Task~2, the collapse of inter-team agreement to near-zero Cohen's $\kappa$ ($\bar{\kappa}=0.096$
on Brest; $0.043$ on Algeria) and the absence of any statistically significant pairwise
McNemar differences tells a clear story: no team has found a reliable signal for the
longitudinal AMD progression prediction problem as formulated here.
The best Brest Task~2 score (MIPLAB, 0.306) is driven almost entirely by accurate
prediction of the majority \textit{Stable} class, not by genuine discrimination of
\textit{Reduced} vs.\  \textit{Worsened} outcomes.

\subsection{Cross-Site Generalisation and Domain Shift}

The cross-site comparison (Figure~\ref{fig:crosssite}) reveals a consistent Task~1 F1 penalty
when algorithms developed on Brest data are applied to the Algeria cohort,
with a median drop of $\approx 0.25$ and an extreme case of $-0.784$ for the STEP team.
The STEP team's near-complete failure (Algeria T1 F1$=$0.007) is traceable to its
bidirectional cross-attention architecture, which was calibrated on the Brest scan-volume
statistics: the Algeria dataset contains B-scan sequences that do not reconstruct a full
OCT volume, violating the architectural assumptions of the model.
This is a textbook case of \textit{acquisition-protocol shift}~\cite{MATTA2024109256}
and underscores the importance of challenge designs that explicitly test robustness
to scanner and protocol variation.

The more gradual degradation observed in other teams reflects a combination of
\textit{covariate shift} (different age and sex distributions) and
\textit{prior probability shift} (the Algeria cohort has proportionally more
\textit{Reduced} and \textit{Worsened} examples).
Teams that relied on domain-specific pretraining (e.g.\  FERLIV using RETFound)
maintained relatively higher Algeria performance, suggesting that ophthalmic-domain
knowledge provides some protection against site-level generalisation failures.

Beyond its technical framing, this cross-site gap is a fairness concern. The Brest and
Tlemcen cohorts differ in ethnicity, age and sex distribution, and imaging protocol, and a
model that performs at expert level for the former while failing for the latter would deliver
inequitable care if deployed globally~\cite{ueda2024fairness,LIM2024100096}. That the entire
ranking reordered between the two sites---the Brest winner falling to fifth on Algeria---shows
that a leaderboard established on a single population is a poor predictor of real-world equity.
We therefore argue that external, demographically distinct validation should be a default
component of AMD monitoring benchmarks rather than an optional extra, and that reporting
site-stratified performance is as important as reporting a single aggregate score. The
comparatively smaller degradation of domain-pretrained models is encouraging, but the Tlemcen
cohort is small (five patients); the wide confidence intervals it produces are themselves a
reminder that equitable evaluation ultimately requires larger and more diverse external data.

\subsection{Error Patterns and Clinical Significance}

The per-class analysis (Figure~\ref{fig:perclass}) highlights two clinically critical failure modes.
First, the \textit{Uninterpretable} class in Task~1 (mean F1$=$0.357) is consistently
misclassified as \textit{Stable}, likely because the visual boundary between a
motion-artefacted scan (\textit{Uninterpretable}) and a genuinely stable retina is subtle.
Second, and more clinically concerning, the \textit{Worsened} class in Task~2 achieves
near-zero F1 for virtually every team (mean $0.018$, range [$0.000$, $0.089$]).
In clinical practice, missing a \textit{Worsened} visit has direct consequences:
delayed anti-VEGF re-treatment risks irreversible photoreceptor loss.
Future work must address this asymmetric cost structure, for example through
class-weighted losses, ordinal regression losses, or threshold-calibrated decision rules.

Read positively, the Task~1 result has a concrete clinical corollary. Deciding, from two
consecutive B-scans, whether neovascular activity has reduced, stayed stable, or worsened is
precisely the judgement that drives the treat-and-extend decision at each monitoring visit
(Section~\ref{sec:sect2}). A model that reproduces this judgement at close to inter-expert
agreement could support---though not yet replace---the clinician in triaging stable eyes for
interval extension, helping to relieve the injection and imaging burden of nAMD care.
Translating this into practice would require prospective validation, calibrated uncertainty
estimates so that borderline cases are routed to a human reader, and the asymmetric-cost
handling above so that missed worsening remains rare. Task~2, by contrast, is not yet at a
stage where clinical use can be contemplated: predicting three-month progression from a single
visit likely demands richer inputs than one B-scan---longitudinal history, treatment timing,
and volumetric context---rather than a better classifier on the current formulation.

The qualitative examples in Figure~\ref{fig:octcases} (Section~\ref{sec:error_analysis})
make these failure modes concrete. Beyond the shared solved/missed extremes reported there,
the intermediate strata are informative too: when only a few teams succeed, the successful
set is small and inconsistent across cases (e.g.\ a single team such as yyama on one case),
and when only a few fail, the failing set is likewise idiosyncratic (e.g.\ DF41 and
jkulinzstudents). Because the same hard visit-pairs defeat most methods simultaneously, the
bottleneck is the discriminability of subtle exudative change in a single B-scan pair rather
than any particular architecture---which is why ensembling the current models offers little
headroom and why richer inputs are the more promising direction.

\subsection{Temporal Dynamics}

The $\Delta t$ heatmap (Figure~\ref{fig:panel_temporal}A--B) shows that inter-visit time interval
is not the primary driver of classification difficulty: performance does not degrade
monotonically with longer gaps.
Instead, patient-specific disease dynamics appear to dominate.
The visit-level F1 trajectory (Figure~\ref{fig:panel_temporal}C--D) further reveals that performance
at a given visit is highly patient-dependent (Figure~\ref{fig:patient}), with some patients
producing consistent predictions across teams while others are systematically misclassified.
This patient-level heterogeneity motivates personalised longitudinal modelling approaches,
such as recurrent architectures conditioned on individual treatment histories~\cite{Rivail2019}.

\subsection{Methodological Trends}

Beyond the quantitative results, the finalists' design choices (summarised in
Section~\ref{sec:sect6} and Supplementary Sections~\nameref{sec:supp_methods_t1}
and~\nameref{sec:supp_methods_t2}) point to a few trends worth interpreting explicitly.
All twelve teams implemented their solutions in PyTorch~\cite{paszke2019pytorchimperativestylehighperformance},
mirroring its dominance in other recent Medical Image Analysis challenges~\cite{ANDREARCZYK2023102972,Nwoye2023,Dorent2023}.
Domain-specific pretraining consistently outperformed generic ImageNet initialisation:
teams that fine-tuned RETFound or another retinal foundation model (FERLIV, MIPLAB, STEP,
OptimaTeam) tended to rank higher and generalise better to Algeria
(Section~\ref{sec:cross_site}) than teams relying solely on natural-image pretraining,
reinforcing the value of large-scale, domain-relevant pretraining data over dataset size
alone. Multi-modal fusion, by contrast, remained rare: MIPLAB was the only team to combine
OCT, localiser images, and clinical variables into a single model, and its consistent
top-two ranking on both sites suggests this is an under-exploited direction rather than a
saturated one. Two teams attempted to directly address Task~2's core information deficit
by predicting the missing future state before classifying it -- DF41 via masked-patch
reconstruction (PPMAE) and jkulinzstudents via latent-embedding matching -- which is
conceptually the closest any submission came to tackling the single-visit limitation
discussed above, though neither out-performed the simpler single-image classifiers overall.
Standardised pretraining resources for ophthalmology, and closer integration of
interpretability tools (saliency maps, attention visualisation) into clinical-facing
pipelines, remain open needs that this challenge did not resolve.

\subsection{Limitations and Future Directions}

The MARIO challenge provided a valuable platform for evaluating cutting-edge deep learning methodologies in medical image analysis. Key trends, such as multi-modal learning, domain-specific pretraining, and generative approaches, emerged as significant drivers of success. However, several challenges remain, including limited adoption of multi-modal strategies, the need for standardized pretraining datasets, and the necessity of improved interpretability for clinical applications.

While the MARIO challenge facilitated significant advancements in medical image analysis, several limitations were observed, which provide critical insights for future iterations of the competition and broader research efforts.

\begin{enumerate}
    \item \textbf{Data Augmentation and Diversity:} Most teams implemented conventional augmentation techniques such as cropping, flipping, and color jittering. However, the use of synthetic data generation techniques, such as generative adversarial networks (GANs) and diffusion models was relatively underexplored. These approaches could play a crucial role in addressing class imbalances and enhancing model robustness by augmenting underrepresented pathological cases.
    
    \item \textbf{Over-reliance on Public Datasets:} A considerable number of teams pretrained their models on ImageNet and other general-purpose datasets, which may not align well with the specialized nature of the challenge. While domain-specific models such as RETFound demonstrated superior performance, the lack of large-scale, publicly available ophthalmic datasets remains a bottleneck. Future work should focus on curating diverse, high-quality datasets for improved model training and evaluation.
    
    \item \textbf{Explainability and Interpretability:} Despite high predictive performance, few models prioritized explainability, which is essential for clinical adoption. Methods such as saliency maps, class activation mappings, and attention mechanisms should be further explored to provide clear justifications for model predictions, thereby enhancing trust among clinicians.
    
    \item \textbf{Generative Techniques:} Teams such as \textbf{jkulinzstudents} and \textbf{MIC Group 6} demonstrated the potential of synthetic scans in Task 2, but broader adoption of generative models could enhance dataset diversity and improve generalizability. Future work should consider exploring diffusion models and variational autoencoders (VAEs) for more effective synthetic data generation.
\end{enumerate}

Beyond addressing these limitations, future editions of the challenge could introduce additional tasks that reflect more clinically relevant scenarios. Some proposed directions include:

\begin{itemize}
    \item \textbf{Integration of Anti-VEGF Treatment Context:} A dedicated task could be introduced to incorporate patient treatment history, specifically anti-VEGF therapy, as an additional predictive variable. This would provide a more comprehensive understanding of disease progression and treatment response, ultimately improving clinical decision support.
    
    \item \textbf{Volume-Based Pathology Prediction:} Instead of limiting pathology predictions to individual slices (B-scan level), future challenges could introduce volumetric predictions (C-scan level). This shift would align more closely with clinical assessment practices, where volumetric changes provide critical insights into disease progression.
    
    \item \textbf{Federated Learning for Privacy-Preserving Model Training:} Given the sensitivity of medical data, future iterations of the challenge could explore federated learning paradigms, enabling models to be trained across multiple institutions without sharing raw patient data. This would facilitate the development of more generalized and privacy-conscious AI models.
    
    \item \textbf{Incorporation of Temporal Analysis:} Longitudinal analysis of disease progression using time-series OCT scans could be introduced as a novel task. Predicting future disease states based on past imaging data would be highly valuable for clinical prognostics and treatment planning.
\end{itemize}

Taken together, these limitations point to a single conclusion: the ceiling on Task~2 is set
less by model architecture than by the information available at a single visit and by the
scarcity of diverse, longitudinally rich training data. The most promising directions above---
incorporating treatment context, moving from the B-scan to the volume level, and modelling the
full temporal trajectory---all add information rather than merely refine the classifier, which
is consistent with the near-identical, chance-level agreement observed across otherwise very
different Task~2 submissions.

In summary, MARIO shows that automated short-term monitoring of neovascular AMD is close to
clinically useful, that single-visit progression prediction is an open problem requiring richer
inputs, and that robustness across populations and devices cannot be inferred from single-site
performance. Future editions that pair these harder tasks with larger, multi-site data will be
well placed to translate the first finding into clinical practice and to make genuine progress
on the second.

\section*{Data Availability}

The dataset has been made publicly available via Zenodo\footnote{\url{https://zenodo.org/records/15270469}}. To ensure consistency and reproducibility, we adhered to the original data split and folder structure that were provided during the challenge phase. Currently, only the dataset from Brest has been released publicly. The Tlemcen dataset, due to privacy and regulatory considerations, remains private and is not available for distribution at this time.

\section*{Ethics}

This study was conducted in accordance with the tenets of the Declaration of Helsinki. All imaging data were de-identified prior to annotation and distribution, and metadata files were stripped of personally identifying information.

\paragraph{Data from Brest, France.} This study follows the French MR-004 methodology (CNIL), designed for research involving the re-use of anonymized data, which requires that the research be demonstrably in the public interest.\footnote{A bilingual article explaining this procedure in more detail is available at \url{https://www.sciencedirect.com/science/article/abs/pii/S2352580023001399}.} All incoming patients sign a written informed consent form authorizing the use of their anonymized data for research purposes. The study has received approval from the Institutional Review Board (IRB) of the French Society of Ophthalmology (Soci\'et\'e Fran\c{c}aise d'Ophtalmologie), IRB~00008855, IRB1.

\paragraph{Data from Tlemcen, Algeria.} As in Brest, all patients in Tlemcen provided written informed consent for the use of their anonymized data for research purposes, and the research adheres to the ethical principles outlined in the Declaration of Helsinki. Approval for this research was granted by the Ethics Committee Board of the Lazouni Ophthalmology Clinic for the use of anonymized OCT images of patients with wet AMD for the research purposes described above. For further inquiries or verification of this ethical approval, contact: Lazouni Ophthalmology Clinic, Imama, Mansourah, 13000, Tlemcen, Algeria; email \texttt{cliniquelazouni@gmail.com}; phone +213-500-568-090.

\paragraph{Data Use Agreement.} Each participating team was required to submit a signed Data Use Agreement (DUA) verifying their institutional affiliation and agreeing to all conditions specified in the Terms of Use.

\section*{Declaration of generative AI in scientific writing}
During the preparation of this work the author(s) used ChatGPT in order to improve readability and language. After using this tool/service, the author(s) reviewed and edited the content as needed and take(s) full responsibility for the content of the publication.

\section*{Acknowledgements}

This work is conducted within the framework of the \textbf{Evired} project, a research initiative funded by the French National Research Agency (\textit{Agence Nationale de la Recherche}, ANR) as part of the \textit{RHU} program. It has received financial support from the French government under the ``\textit{Investissements d'Avenir}'' program, with the reference \textbf{ANR-18-RHUS-0008}.  

We extend our gratitude to the \textbf{CHU of Brest} for their significant contributions to patient care and for their efforts in constructing the dataset. We particularly acknowledge the valuable time and expertise of the medical professionals who meticulously annotated the data, ensuring its quality and reliability.  

Furthermore, we sincerely thank the \textbf{Lazouni Clinic} for granting access to the Algerian dataset and for their essential role in patient care. We also acknowledge the dedicated efforts of the medical team in providing expert annotations, which were instrumental in the success of this study.

\clearpage
\bibliographystyle{model2-names.bst}\biboptions{authoryear}
\bibliography{refs}


\clearpage
\section*{Supplementary Material}
\label{sec:supp}

\setcounter{figure}{0}
\renewcommand{\thefigure}{S\arabic{figure}}

\noindent This supplement collects the full per-team detail underlying the summary
figures of the main text. No additional experiments or results are introduced;
these panels expand the aggregated views already reported in Section~\ref{sec:sect7}.

\begin{figure*}[tp]
  \centering
  \includegraphics[width=\textwidth,height=0.9\textheight,keepaspectratio]{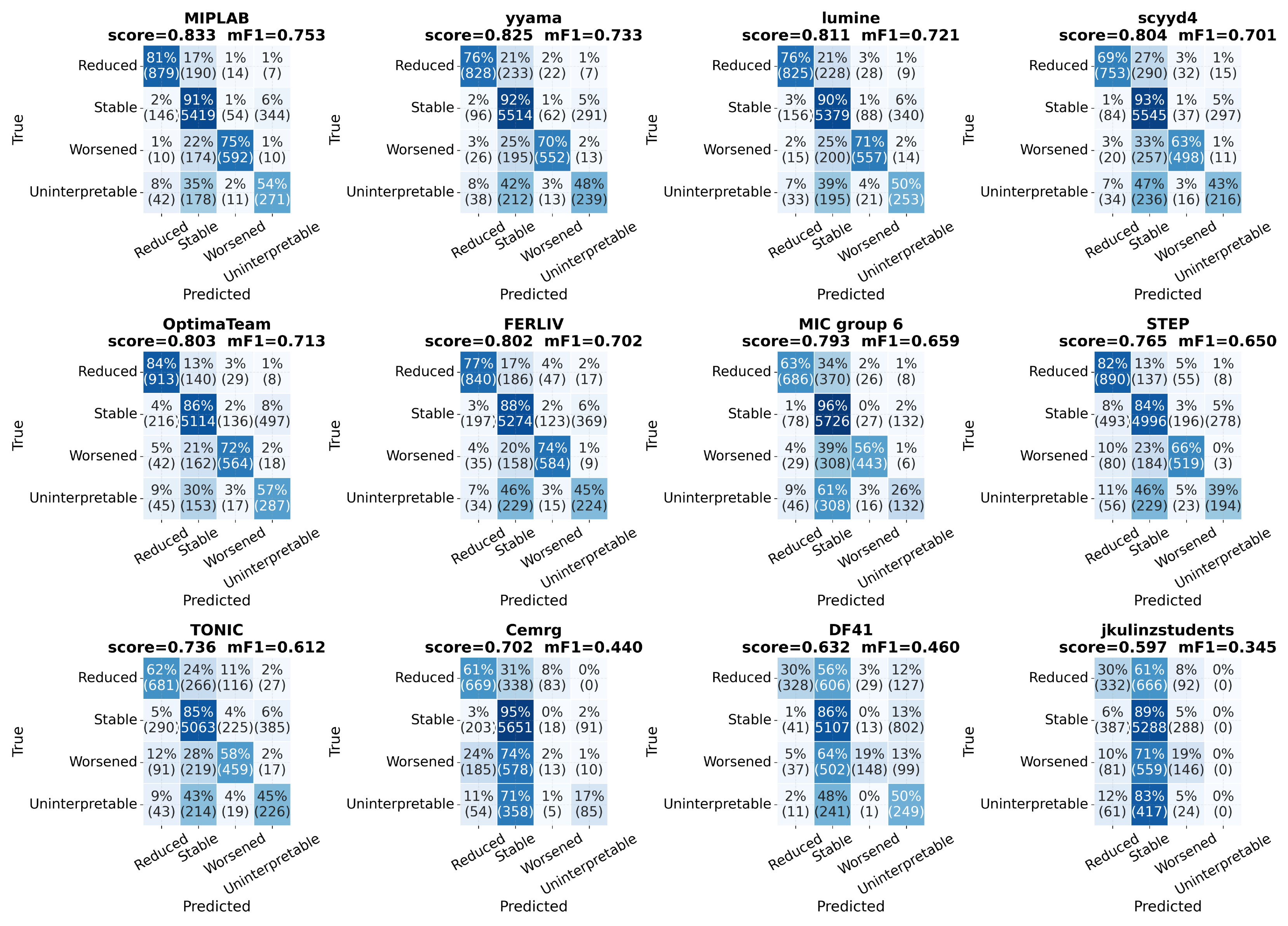}
  \caption{%
    \textbf{All-team confusion matrices -- Brest, Task~1.}
    Row-normalised confusion matrices for all 12 teams on the Brest test set,
    Task~1 (four classes: Reduced, Stable, Worsened, Uninterpretable).
    Teams are sorted in descending order of composite score.
    Cell values show the fraction of true-class samples predicted in each column
    (percentage) with absolute counts in parentheses.
  }
  \label{fig:panel_allcm_brest_t1}
\end{figure*}

\begin{figure*}[tp]
  \centering
  \includegraphics[width=\textwidth,height=0.9\textheight,keepaspectratio]{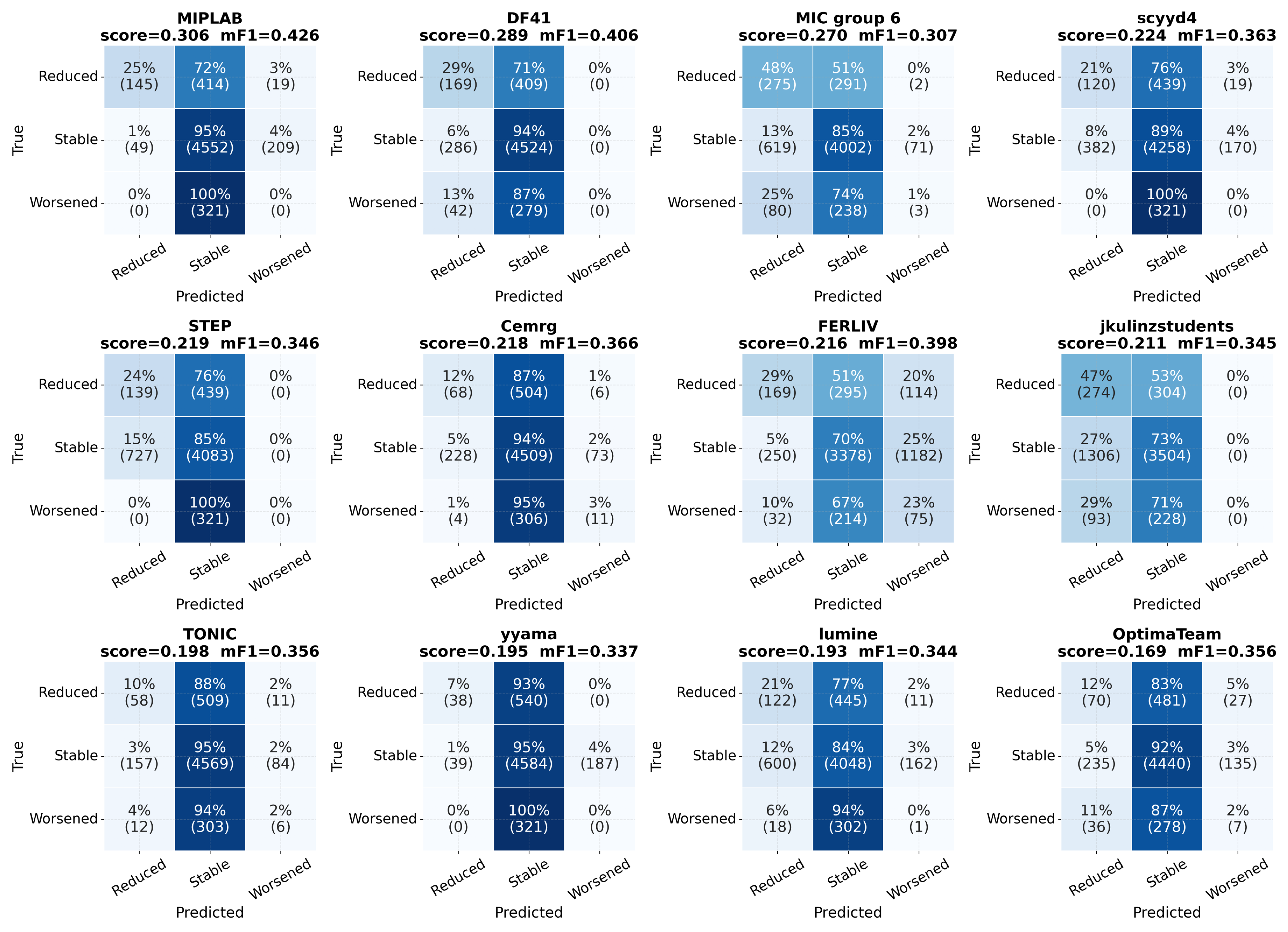}
  \caption{%
    \textbf{All-team confusion matrices -- Brest, Task~2.}
    Row-normalised confusion matrices for all 12 teams on the Brest test set,
    Task~2 (three classes: Reduced, Stable, Worsened).
    Teams are sorted in descending order of composite score.
    Cell values show the fraction of true-class samples predicted in each column
    (percentage) with absolute counts in parentheses.
  }
  \label{fig:panel_allcm_brest_t2}
\end{figure*}

\begin{figure*}[tp]
  \centering
  \includegraphics[width=\textwidth,height=0.9\textheight,keepaspectratio]{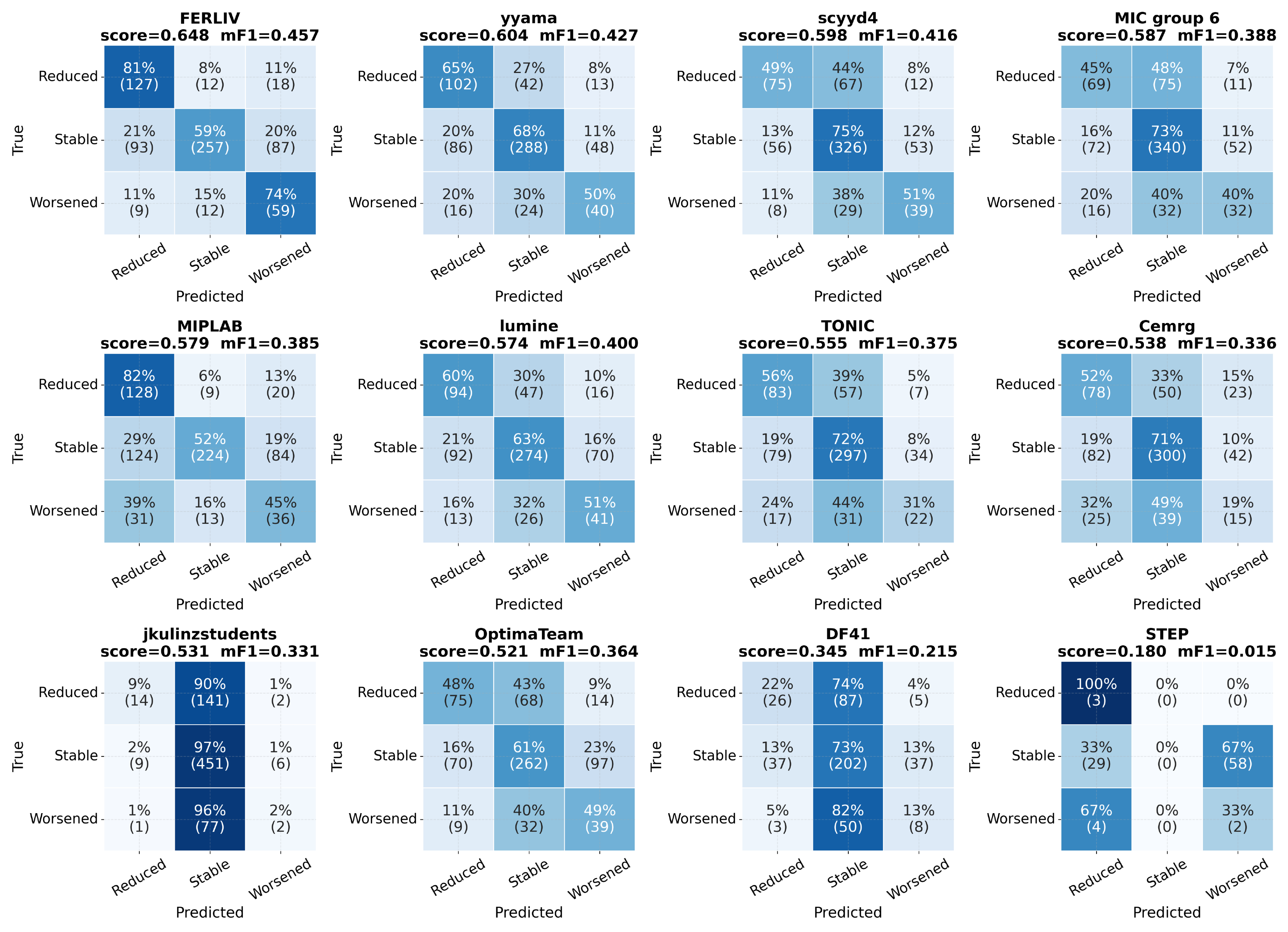}
  \caption{%
    \textbf{All-team confusion matrices -- Algeria, Task~1.}
    Row-normalised confusion matrices for all 12 teams on the Algeria
    (external validation) dataset, Task~1 (three classes, \textit{Uninterpretable}
    absent from this cohort).
    The wider off-diagonal spread relative to Figure~\ref{fig:panel_allcm_brest_t1}
    illustrates the domain-shift challenge faced by all models.
  }
  \label{fig:panel_allcm_algeria_t1}
\end{figure*}

\begin{figure*}[tp]
  \centering
  \includegraphics[width=\textwidth,height=0.9\textheight,keepaspectratio]{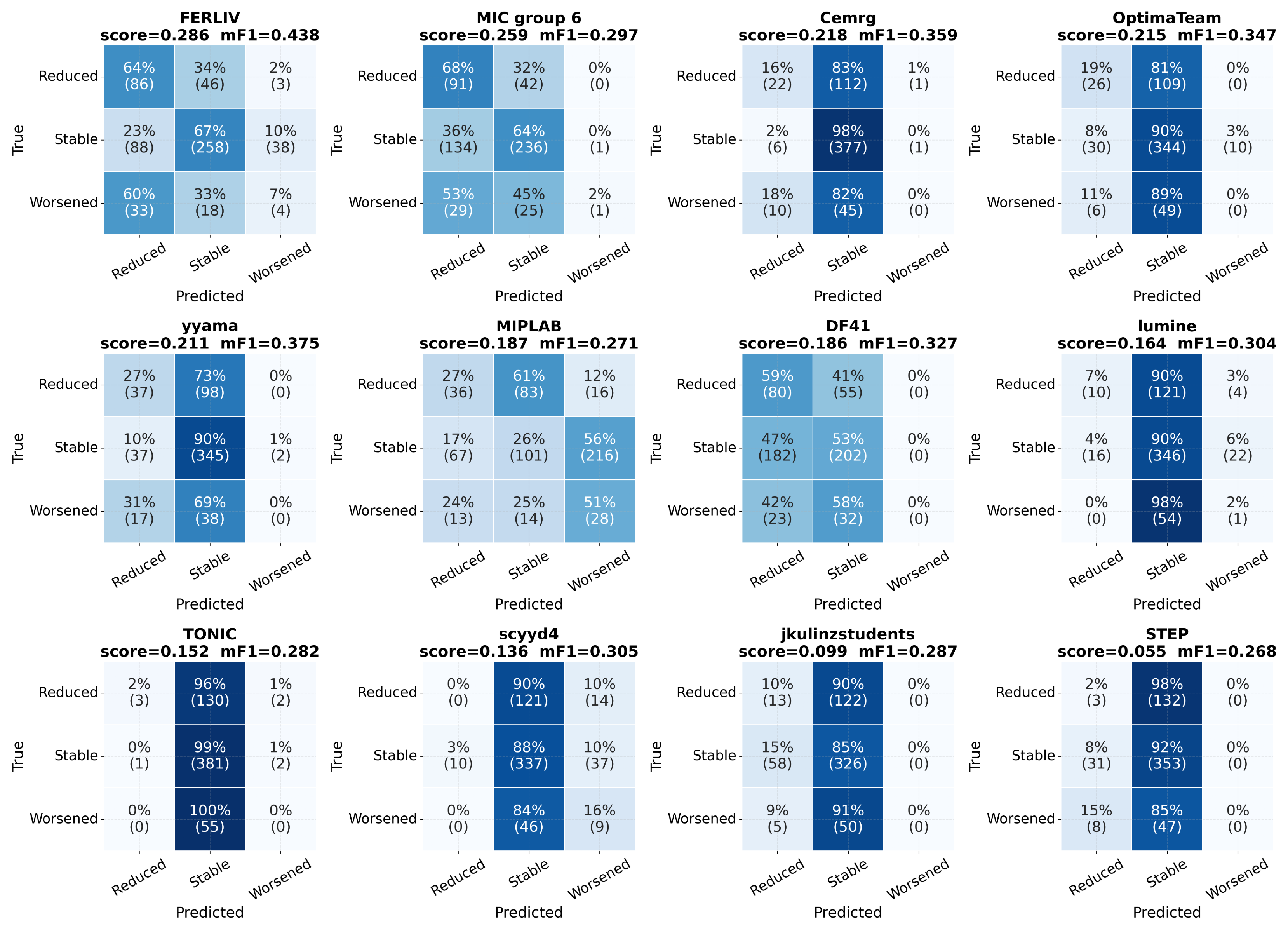}
  \caption{%
    \textbf{All-team confusion matrices -- Algeria, Task~2.}
    Row-normalised confusion matrices for all 12 teams on the Algeria
    (external validation) dataset, Task~2 (three classes).
  }
  \label{fig:panel_allcm_algeria_t2}
\end{figure*}

\begin{figure*}[t]
  \centering
  \includegraphics[width=\textwidth]{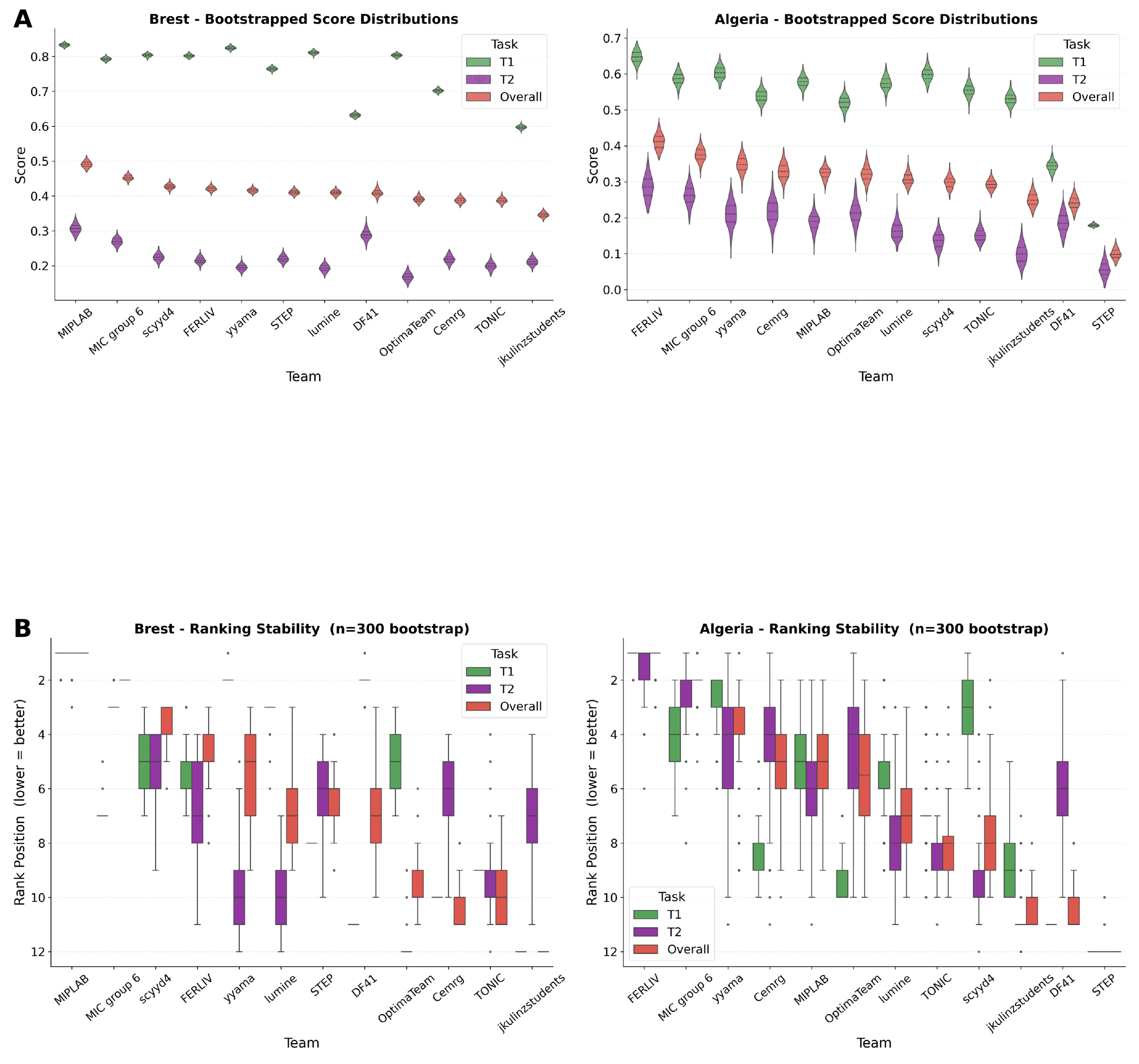}
  \caption{%
    \textbf{Combined bootstrap ranking analysis.}
    \textbf{A:} Bootstrapped composite score distributions ($n=500$ resamples)
    for all 12 teams on both sites, shown as violin plots with quartile markers.
    Colours distinguish Task~1 (blue), Task~2 (orange), and overall (green) scores.
    \textbf{B:} Rank stability under bootstrap resampling shown as box plots.
    Teams are ordered by median overall rank; lower rank position indicates better
    performance.
    Boxes span the interquartile range; whiskers show the 5th--95th percentile.
  }
  \label{fig:panel_bootstrap}
\end{figure*}

\begin{figure*}[t]
  \centering
  \includegraphics[width=\textwidth]{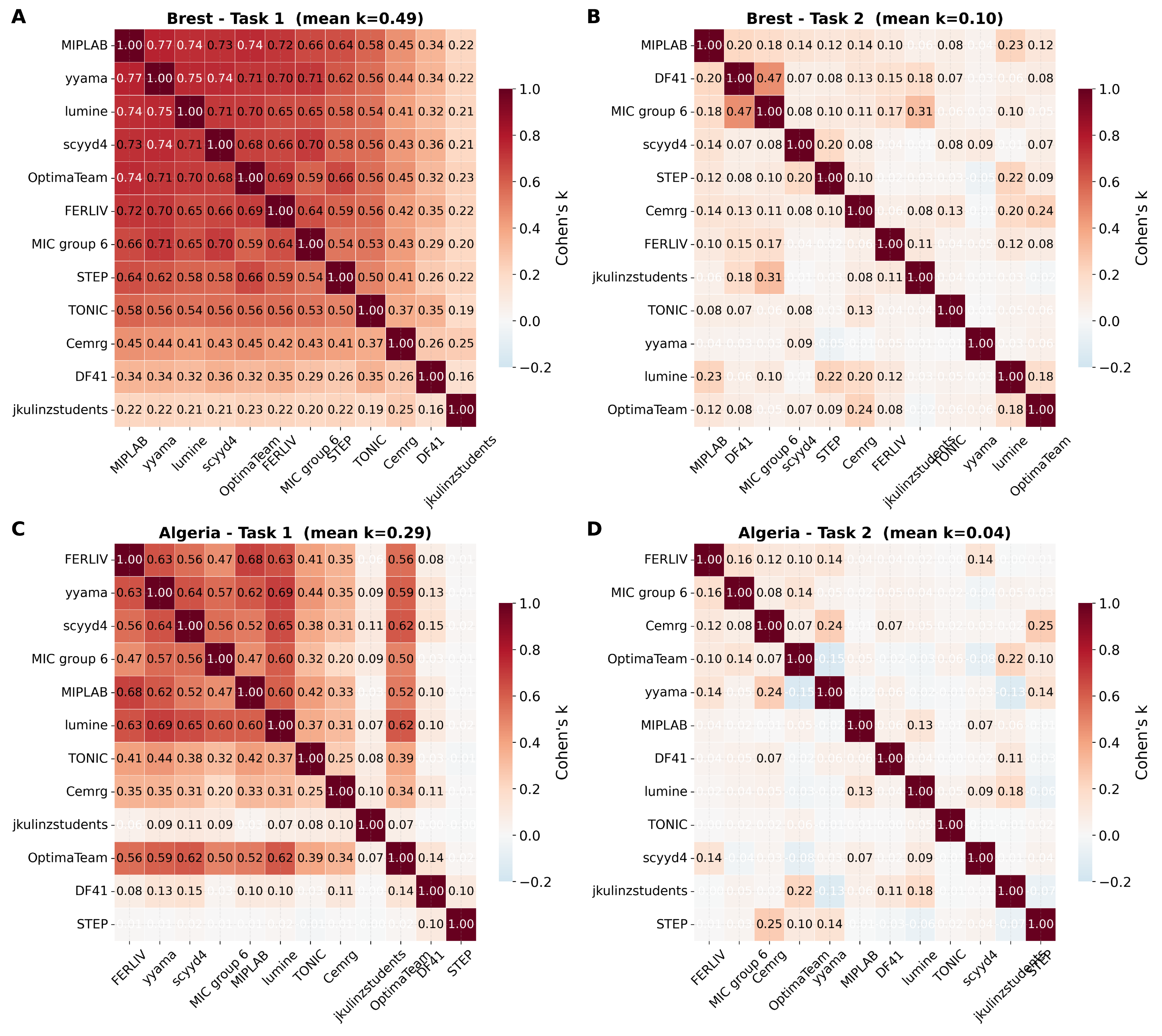}
  \caption{%
    \textbf{Inter-team prediction agreement.}
    Symmetric Cohen's $\kappa$ heatmaps for all 66 pairwise team combinations
    across all four site\,$\times$\,task configurations.
    \textbf{A:} Brest Task~1.  \textbf{B:} Brest Task~2.
    \textbf{C:} Algeria Task~1.  \textbf{D:} Algeria Task~2.
    Green = strong agreement ($\kappa\!\to\!1$); red = systematic disagreement
    ($\kappa\!\to\!-0.2$).  Mean $\kappa$ per panel is shown in the subplot title.
  }
  \label{fig:panel_agreement}
\end{figure*}

\begin{figure*}[t]
  \centering
  \includegraphics[width=\textwidth]{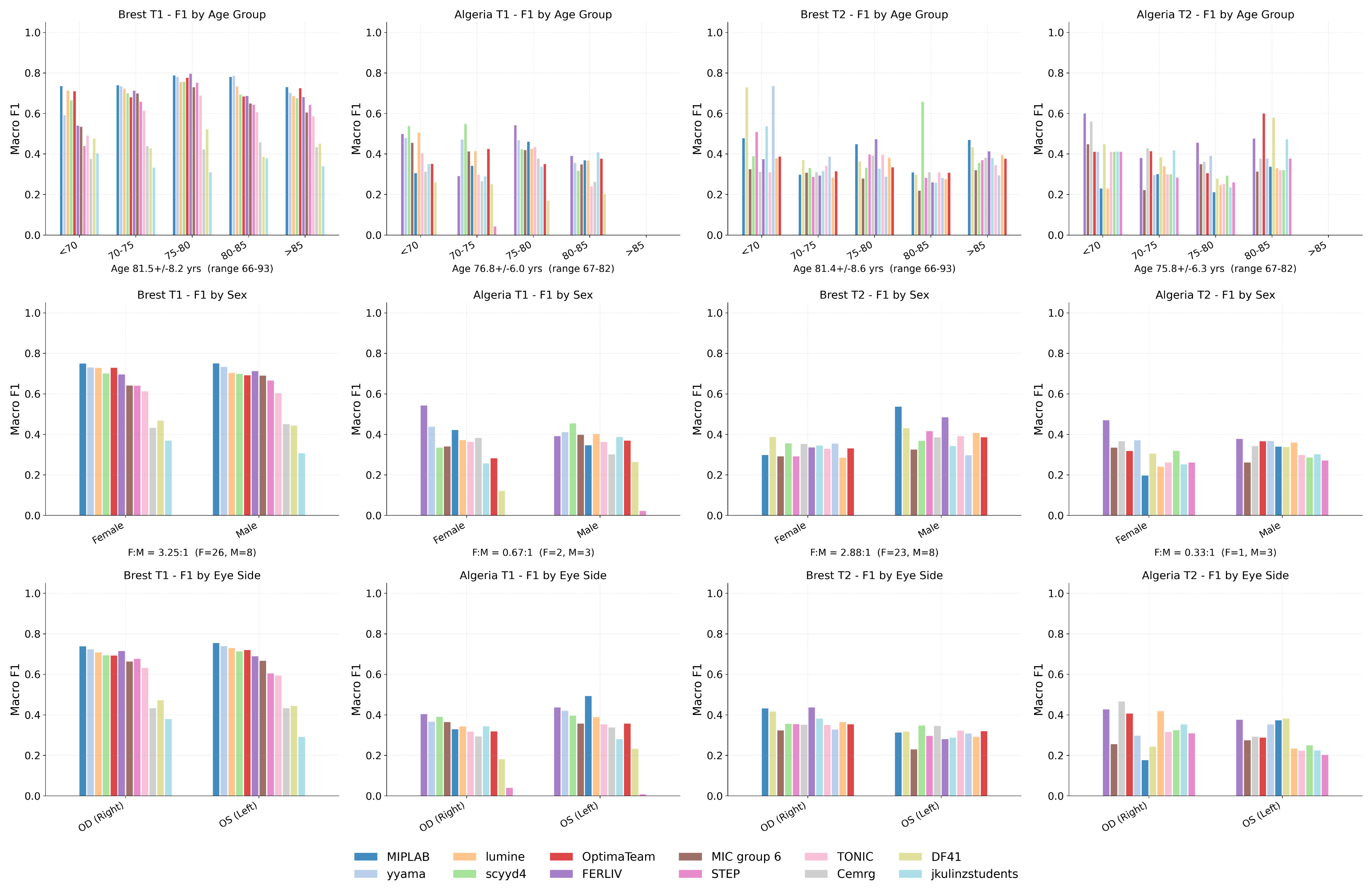}
  \caption{%
    \textbf{Demographic stratification.}
    Performance stratified by demographic variables
    (age group, sex, laterality) for Task~1 and Task~2 on both sites.
  }
  \label{fig:panel_clinical}
\end{figure*}

\clearpage
\section*{Supplementary Note: Per-Team Method Descriptions}

\noindent This note provides the full narrative description of each finalist team's
methodology for Task~1 (Section~\nameref{sec:supp_methods_t1}) and Task~2
(Section~\nameref{sec:supp_methods_t2}), summarised in the main text in
Section~\ref{sec:sect6} and Figures~\ref{fig:description_method_task1}
and~\ref{fig:description_method_task2}.

\subsection*{Task 1 -- Full Method Descriptions}
\label{sec:supp_methods_t1}

\subsubsection{lumine - Team Summary -- Task 1}

The Lumine team developed a deep learning model designed to classify progression in Age-related Macular Degeneration (AMD) using OCT images captured over two consecutive sessions. The preprocessing methods include resize and Z-score normalization, each image was resized to 224x224 pixels to match the pre-trained network. To address data imbalance, the team utilized data augmentation methods including ColorJitter, GaussianBlur, RandomAffine, HorizontalFlip, and GaussianNoise. These methods are not only used during training but also before training to generate different quantities of images for different classes to mitigate data imbalance.
For Task 1, a Siamese network was utilized, with both branches consisting of a ConvNeXt\_large feature extractor to process each image in a pair simultaneously. The model also incorporated a multi-head attention mechanism to better extract features. Final classification is derived by combining the extracted features and passing them through a linear classifier with ReLU activation. The training utilized the AdamW optimizer with learning rate decay, and cross-entropy loss with a balanced weighting scheme to counteract the class imbalance. Several models using different hyperparameters are trained and used to classify the image. Results produced by these networks are combined and weighted to produce the final classification result.

\subsubsection{yyama - Team Summary -- Task 1}

Yamagishi presented an innovative approach using MaxVit Tiny architecture for AMD progression classification between consecutive OCT scans. The preprocessing strategy uniquely mimicked clinical practice by vertically concatenating pairs of consecutive OCT images (each resized to 512x256 pixels) into a single 512x512 pixel image, effectively simulating the side-by-side comparison method used by clinicians. This fusion approach aimed to enhance the model's ability to detect temporal changes between scans.
The model architecture leveraged MaxVit Tiny's hybrid design, combining the strengths of CNNs for local feature detection with Transformers for capturing global dependencies. Training utilized a 5-fold StratifiedGroupKFold cross-validation strategy with patient-wise grouping to prevent data leakage. The training process incorporated comprehensive data augmentation techniques including random resized crops, horizontal flips, rotations, and coarse dropout. The model was optimized using Adam optimizer with cosine learning rate scheduling over 5 epochs. For inference, the team implemented both ensemble prediction across all 5 folds and test-time augmentation with horizontal flips.

\subsubsection{Cemrg - Team Summary -- Task 1}

They employed transfer learning by initializing MobileNetV3 with weights pretrained on ImageNet and fine-tuning it on the Task 1 dataset. To mitigate catastrophic forgetting of the pretrained features, they fine-tuned the deeper layers---containing more task-specific information---using a lower learning rate, while adapting the final layers specifically to the Task 1 data.

To address the limitations of MobileNetV3, they applied Test-Time Augmentation (TTA), generating multiple transformed versions of each image during inference. The resulting predictions were merged to enhance the model's confidence and robustness.

They further trained five different variations of MobileNetV3 models using 5-fold cross-validation, varying initialization strategies, hyperparameters, and data augmentation techniques. This promoted diverse feature learning and resulted in a more generalized final model. The ensemble outputs were aggregated either by averaging or using a weighted sum for the final prediction.

An ablation study was performed by comparing the performance of $MobileNetV3\_base$ (without ensembling or TTA) against $MobileNetV3\_ensemble$ (with ensembling and TTA).

\subsubsection{OptimaTeam - Team Summary -- Task 1}

SiamRETFound is a Siamese neural network designed for longitudinal change detection in retinal images. The input images are preprocessed through intensity normalization and resized to 224x224 pixels. During the training phase, data augmentation techniques such as rotation and horizontal flipping are applied to introduce variability and improve robustness. The network utilizes a large Vision Transformer architecture, initialized with pretrained weights from the RETFound model. As a pretext task, SiamRETFound is first trained on the Kermany dataset for a simulated \textit{change/no-change} detection task, with labels inferred from the disease classes: if both images have the same disease, the pair is assigned to \textit{no-change}, otherwise to \textit{change}. It is then fine-tuned on the MARIO training dataset for target class classification. SiamRETFound employs a late fusion approach, independently extracting feature representations from each input B-scan and concatenating them for the final prediction. Ultimately, five SiamRETFound-based models are trained with variations in training settings, including different optimizers, learning rates, and augmentation schemes. In parallel, another model based on the Shifted WINdows (SWIN) Transformer architecture is trained. This model integrates multiple pretext tasks, such as multi-class B-scan classification and biomarker detection as a multi-label task. Similar to SiamRETFound, the SWIN model uses a late fusion approach for change detection. Input images are preprocessed with intensity normalization and resizing, followed by extensive data augmentation, including brightness adjustment, blurring, salt-and-pepper noise, rotation, and random eraser. To address class imbalance, the SWIN model employs the focal loss function, which assigns greater penalties to misclassified examples. SWIN models are trained using a five-fold cross-validation framework. During post-processing, predictions for a test image are obtained by averaging class probabilities across all ten models. The final classification is determined by selecting the class with the highest mean score.

\subsubsection{TONIC - Team Summary -- Task 1}

During the development phase, they explored four different types of models for Task 1: a baseline model and three variations. \textbf{Baseline model}: The architecture of the Baseline Model is built using two pre-trained ResNet18 models that process different input data streams. The first model gets as input the images taken on t=0, the second model gets as input all the images from t=1. After training, the outcome features from both models are then combined and fed into a new fully connected layer that makes the final classification.\textbf{ Extra Layer}: To improve the model, they added an extra layer to the baseline model after combining the two ResNet models. They expect that this will improve the model because this extra layer could make connections between the two models and learn from the feature interaction. \textbf{Balanced Batches}: To optimize the model in a different way, they focused on the unequal distribution of classes detailed in Section~\ref{sec:sect5}. To make sure that the model learns about every class equally, they trained a model with balanced batches: ensuring that each batch of data used during training contains an equal or proportionate number of samples from each class. Balanced batches are achieved by iterating through separate data loaders for each class. During each iteration, a batch is drawn from each loader, and then the samples are concatenated together to form a single batch that contains an equal number of samples from each class. This approach helps the model to learn equally from all classes, preventing it from becoming biased towards the more frequent `stable' class. \textbf{Balanced Batches and Augmented Data}: Some data augmentation was applied to artificially expand the diversity of the training dataset by applying various transformations. Data augmentation was implemented through a series of image transformations, including random horizontal and vertical flips, rotations, brightness/contrast (with PyTorch ColorJitter), and resized cropping. The datasets are then combined to create a comprehensive training set that includes both the original and augmented images. They combined this method with their balanced-batches approach. For Task 1, they generated 15,000 samples per class. Thus, they end up with a dataset that is four times as big as the original dataset of which the classes are balanced.

\subsubsection{FERLIV - Team Summary -- Task 1}

The FERLIV team proposed a modular, fully transformer-based approach for monitoring the progression of wet AMD using two registered OCT B-scans from two consecutive medical exams. The method consists of three transformer encoder architectures: Feature Encoder, Change Encoder, and Diagnosis Encoder. The Feature Encoder is the first part of the method, where a large Vision Transformer \cite{dosovitskiy2021image} encodes a pair of OCT B-scans into two sets of local feature vectors. The Change Encoder detects visual changes in a retina using a dual Multi-Change Captioning Transformer \cite{qiu2021describing} with a co-attention mechanism \cite{lu2019vilbert}. The outputs of the Change Encoder are concatenated based on their corresponding 2D positions and forwarded to the third and final transformer encoder, i.e., the Diagnosis Encoder, which quantifies changes in disease progression. The Diagnosis Encoder uses a self-attention mechanism and is followed by the classification head.
During preprocessing, OCT B-scans are resized and cropped to $224 \times 224$ pixels, followed by normalization.
Before training, the Feature Encoder is initialized with weights from the pretext task, where RETFound's Vit-Large model \cite{zhou2023foundation} is used as a Segmenter encoder \cite{strudel2021segmenter} for the segmentation of retinal layers on the OCT MS and Healthy Control dataset \cite{he2019retinal}. This public dataset is acquired on the same device as the MARIO dataset, addressing variability caused by inter-device transfer. Pretext task results in improved performance for Task 1, as changes in retinal structures are crucial for monitoring disease progression. All three encoders are trained using the AdamW optimizer with cosine learning rate scheduling and weight decay. A weighted categorical cross-entropy loss function is used to address the class imbalance. Augmentation techniques include random resized cropping and random horizontal flipping to improve generalization.

\subsubsection{MIPLAB Team Summary -- Task 1}

The MIPLAB approach for Task~1 integrates all available data modalities, including OCT B-scans, infrared localizer images, and clinical variables, to classify changes in neovascular activity between two consecutive time points. The team began by preprocessing the image data (resizing OCT B-scans to 496$\times$496 pixels and infrared localizer images to 384$\times$384 pixels) and standardizing clinical variables using z-score normalization (subtracting the mean and dividing by the standard deviation). They finetuned RETFound (a ViT-large based foundation model) and EfficientNetV2 on the MARIO training set using cross-entropy loss, and then used these finetuned models to extract feature representations from the imaging data. These representations are combined with the patient's clinical data to create a unified feature vector for each patient on a slice-by-slice basis. To incorporate volumetric context, global average pooled features of all B-scans within a 3D OCT C-scan are also added to the feature vectors. Principal component analysis (PCA) is used to reduce the dimensionality of the feature space and to mitigate the risk of overfitting. A support vector classification (SVC) model with hinge loss serves as the final classifier. Random data augmentations (horizontal and vertical flips, rotations, translations, and contrast adjustments) were employed to enhance model generalizability. The team further employed a semi-supervised strategy by assigning pseudo labels to unlabelled validation data with a prediction probability above a pre-specified confidence threshold. The SVC model was then retrained on this enriched dataset.

\subsubsection{MIC Group 6 - Team Summary -- Task 1}

In this study, a robust framework was developed to classify changes in AMD progression between two time-point OCT scans. The architecture centered on a Siamese network, specifically designed to harness the temporal information encoded in paired OCT scans. Preprocessing steps included intensity normalization, resizing all images to a standardized resolution, and ensuring precise anatomical alignment between scans for consistency. Extensive experimentation revealed that the best-performing approach for Task 1 utilized a Siamese network with the ImageNet encoder. This encoder efficiently extracted feature representations from both time points, which were subsequently combined through element-wise subtraction to emphasize temporal changes. The resulting feature vector was fed into fully connected classification layers to predict progression states (reduced, increased, stable, or uninterpretable).

\subsubsection{STEP - Team Summary -- Task 1}

The STEP team developed a novel deep learning model, designed to classify the evolution between consecutive OCT B-scans, using contextual information from adjacent B-scans part of an OCT volume and at two time instants. The preprocessing workflow involved constructing OCT volumes from grouped B-scans, applying intensity normalization using computed mean and standard deviation, and resizing each B-scan to a fixed size of 224x224 pixels for consistency. To address dataset variability, the team implemented targeted data augmentation using the MONAI framework during training, applying random flip in all dimensions, random intensity scaling and random Gaussian noise addition to enhance robustness against overfitting. For imbalanced classes, volume-level augmentation was carefully synchronized across paired scans.
The STEP team's architecture combined a pretrained vision transformer (ViT) backbone, RETFound, with a bidirectional cross-attention module designed to capture dependencies between sequential B-scan pairs. CLS tokens extracted from the B-scans were processed to compute temporal relationships, with cross-attention improving the model's sensitivity to subtle changes in progression. A linear layer generated individual predictions for each B-scan. The training pipeline employed the AdamW optimizer with a weight decay of 0.05 with cosine learning rate scheduling and 10 warm-up epochs to ensure smooth convergence. Evaluation metrics guided model selection, with predictions aggregated using batched inference for OCT volumes. Post-processing involved assigning predictions slice-by-slice, leveraging the context learned by the model. This approach demonstrated consistent classification improvements across metrics, highlighting the efficacy of integrating sequential context and advanced transformer-based methods.

\subsubsection{DF41 - Team Summary -- Task 1}

For Task 1, the DF41 team used a Late Fusion CNN architecture to classify progression changes between two consecutive retinal OCT scans. The preprocessing pipeline leveraged the OCT image segmentation library Optical Coherence Tomography Image Preprocessing (OCTIP). Two segmentation models, FPN-EfficientNet-B6 and FPN-EfficientNet-B7, were employed to generate segmentation masks. The median of the outputs from these models was applied to ensure robust predictions. Once the segmented regions were extracted, the upper boundary of the retina (inner limiting membrane) was aligned with the top of the image, effectively "flattening" the retina and realigning the scans along the depth axis. This preprocessing step is crucial for improving analysis quality and enhancing classification performance by eliminating noise and irrelevant information. The output of this step was an image with dimensions of \( 512 \times 200 \) pixels. Next, the Late Fusion CNN network employed the ResNet50 architecture to extract two feature maps of size 2048 from the paired OCT images without resizing, ensuring that no information was lost. These feature maps were concatenated into a single map of size 4096, which was then passed through a fully connected layer to produce the classification output. To enhance the diversity of the training dataset and improve model robustness, various data augmentation techniques were applied, including RandomHorizontalFlip, RandomVerticalFlip, RandomRotation, ColorJitter, RandomPerspective, and GaussianBlur. To address class imbalance and further boost performance, an ensemble method was used. Four models were trained through cross-validation, and the predictions from all models were averaged during post-processing. These combined strategies contributed to strong performance in classifying the progression between two consecutive OCT slices. In particular, OCTIP showed improved performance with enhanced F1 scores and rank correlation.

\subsubsection{scyyd4 - Team Summary -- Task 1}

The scyyd4 team proposed a novel AI framework, OCT-DiffNet, designed for detecting neovascular activity in Age-related Macular Degeneration (AMD) using sequential OCT images. The preprocessing pipeline
included cropping, resizing each image to \( 224 \times 224 \) pixels, and intensity normalization to enhance the input consistency. To further augment the training data and improve model generalization, the team employed
targeted data augmentation techniques, such as random rotations, zoom, and brightness adjustments, simulating realistic variations in the OCT imaging process.
OCT-DiffNet is built upon a modified ConvNeXt V2-Large architecture, which was fine-tuned for this
task. The model processes pairs of consecutive OCT images and computes temporal differences using a
Siamese network structure. Key architectural modifications included adjusting the first convolutional layer
to accommodate single-channel (grayscale) OCT images and introducing custom fully connected (FC) layers
to analyze the extracted features and their differences. These enhancements allow the model to effectively
capture subtle changes indicative of disease progression.
The training pipeline utilized a cross-entropy loss function, optimized with the Adam optimizer. To
further boost prediction accuracy and robustness, an ensemble method was employed, combining predictions
from five models trained with 5-fold cross-validation. Post-processing involved majority voting across the
ensemble predictions, ensuring stable and reliable outputs.
The proposed method achieved significant improvements in performance metrics, demonstrating its potential for precise AMD progression detection and clinical applicability.

\subsubsection{jkulinzstudents - Team Summary -- Task 1}

This work proposes a ViT-based deep learning approach to classify progression changes between consecutive retinal OCT B-scans. The preprocessing pipeline includes resizing all images to \( 224 \times 224 \) pixels and intensity normalization to enhance feature consistency. Data augmentation techniques, such as random flipping, zooming, cropping, rotation, and noise injection, are applied to improve robustness. The method employs a pretrained DinoV2 ViT model with registers to extract feature embeddings from each B-scan. These features are concatenated and passed through a two-layer MLP to generate classification predictions for disease progression. A 10-fold cross-validation strategy ensures reliable performance by varying stratification seeds to create diverse patient splits. The final prediction is obtained by aggregating softmax scores across all folds. This approach effectively captures subtle changes in neovascular activity and demonstrates strong performance in classification metrics.

\subsection*{Task 2 -- Full Method Descriptions}
\label{sec:supp_methods_t2}

\subsubsection{lumine - Team Summary -- Task 2}

For Task 2, they utilized a standalone ConvNeXt\_large model to predict disease progression based on OCT scans taken three months apart. The preprocessing methods are also resize and Z-score normalization. To ensure the model was exposed to a wide variety of augmented samples, they also used data augmentation methods similar to those in the previous task. However, this process was applied only during training.
The model architecture contains a ConvNeXt\_large backbone with a multi-head attention mechanism to capture relevant features. The network was trained using the AdamW optimizer, combined with learning rate decay, and used cross-entropy as the loss function, with class-specific weights assigned to improve the prediction of minority classes. For post\-processing, they adjusted labels based on the proportion of instances belonging to a particular category within the same localizer, improving consistency across related data points.

\subsubsection{yyama - Team Summary -- Task 2}

For the three-month progression prediction task, Yamagishi developed a multimodal approach combining EfficientNet V2 S for image processing with a Multi-Layer Perceptron for handling patient metadata. The preprocessing pipeline was enhanced to include both OCT and Localizer images, concatenated vertically into a 512x512 pixel image. Patient metadata (age, gender, number of visits) was normalized and processed through a separate network branch before fusion with image features.
The architecture consisted of three main components: an EfficientNet V2 S image encoder, a metadata encoder with two Dense layers, and a feature fusion classifier. The training process maintained similar augmentation and optimization strategies to Task 1, but utilized single-fold prediction with test-time augmentation for inference. The team's implementation demonstrated how multimodal fusion of image data with patient metadata could be leveraged for long-term disease progression prediction.

\subsubsection{Cemrg - Team Summary -- Task 2}

The approach for Task 2 mirrored that of Task 1. They initialized MobileNetV3 with ImageNet-pretrained weights and fine-tuned it on the Task 2 dataset, carefully adjusting deeper layers at a lower learning rate to prevent catastrophic forgetting and adapting the final layers to the new task.

They again applied Test-Time Augmentation (TTA) to strengthen model robustness by merging predictions from multiple augmented versions of the input image. Five different MobileNetV3 models were trained with 5-fold cross-validation, each varying in initialization, hyperparameters, or data augmentation strategies, leading to diverse representations and a more generalized model.

Ensemble outputs were combined through averaging or a weighted sum. An ablation study comparing $MobileNetV3\_base$ and $MobileNetV3\_ensemble$ models was conducted to highlight the benefits of ensembling and TTA.

\subsubsection{OptimaTeam - Team Summary -- Task 2}

In Task 2, the goal is to predict the change in AMD patients within a three-months window. For preprocessing, a dataset level mean and variance are calculated to normalize the pixel values, then all images are resized to 224x224. The ViT16 model is initially pretrained with Masked Auto Encoder. In the fine-tuning step, 3-fold cross-validation is used for training 3 different models. In order to address class imbalance, focal loss is employed to put more importance to the minority classes. Additionally, the change labels are ordinal, meaning that there is an inherent ordering between the classes. a discrete Wasserstein-2 loss is added to exploit the ordinality. Finally a majority voting between these 3 models is applied to get a single prediction.

\subsubsection{TONIC - Team Summary -- Task 2}

Task two has similar approaches as Task 1. First, the baseline model is a pre-trained ResNet18 model that was trained on 70\% of the training data. Second, the same model is retrained with balanced batches, to exclude the effect of the class distribution; similar to Task 1. Lastly, the balanced batches model was extended with augmented data by \textit{torchvision.transforms}; also similar to Task 1.

\subsubsection{FERLIV - Team Summary -- Task 2}

For Task 2, the FERLIV team utilizes the modular approach from Task 1, modifying the three-part method by removing the middle component, i.e., the Change Encoder, from the pipeline. The Feature Encoder and Diagnosis Encoder are initialized with weights from Task 1, enabling knowledge transfer from the simpler task to the more challenging one. In Task 2, the Feature Encoder extracts local feature vectors from a single OCT B-scan, while the Diagnosis Encoder is followed by a new classification head. Preprocessing, augmentation techniques, and most training details remain the same as in Task 1.
To better address the more apparent class imbalance in Task 2, weighted random sampling is introduced in addition to the weighted categorical cross-entropy loss function.

\subsubsection{MIPLAB Team Summary -- Task 2}

For Task 2, the team utilized all available data modalities, including OCT B-scans, infrared localizer images, and clinical variables, to predict three-month progression in neovascular activity from a single time point. The image data were preprocessed by resizing OCT B-scans to 496$\times$496 pixels and infrared localizer images to 384$\times$384 pixels. Clinical variables were standardized using z-score normalization (subtracting the mean and dividing by the standard deviation). The team finetuned RETFound and EfficientNetV2 on the MARIO training set using cross-entropy loss, and subsequently utilized these models to extract feature representations from the imaging data. These representations are combined with the patient's clinical data to create a unified feature vector for each patient on a slice-by-slice basis. To incorporate volumetric context, global average pooled features from all B-scans within a 3D OCT C-scan are added to the feature vectors. PCA is employed to reduce the dimensionality of the feature space and mitigate the risk of overfitting. An ordinal logistic classification model with an immediate-threshold loss variant is trained to directly predict graded progression outcomes. To improve model generalizability, random data augmentations were applied, including horizontal and vertical flips, rotations, translations, and contrast adjustments.

\subsubsection{MIC Group 6 - Team Summary -- Task 2}

A novel hybrid framework was designed to predict AMD progression from a single OCT scan, building upon the findings from Task 1. The preprocessing pipeline mirrored that of Task 1, ensuring consistency across both tasks. Additionally, the fine-tuned encoder from Task 1 was adapted for use in Task 2. Although extensive evaluation on more complex models was performed, the Siamese Network with ImageNet encoder emerged as the top-performing model. It demonstrated superior capability in extracting detailed features from single scans, making it well-suited for capturing subtle disease-specific patterns necessary for accurate progression prediction.

\subsubsection{STEP - Team Summary -- Task 2}

For Task 2, the STEP team designed a specialized Multiple Instance Learning (MIL)-based architecture to identify the most significant slices within an OCT volume to predict disease progression within a 3-month period. The preprocessing pipeline involved constructing fixed-size OCT volumes (25 B-scans each) by selecting central slices and replicating boundary slices for smaller volumes. Intensity normalization was applied based on mean and standard deviation, ensuring consistent input data. To address variability, targeted data augmentation was employed using the MONAI framework, incorporating random flip in all dimensions, random intensity scaling and random Gaussian noise addition transformations. The proposed solution utilized a vision transformer (ViT) backbone, RETFound, pre-trained on retinal imaging tasks. This backbone extracted feature vectors from each B-scan, which were then processed by a MIL attention module. The attention mechanism assigned importance scores to slices within the volume, highlighting regions with critical biomarkers such as retinal fluid. A final linear layer aggregated these features to predict progression of activity for the entire volume. The training process used the AdamW optimizer with a weight decay of 0.05, cosine learning rate scheduling and class-weighted sampling to counter data imbalance. During inference, predictions were made at the volume level, with results extended to the original B-scans.

\subsubsection{DF41 - Team Summary -- Task 2}

For Task 2, DF41 introduced a novel method called Patch Progression Masked Autoencoder (PPMAE), which predicts a future OCT image based on the current scan, utilizing the dataset from Task 1. They approach this task by reconstructing the future state of the current image and then performing disease progression classification. The PPMAE model works by masking 75\% of the current OCT image at time \( t_0 \) and predicting the corresponding patches from the \( t_1 \) image, allowing the model to capture temporal changes and disease progression between the patches. After predicting the future patches, they are aligned with the unmasked regions of the \( t_0 \) image, resulting in a reconstructed future OCT image that corresponds to the image several months later. During training, RandomResizedCrop and RandomHorizontalFlip augmentations were applied, and the reconstruction error was evaluated using Mean Squared Error (MSE) between the predicted and actual patches at \( t_0 \). OCTIP preprocessing was also used for this reconstruction task, improving the quality of the image reconstruction. Finally, the model from Task 1 was re-trained to predict AMD progression by using both the current OCT image and the predicted future image.

\subsubsection{scyyd4 - Team Summary -- Task 2}

The scyyd4 team presented an innovative solution for predicting disease progression in Age-related Macular Degeneration (AMD) over a three-month period, leveraging a multi-instance learning framework, CLAM- SB, combined with the ConvNeXt V2 feature extractor. Preprocessing steps included converting OCT images to grayscale, resizing them to \( 224 \times 224 \) pixels, and normalizing pixel intensities based on dataset-specific mean and standard deviation values to enhance consistency. A weighted random sampler was employed to address the class imbalance, ensuring proportional representation of each class during training. The core architecture utilized ConvNeXt V2 as a frozen feature extractor to generate robust image embeddings, which were fed into the CLAM-SB model. This multi-instance learning approach aggregated
features across multiple OCT scans (bags), enabling the model to capture nuanced patterns indicative of
disease progression. Custom fully connected layers within the CLAM-SB framework were optimized to
classify cases into three progression categories: no progression, mild progression, and significant progression. Additionally, a gated attention mechanism in CLAM-SB highlighted the most relevant instances within each bag, improving interpretability and prediction accuracy.
The training pipeline employed a weighted cross-entropy loss function to handle class imbalance and was
optimized using the Adam optimizer with a learning rate scheduler. Dropout regularization was excluded
to preserve model capacity, while data augmentation was omitted following preliminary experiments that
revealed augmentation-induced instability. The final model ensemble was built using predictions from five
models trained with different random seeds, ensuring stable and reliable outputs. Post-processing involved
majority voting across bags, enhancing prediction consistency at the patient level.

\subsubsection{jkulinzstudents - Team Summary -- Task 2}

The jkulinzstudents team employed a two-step deep learning approach to predict disease progression over a 90-day period using a single OCT B-scan. The preprocessing pipeline involved resizing each OCT scan to \( 224 \times 224 \) pixels, aligning with the input requirements of their feature extraction backbone, RETFound. The images were then normalized using standard ImageNet statistics to ensure consistency across the dataset.

Their approach was divided into two stages: Latent Matching and Disease Progression Classification. In the first stage, the Latent Matching model estimated the future state of the disease by predicting the embedding of the OCT scan at \( t_{90} \), leveraging the frozen feature extractor of RETFound. This step used a three-layer MLP trained with negative cosine similarity loss, ensuring that the predicted embedding closely aligned with the true future representation. The Adam optimizer was employed with an initial learning rate of \( 1e^{-3} \), reduced adaptively when training plateaued.

In the second stage, the Disease Progression Classification model utilized the embeddings from \( t_0 \) and the predicted embeddings at \( t_{90} \) to classify the disease progression into three categories: Reduced, Stable, or Increased activity. The concatenated embeddings were passed through another three-layer MLP, optimized with cross-entropy loss. This stage was trained using the available dataset from Task 1 to ensure sufficient data coverage for all progression classes.

To further improve performance, They experimented with fine-tuning the RETFound model on the provided OCT images using a masked image modeling approach. However, as this did not yield significant performance improvements, They opted to retain the original pre-trained RETFound model to avoid overfitting.

Post-processing was minimal, with the only adjustment being the reassignment of the Uninterpretable category to Stable, due to class imbalances. While effective for competition purposes, they note that uninterpretable predictions should ideally be flagged for further medical review.

\end{document}